\pdfoutput=1

\documentclass[11pt]{article}

\usepackage[final]{acl}

\usepackage{times}
\usepackage{latexsym}
\usepackage{floatrow}
\usepackage{hyperref}

\usepackage{caption}
\usepackage{subcaption}

\usepackage[T1]{fontenc}

\usepackage[utf8]{inputenc}

\usepackage{microtype}

\usepackage{inconsolata}

\usepackage{pifont}
\newcommand{\xmark}{\ding{55}}
\usepackage{inconsolata}
\usepackage{tabularx}
\usepackage{pdfpages}
\usepackage{colortbl}
\usepackage{graphicx}
\usepackage{array}
\usepackage{algorithm}
\usepackage{algorithmic}
\usepackage{amsmath}
\usepackage{amssymb}
\usepackage{color}
\usepackage{multirow}
\usepackage{array}
\newcolumntype{M}[1]{>{\centering\arraybackslash}m{#1}}
\usepackage{booktabs}
\usepackage{tikz}
\usepackage{makecell}
\usepackage[frozencache,cachedir=.]{minted} 
\usepackage{fancyhdr}
\usepackage{lastpage}
\usepackage{xspace}
\usepackage{wrapfig}

\appto\appendix{\addtocontents{toc}{\protect\setcounter{tocdepth}{0}}}

\appto\listoffigures{\addtocontents{lof}{\protect\setcounter{tocdepth}{1}}}
\appto\listoftables{\addtocontents{lot}{\protect\setcounter{tocdepth}{1}}}

\definecolor{color1}{rgb}{1, 1, 1} 
\definecolor{color2}{rgb}{1, 1, 1} 
\definecolor{color3}{rgb}{1, 1, 1} 
\definecolor{color4}{rgb}{1, 1, 1} 
\definecolor{color5}{rgb}{1, 1, 1} 
\definecolor{color6}{rgb}{1, 1, 1} 
\definecolor{color7}{rgb}{1, 1, 1} 

\definecolor{Gray}{gray}{0.8}

\definecolor{light-gray}{gray}{0.95}
\usepackage{tcolorbox}
\newfloatcommand{capbtabbox}{table}[][\FBwidth]

\usepackage{arydshln}

\newcolumntype{P}[1]{>{\centering\arraybackslash}p{#1}}

\newcommand{\Thepage}{\thepage \xspace\xspace}

\usepackage[inline]{enumitem}
\title {Under the Surface: Tracking the Artifactuality of LLM-Generated Data}

\author{
\normalfont
\normalsize
 Debarati Das\thanks{Core Contributor $^\spadesuit$Project Lead }$^\spadesuit$, Karin de Langis$^*$, Anna Martin-Boyle$^*$, Jaehyung Kim$^*$, Minhwa Lee$^*$, Zae Myung Kim$^*$,
 \\ \normalsize Shirley Anugrah Hayati, Risako Owan, Bin Hu, Ritik Sachin Parkar, Ryan Koo, \\
 \normalsize Jong Inn Park, Aahan Tyagi, Libby Ferland, Sanjali Roy, Vincent Liu, \\
\normalsize Dongyeop Kang \\\\
\small Minnesota NLP Lab, University of Minnesota\\
\small  \texttt{\{das00015, dento019, mart5877, kim01746, lee03533, kim01756, dongyeop\}@umn.edu}}

\begin{document}
\maketitle

\begin{abstract}\vspace{-2mm}
This work delves into the expanding role of large language models (LLMs) in generating artificial data. LLMs are increasingly employed to create a variety of outputs, including annotations, preferences, instruction prompts, simulated dialogues, and free text. As these forms of LLM-generated data often intersect in their application, they exert mutual influence on each other and raise significant concerns about the quality and diversity of the artificial data incorporated into training cycles, leading to an \textit{artificial data ecosystem}. To the best of our knowledge, this is the first study to aggregate various types of LLM-generated text data, from more tightly constrained data like ``task labels'' to more lightly constrained ``free-form text''. We then stress test the quality and implications of LLM-generated artificial data, comparing it with human data across various existing benchmarks. Despite artificial data's capability to match human performance, this paper reveals significant hidden disparities, especially in complex tasks where LLMs often miss the nuanced understanding of intrinsic human-generated content. This study critically examines diverse LLM-generated data and emphasizes the need for ethical practices in data creation and when using LLMs. It highlights the LLMs' shortcomings in replicating human traits and behaviors, underscoring the importance of addressing biases and artifacts produced in LLM-generated content for future research and development.
All data and code are available on our project page.\footnote{\href{https://minnesotanlp.github.io/artifact}{https://minnesotanlp.github.io/artifact}} 
\end{abstract}

\section{Introduction}\label{sec:intro}

\begin{figure}[h!]
  \centering
  \vspace{-5mm}
  \includegraphics[width=0.85\columnwidth,clip, trim=0.1cm 2.0cm 0.1cm 0.6cm]{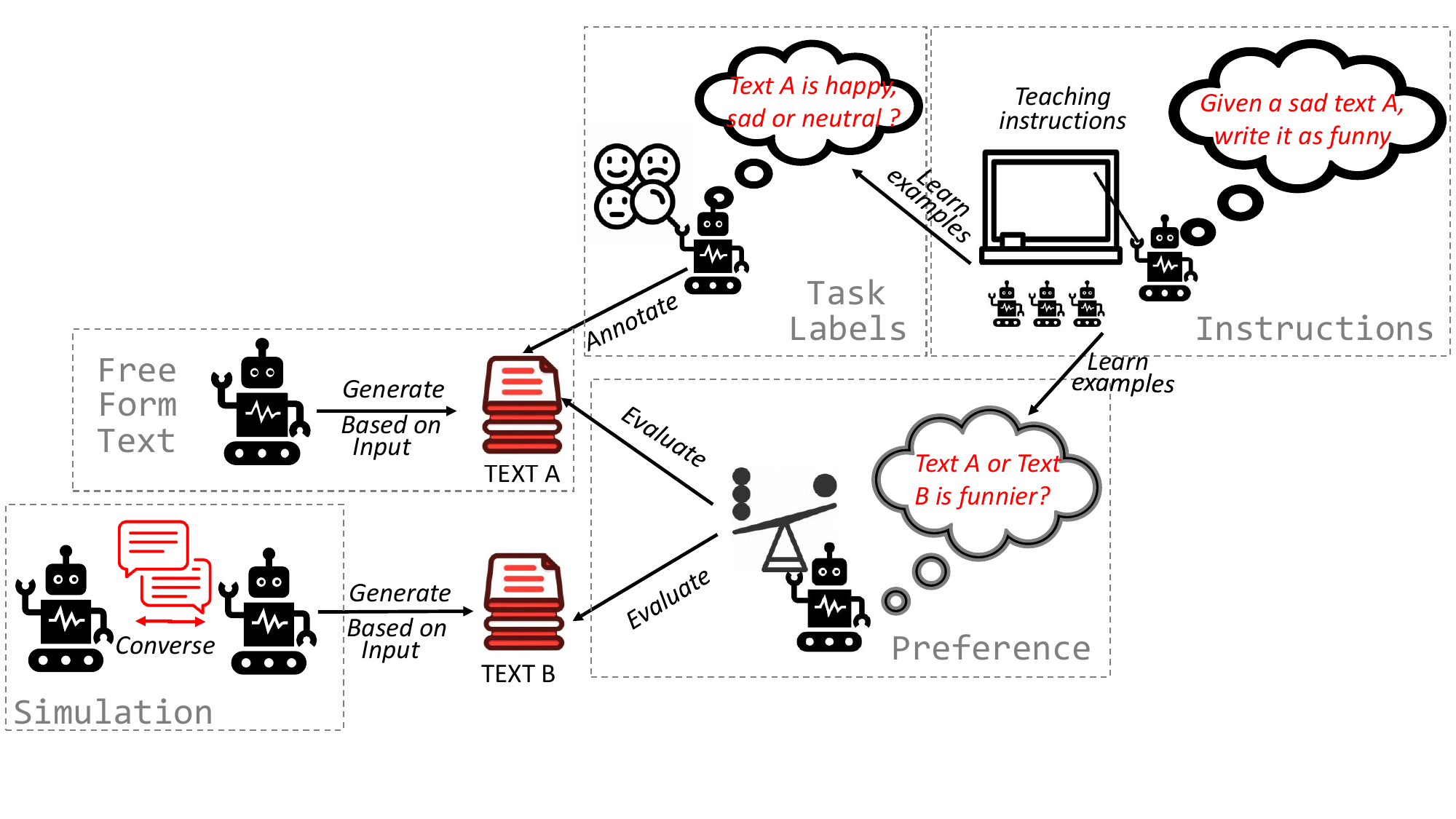}
  \caption{
  An example of an artificial data ecosystem in which LLMs are increasingly employed to create a variety of outputs, including annotations, preference labels, instruction prompts, simulated dialogues, and free text. As these forms of LLM-generated data are often intertwined in their application, they exert mutual influence on each other within interconnected use cases. This interdependence raises significant concerns about the quality and diversity of the artificial data incorporated into training cycles. \textit{In this LLM ecosystem, there is a risk that AI systems may become predominantly or entirely dependent on artificially generated inputs.}
  }
  \label{fig:llm_world}
\end{figure}

\noindent
The advancement of large language models (LLMs) has led to their widespread adoption for complex data generation tasks once exclusive to humans, such as writing, labeling, and data evaluation \cite{ding2022-is-gpt3-good, moller2023prompt, huang2023learning}. This proliferation of artificial data, especially from models like ChatGPT, calls for detailed research to understand the capabilities and limitations of LLMs in various economically and efficiently executed data-intensive tasks. LLM-generated artificial data can match human data in performance across various benchmarks, yet the full implications and impact of its use in subsequent tasks remain elusive.

As illustrated in Figure \ref{fig:llm_world}, LLMs are now being used to generate data for many use cases including annotation, evaluation, and generation. Artificial data generated from one LLM could be used for one or more related use cases, leading to the rise of an LLM ecosystem. Limited research has highlighted the disparity between LLM-generated labels and human annotations \cite{reiss2023testing}, prompting calls for cross-domain evaluation against human data. There is no consensus on how artificial data differs from human data because it is a fairly new and highly fluent source of data \cite{sadasivan2023can}. Recent studies, notably those by \citet{shumailov2023curse} and \citet{briesch2023large}, have highlighted concerns regarding the quality and diversity of artificial data employed in, and subsequently produced by, training loops because of the problem of mode collapse, where models trained on their outputs might lose touch with the true data distribution. 

The growing dependence on such data for its convenience and lower costs raises concerns about an emerging cycle of AI- reliant solely on artificial inputs. These concerns are further fueled by speculations of the next version of GPT training on its own generated data\footnote{Tweet Link: \url{https://shorturl.at/kmAY2}} and a growing body of research highlighting troublesome aspects of artificial data. For instance, numerous studies show that various marginalized groups are often portrayed in biased and harmful ways in LLM-generated data \citep{hosseini-etal-2023-empirical,gallegos2023bias,wald2023exposing,pmlr-v139-liang21a,10.1145/3597307,10.1145/3531146.3533088,10.1145/3442188.3445922}, and other research highlights the difficulty in identifying misinformation produced by LLMs, such as incorrect articles \citep{pan-etal-2023-risk, chen2023llmgenerated}. This destructive cycle could amplify existing biases and stereotypes, reducing the models' ability to accurately reflect human data's diversity and complexity. With the increasing use of LLMs across various domains, the persistent engagement with these generative processes threatens the preservation of human diversity over time.

These issues underpin our inquiry into the nature of data produced by LLMs. Our objective is not an in-depth empirical analysis of LLM-generated data but rather a preliminary examination to detect certain anomalies that might suggest biases or a trend toward mode collapse. By adopting this method, we can examine the attributes of artificial data and pinpoint areas that might benefit from more thorough research. Through this investigation of such anomalies, our aim is to offer initial observations on the characteristics of LLM-generated data, especially in comparison to human-generated data. Additionally, we seek to emphasize the need for careful and responsible methods in creating datasets and developing LLMs, particularly given their growing relevance to significant societal issues.


\subsection{Research Focus, Questions, and Scope}

This work contributes to this understanding by taking a broad inventory of artificial data, possible ``artifacts''(anomalous behavior or divergences from the corresponding human data), and the effects of using this data. We prioritize breadth, covering five different data types (examples of each shown in Figure \ref{fig:figure1}), rather than depth, to understand overarching themes in artificial data. 

To forward the understanding of artificial data and its effects, the research questions we investigate are: 
\begin{itemize}[noitemsep,topsep=5pt]
\item \textbf{RQ1} What is the nature of SOTA artificial data? How does it compare to human data, and what artifacts does it contain?
\item \textbf{RQ2} Does training on artificial data compromise performance outcomes compared to similar human-generated data? 
\item \textbf{RQ3} How specific are the artifacts of artificial data to certain types of data produced by LLMs, and how much do they apply to all types of data generated by LLMs?
\end{itemize}

To address RQ1, we examine \textit{first-order effects} for each data type. We define first-order effects as the nature of the data itself, i.e., any differences or inconsistencies between the artificial data and the corresponding human data. To address RQ2, we examine \textit{second-order effects}, which we define as any differences or inconsistencies between the outcomes of NLP pipelines containing artificial data and the outcomes from corresponding pipelines containing human data. We address RQ3 through an overview of the results of all first-and second-order experiments.

Our research is designed to offer preliminary insights into the quality and effects of data produced by LLMs rather than providing final, conclusive findings. We recognize the challenge of dealing with stochastic outputs, a common issue in LLM studies. Our paper examines a broad spectrum of LLM-generated data types from existing benchmarks, whose variety and complexity make it difficult to create a controlled experiment that adequately addresses all their intricacies and variables. Furthermore, such an endeavor would demand significant time, financial, and computational resources. The primary focus of our study is instead to identify and discuss the differences between LLM-generated and human-generated outputs, to identify potential areas for further research, and to bring attention to the ethical considerations inherent in LLM-generated data. 

\begin{figure}[t!]
    \centering
    \includegraphics[width=0.75\linewidth, trim={0.5cm 0.5cm 0.5cm 0cm},clip]{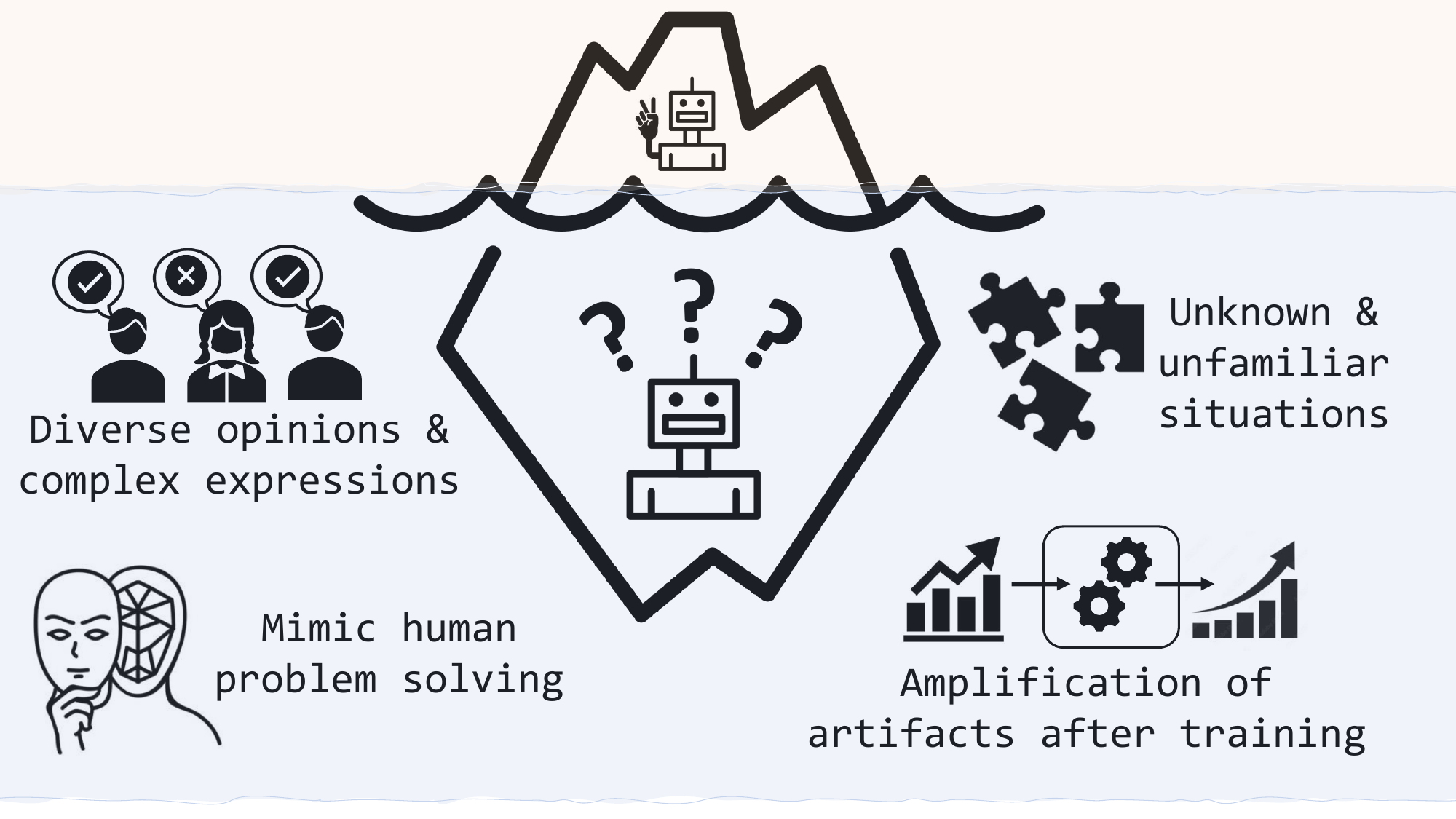}
    \caption{Large Language Models often falter in unfamiliar scenarios, exhibiting biases and a lack of nuanced understanding of complex human opinions, and thus struggle to replicate human behavior in tasks such as problem-solving accurately. This leads to \textit{decreased performance in models trained on LLM-generated data containing biases and artifacts}, underscoring the critical need to monitor and address these issues in LLM-generated content.} 
    \label{fig:iceberg}
\end{figure}
\subsection{Main Findings}
We contend that the general assertion that LLM-generated data matches human data is \textit{superficial}, as we find that when we look below the surface (see Figure \ref{fig:iceberg}), there are many dimensions in which LLM-generated data does not match human data:
\begin{itemize}[noitemsep,topsep=5pt]
\item LLMs tend to fall short in complex tasks, often \textit{missing the subtleties and nuanced understanding} found in human-generated data, particularly in areas like minority perspectives and complex preferences. 
\item LLM-generated texts frequently \textit{lack human traits} such as humor and sarcasm and \textit{do not mirror the structural complexities and styles} typical in human writing and conversations, especially in problem-solving scenarios. 
\item These shortcomings become evident as LLMs struggle in \textit{unfamiliar situations}, generating \textit{incorrect outputs}, \textit{exhibiting biases}, and failing to accurately replicate human behavior, leading to \textit{decreased performance in models trained on such LLM-generated data}. 
\end{itemize}
This highlights the essential need to monitor and address the limitations and biases inherent in LLM-generated content.
We summarize more detailed major findings in Section \ref{sec:sec4}.

\subsection{Contributions}
\noindent Our main contributions are: 
\begin{itemize}[noitemsep,topsep=5pt]
\item We present a \textit{pioneering effort in gathering a diverse range of text data produced by LLMs}, covering everything from more structured ``task labels'' to more open-ended ``free-form text.'' This comprehensive collection is significant as it allows for a unique and holistic examination of LLM outputs. By encompassing a wide spectrum of data types, our study offers a broader understanding of LLM capabilities and limitations across different contexts and constraints. This is a crucial contribution to the field, as it provides insights into how LLMs perform under varying degrees of structure and freedom, which is essential for both understanding their current state and guiding future improvements and applications.
\item We \textit{aggregate and conduct comprehensive stress tests} on various data generated by LLMs using the existing benchmarks, offering a thorough evaluation of the quality, consistency, and reliability of LLM outputs across diverse models and scenarios, thereby providing a groundbreaking insight into their strengths and weaknesses for future research and development.
\item Our research \textit{emphasizes the critical need for responsible and ethical practices in creating and using LLM-generated data}, advocating for collaborative efforts among stakeholders to address biases, increase diversity, and deepen the understanding of complex human opinions in LLM outputs, thereby ensuring their development benefits society ethically and sustainably.
\end{itemize}



\subsection{Structure of Paper}
We structure the rest of the paper as follows: Section \ref{sec:sec2} discusses the different types of LLM-generated data we have explored. Section \ref{sec:sec3} provides an overview of the stress testing methods used in our analysis. Section \ref{sec:sec4} elaborates on key research gaps we unearthed from previous work and our broad takeaways from our research. Sections 5 - 9 discuss each data type and related experiments and analysis. Section \ref{sec:sec9} discusses the overall findings from our analysis of LLM-generated data and specific findings from each data type. This section also discusses alternative evaluation methodologies for each data type, suggestions for mitigating bias, and a call for action to the creators of this data. Section \ref{sec:sec10} is our conclusion from this research. Section \ref{sec:contributions} elaborates on the contributions of the authors of this research.
\newpage
{\small\tableofcontents}

\newpage
\section{Types of Artificial Data}\label{sec:sec2}

\begin{figure*}[h!]
\hspace*{-0.5cm} 
  \centering
  \includegraphics[width=1.1\textwidth]{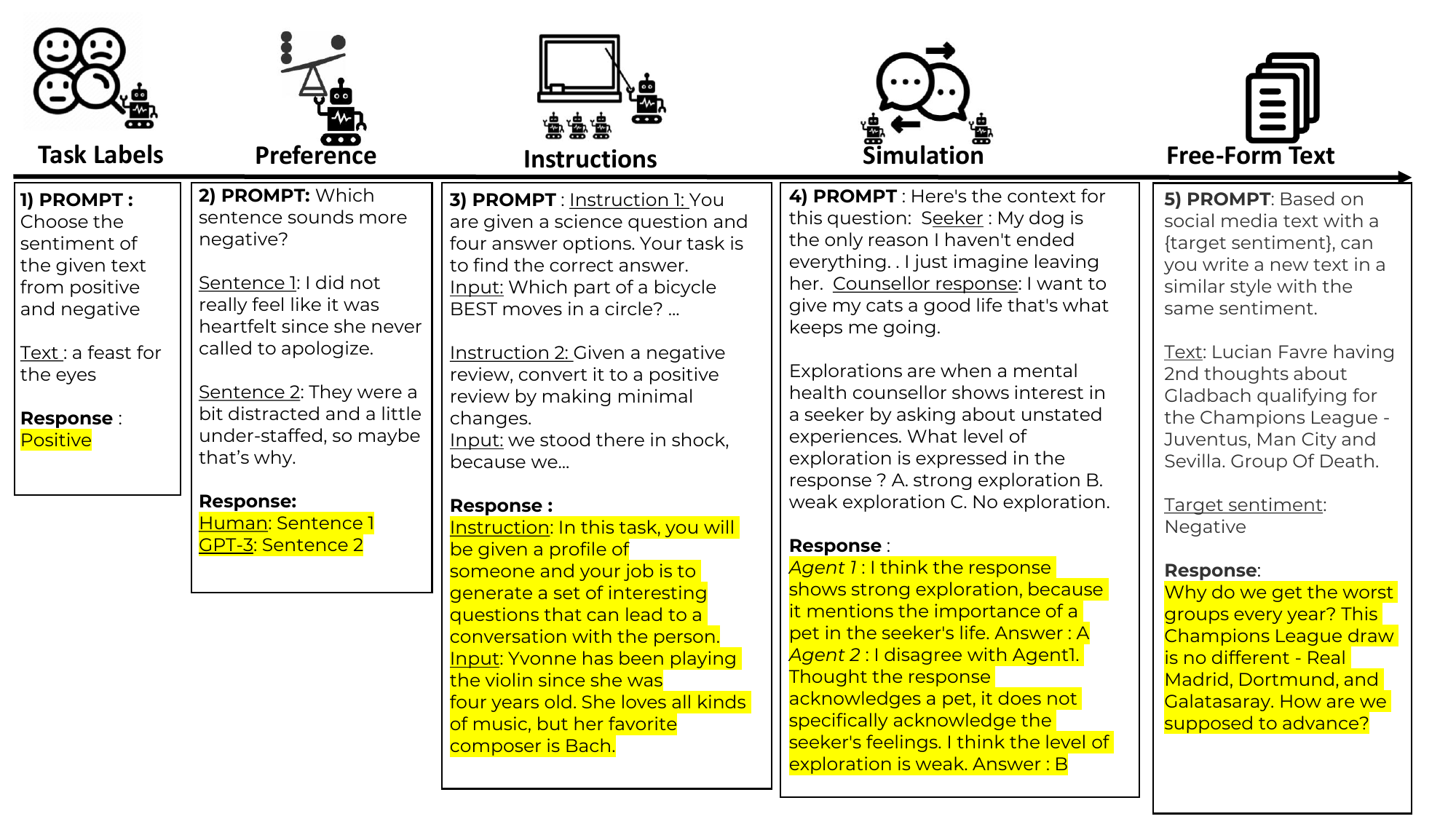}
  \caption{Overview of the \textit{five types of LLM-generated data and associated examples from the most tightly constrained output (left) to the most lightly constrained output (right)} -- (1) Task Labels, (2) Preference, (3) Instructions, (4) Simulation, and (5) Free Form text. Sources for these examples in order: \citet{diaz2018addressing}, \citet{kim2023p2c},  \citet{honovich2022unnatural}, \citet{liang2023encouraging} and  \citet{moller2023prompt}.
  }
  \label{fig:figure1}
\end{figure*}


\noindent
In our experiments, we closely examine and analyze five types of artificial data, categorizing them to differentiate their applications and functionalities. This categorization enables our understanding of LLM strengths and limitations in different contexts. The five types are: 
\begin{itemize}[noitemsep,topsep=5pt]
\item \textbf{Task Labels} for classification tasks over textual input, bypassing the need for human annotators.
\item \textbf{Preference}, specifically evaluating which text is better, is useful for human alignment tasks such as training reward models used in Reinforcement Learning from Human Feedback (RLHF).
\item \textbf{Instruction prompts} written by LLMs for instruction fine-tuning, eliminating the need for humans to write comprehensive sets of diverse instructions.
\item \textbf{Simulation}, where two LLM agents converse, has the potential for enhancing model performance and simulating intricate social interactions, according to initial studies.
\item \textbf{Free-Form Text} written by an LLM, often used to compensate for data scarcity issues and used for pretraining or finetuning.
\end{itemize}

We defined boundaries between different types of LLM-generated text data based on the level of output constraint, ranging from tightly constrained (such as task labels data, which consists of discrete labels for a task) to lightly constrained (like free-form text data, where output is freely generated from an input). 

Examples of each data type are further illustrated in Figure ~\ref{fig:figure1}. Table \ref{tab:tab1_part2} provides some more specific examples from some of the datasets we analyzed in our study. This spectrum of data types considered captures the wide-ranging capabilities and applications of large language models, each representing a specific type of output and process integral to the functioning of LLMs. This classification helps understand how each data type contributes to the broader AI and machine learning development landscape.
To make our study reproducible, we publicly release the datasets we studied and the artificial data we created in HuggingFace's data repository:
\href{https://huggingface.co/datasets/minnesotanlp/LLM-Artifacts}{https://huggingface.co/datasets/minnesotanlp/LLM-Artifacts}

\begin{table*}[htbp!]
\small
\centering
\begin{tabularx}{\textwidth}{@{}X@{}}
\toprule
\rowcolor{Gray} \multicolumn{1}{c}{\textbf{Task Labels}} \\
\textsc{Social Chemistry 5} \cite{forbes2020social} \newline 
     {\begin{tabularx}{\textwidth}{X}
    \textbf{Text}:  It is wrong to be racist to other people. \newline \textbf{Choice of Labels}: [(0) Almost no one believes this, (1) People occasionally think this, (2) Controversial, (3) Common belief, (4) Universally true]
    \newline \textbf{Human Label}: 1, \textbf{GPT Label}: \textit{3}
    \end{tabularx}}\\ 
\midrule
\textsc{SBIC} \cite{sap2019social} \newline
    {\begin{tabularx}{\textwidth}{X}
    \textbf{Text}: sup bitches \newline \textbf{Choice of Labels}: [(0) No offensive, (1) maybe, (2) offensive]
    \newline \textbf{Human Label}: 0, \textbf{GPT Label}: \textit{1}
    \end{tabularx}}\\
\midrule
\rowcolor{Gray} \multicolumn{1}{c}{{\textbf{Preference}}} \\
\textsc{P2C} \citep{kim2023p2c} \newline 
    {\begin{tabularx}{\textwidth}{X}
    \textbf{Instruction: }Given a pair of two sentences A and B, which sentence sounds more negative? \newline \textbf{Sentence A: } When the guy came to the door he said it was \$24. I was disappointed. \newline \textbf{Sentence B: } Don’t try this mango tango curry.
     \newline \textbf{Human Preference: } B /  \textbf{GPT-Preference } \textit{A} 
    \end{tabularx}}\\ 
\midrule 
\textsc{CoBBLEr} \citep{koo2023benchmarking} \newline
    {\begin{tabularx}{\textwidth}{X}
    \textbf{Prompt: } Given a pair of two sentences A and B, which sentence aligns more with the instruction and the reference? \newline \textbf{Instruction: } Did people in Korea under Japanese Rule watch a lot of Iron Chef? \newline \textbf{Reference:} No. The first televisions were sold in 1946. Iron Chef started airing in 1993. \newline \textbf{Model A (GPT-4): } No, people in Korea under Japanese rule did not watch a lot of Iron Chef. This is mainly because .. \newline \textbf{Model B (Dolly): } Iron Chef was a popular Japanese cooking show that was adapted into a Korean version that was watched by people in Korea under Japanese Rule. \newline \textbf{Human Preference } Model A (GPT-4)  /  \textbf{GPT Preference: } \textit{Model B (Dolly)}
    \end{tabularx}}\\
\midrule
\rowcolor{Gray} \multicolumn{1}{c}{\textbf{Instructions}} \\
\textsc{Self-Instruct} \citep{wang2023selfinstruct}\newline
    {\begin{tabularx}{\textwidth}{X}
    \textbf{Instruction:} Replace each verbs with its past tense form.\newline\textbf{Input:} I am a programmer.\newline\textbf{Output:} \textit{I was a programmer.}  \\
    \end{tabularx}}\\
    \midrule
\textsc{Cleaned-Alpaca} \citep{alpaca}\newline
{\begin{tabularx}{\textwidth}{X} \textbf{Instruction:} Generate a sentence that reflects the tone of the following excerpt.\newline\textbf{Input:} The icy cold waters of the lake chilled her to the bone.\newline\textbf{Output:} \textit{She shivered uncontrollably as the freezing water engulfed her body, leaving her feeling numb and frozen.}
     \end{tabularx}}
    \\
\midrule
\rowcolor{Gray} \multicolumn{1}{c}{{\textbf{Simulation}}} \\
\textsc{Solo Performance Prompting\cite{wang2023unleashing}}\newline 
    {\begin{tabularx}{\textwidth}{X}
    \textbf{Task}: Use numbers and basic arithmetic operations (+ - * /) to obtain 24. You need to use all numbers, and each number can only be used once.\newline \textbf{Input}: 6 12 1 1 \newline \textbf{Participants}: AI Assistant; Math Expert\newline \textbf{Math Expert}: \textit{Let's analyze the task in detail. You need to make sure that you meet the requirement and use exactly the four numbers (6 12 1 1) to construct 24. To reach 24} {\ttfamily[Detailed methods]}\newline \textbf{AI Assistant}: \textit{Thanks for the hints! Here's one initial solution: (12 / (1 + 1)) * 6 = 24}\newline \textbf{Math Expert}: \textit{Let's check the answer step by step} {\ttfamily[Process of checking]} \textit{The answer is not correct} {\ttfamily...}\newline \textbf{AI Assistant}: \textit{Thanks for pointing out the mistake. Here is a revised solution considering 24 can also be reached by 3 * 8: (6 + 1 + 1) * (12 / 4) = 24.}
    \end{tabularx}} \\
    \rowcolor{Gray} \multicolumn{1}{c}{\textbf{Free Form Text}} \\ 
\textsc{HC3} \cite{guo2023close} \newline 
    {\begin{tabularx}{\textwidth}{X}
    \textbf{Question}: Should I avoid credit card use to improve our debt-to-income ratio? \newline \textbf{Human Answer}: If you pay it off before the cycle closes it will look like you have 100\% available credit. So if you credit card statement closes on the 7th pay it off on the 6th in full .. \newline \textbf{GPT Answer}: \textit{It can be a good idea to avoid using credit cards if you are trying to improve your debt-to-income ratio. Your debt-to-income ratio is a measure of how much debt you have compared to your income..} 
    \end{tabularx}}
\\
\midrule
\textsc{Workers} \cite{moller2023prompt} \newline 
    {\begin{tabularx}{\textwidth}{X}
    \textbf{Prompt:} The following social media text conveys the social dimension fun. Fun in a social context is defined by experiencing leisure, laughter, and joy. Write new semantically similar examples in style of a social media comment, that show the same intent and social dimension. \newline
    \textbf{Human Input:} on the 25th of march can we celebrate zayn's time in the band and create cute hashtags and post funny things and really cute things pls \newline
    \textbf{GPT Generation:} \textit{Hey guys, let's all share our favorite moments of Zayn on March 25th and use adorable hashtags to express our love for his time in the band!} 
    \end{tabularx}}
\\
\bottomrule
\end{tabularx}
\caption{Examples of paired human and LLM responses in different data types. The LLM-generated data for each data type is italicized.}
\label{tab:tab1_part2}
\end{table*}

\newpage
\section {Types of Stress Testing Methods}\label{sec:sec3}
This section provides an overview of the stress testing methods applied to analyze LLM-generated data.

\begin{figure*}[ht!]
  \centering
  \includegraphics[width=0.8\textwidth,trim=3cm 2.2cm 2.2cm 2.8cm,clip]{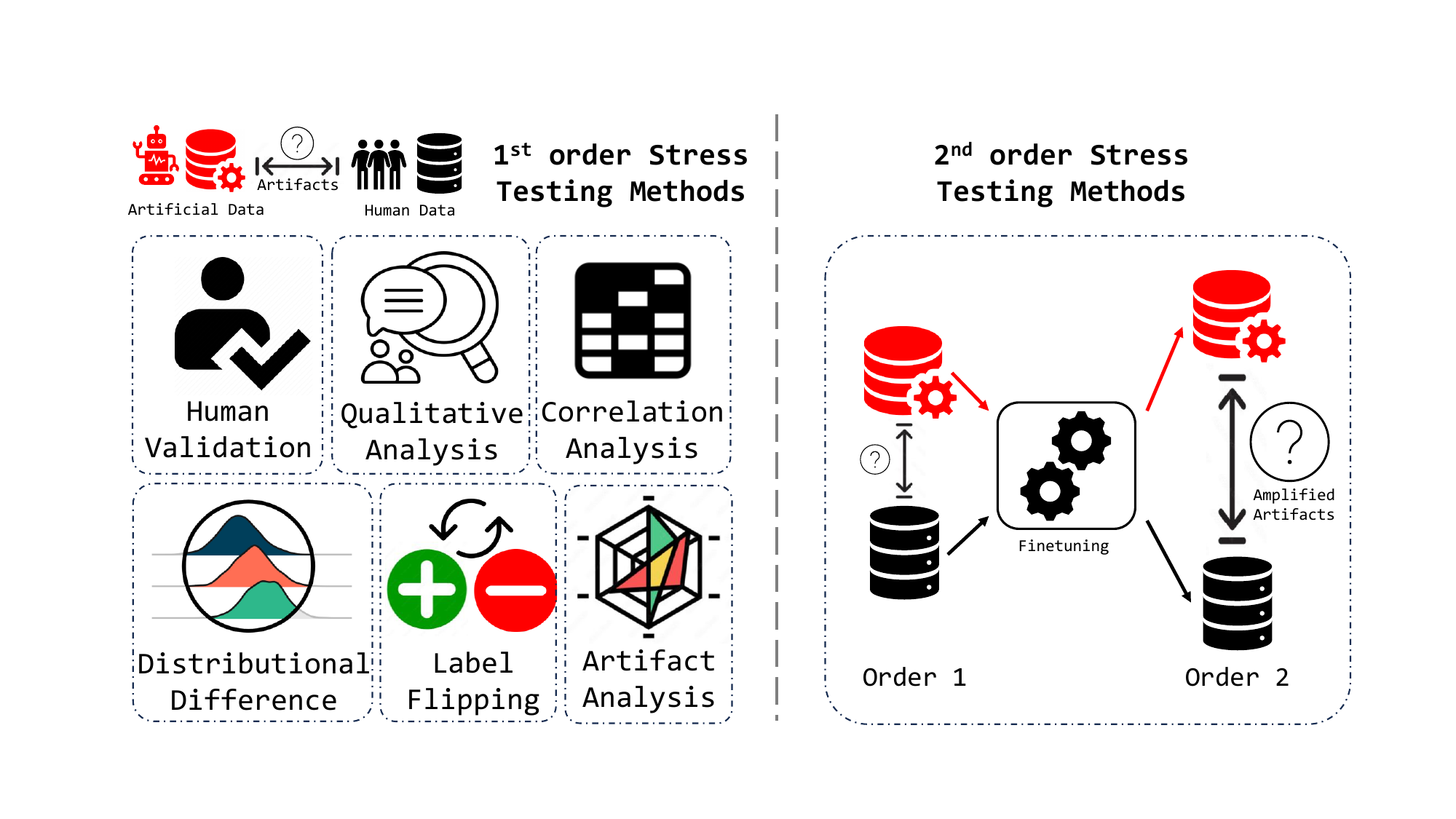}
  \caption{Methods to stress test LLM-generated data for the first- and second-order experiments. In summary, the first-order experiments investigate the data \textit{``as-is,''} for example, focusing on their distributional differences and correlation patterns among human- and LLM-generated data; validating and analyzing using manual inspection; and counting how often labels flip between the original human and the resulting machine text. The second-order experiments involve fine-tuning LLMs on the machine-generated data and investigating whether the existing artifacts or biases are amplified.}
  \label{fig:figure2}
\end{figure*}

Figure \ref{fig:figure2} illustrates the range of approaches utilized in our paper for stress testing artificial data, while Table \ref{tab:stress_testing_overview} summarizes which methods were applied to different data types. 
 Key methods include analyzing distributional differences between human and LLM-generated data to uncover biases and stylistic variances, thereby assessing how well machines mimic human language and thought. Qualitative analysis probes deeper into machine-generated text, evaluating aspects like bias, coherence, and creativity to gauge subjective quality. On the other hand, human validation is a quality check, ensuring that machine outputs are contextually and culturally appropriate. The label-flipping technique is used to study changes in label predictions between human and machine inputs, highlighting inconsistencies in the model's understanding. Lastly, artifact analysis scrutinizes LLM-generated data to identify unique patterns and biases crucial for enhancing content diversity and authenticity. 

 We discover that techniques such as \textit{artifact analysis} are particularly suited for less restricted LLM-generated data like simulations or free-form text. In contrast, \textit{label flipping} is effective for more constrained data types. Methods like \textit{distributional difference} and \textit{label flipping} are useful for comparing various LLM-generated data types with human data. While most data types are analyzed for secondary effects, simulation data is still in the early stages of examination and requires more foundational analysis. Together, these methods comprehensively evaluate LLM-generated data, uncovering trends, biases, and improvement opportunities.


\begin{table*}[ht]
\footnotesize
\centering
\begin{tabular}{P{0.14\linewidth}P{0.08\linewidth}P{0.10\linewidth}P{0.08\linewidth}P{0.10\linewidth}P{0.10\linewidth}P{0.09\linewidth} : P{0.10\linewidth}}
\rowcolor{Gray}
\textit{} & \textit{Artifact Analysis} & \textit{Distributional Difference} & \textit{Label Flipping} & \textit{Correlation Analysis} & \textit{Human Validation} & \textit{Qualitative Analysis} & \textit{2nd order effect} \\
\midrule

\rowcolor{color3}\textbf{Task Labels} (\S\ref{sec:task_labels})& \xmark  & $\checkmark$ & $\checkmark$  & $\checkmark$ & \xmark & $\checkmark$ & $\checkmark$\\
\midrule
\rowcolor{color5}\textbf{Preference} (\S\ref{sec:preference})& \xmark & $\checkmark$ & \xmark  & $\checkmark$  & \xmark & $\checkmark$ & $\checkmark$\\
\midrule
\rowcolor{color4}\textbf{Instructions} (\S\ref{sec:instructions})& \xmark & $\checkmark$ & \xmark  & \xmark & $\checkmark$ & $\checkmark$  & $\checkmark$ \\
\midrule

\rowcolor{color6}\textbf{Simulation} (\S\ref{sec:simulation})& $\checkmark$ & \xmark & $\checkmark$ & $\checkmark$   & $\checkmark$  & $\checkmark$ & \xmark \\
\midrule
\rowcolor{color2}\textbf{Free Text} (\S\ref{sec:free_text}) & $\checkmark$  & $\checkmark$ & $\checkmark$ & \xmark & \xmark & $\checkmark$ & $\checkmark$ \\

\bottomrule
\end{tabular}
\caption{Overview of stress testing methods used across data types}
\label{tab:stress_testing_overview}
\end{table*}

\newpage
\section {Overall Takeaways}\label{sec:sec4}

\begin{table*}[ht]
\small
\centering
\begin{tabular}
{p{0.11\linewidth}p{0.41\linewidth}p{0.41\linewidth}}
\rowcolor{Gray}
\textbf{} & 
\hfil\textit{Challenges Typically Faced} & \hfil\textit{Research Gap Unearthed}\\
\midrule

\rowcolor{color3}\textbf{Task Labels} & Subjectivity loss and consensus bias in LLM-annotated task labels \citep{ding2022-is-gpt3-good,wan2023everyone} & Disregarding minority opinion representation in LLM datasets \\
\midrule
\rowcolor{color5}\textbf{Preference} & LLM bias in evaluating text pairs \citep{zheng2023judging,koo2023benchmarking} & Insufficient scrutiny of LLM locality bias concerning preferences \\
\midrule
\rowcolor{color4}\textbf{Instructions} & 
Error amplification due to overstated diversity \citep{honovich2022unnatural,kung-peng-2023-models} & Lack of attention to the study of diversity and noise in instructions \\
\midrule
\rowcolor{color6}\textbf{Simulation} & Underexplored simulated LLM agents  \citep{li2023camel,chen2023reconcile} & Absence of investigation into artifacts in LLM simulated data, such as role flipping and digression.\\
\midrule
\rowcolor{color2}\textbf{Free-Form Text} & LLM-text indistinguishable from human-written text due to convincing fluency and coherency. \citep{sadasivan2023aigenerated,krishna2023paraphrasing} & Neglect of higher-level figurative attributes and structural document features \\

\bottomrule
\end{tabular}
\caption{We highlight the challenges that are typically faced with each LLM-generated data, as seen through related work.  We then unearth some unexplored research gaps from the related work for each type, and our paper delves into these gaps, conducting thorough \textit{analysis and stress testing to enhance the reliability of LLM-generated data.}}
\label{tab:transposed_related_summary}
\end{table*}

\noindent
We survey each type of LLM-generated data and explore challenging aspects (given in Table \ref{tab:transposed_related_summary}) of each data type to thoroughly evaluate the reliability and practicality of the data produced by LLMs. Table \ref{tab:data_and_exp_setup} summarizes our datasets utilized in the experiments, the rationale behind choosing these datasets, and the experiments that we conducted (1st and 2nd order). Table \ref{tab:transposed_related_summary} summarizes the five distinct data types in the landscape of LLM-generated data, each with its unique set of challenges that is mentioned across previous research. We then unearthed unexplored research gaps from this survey and focused our experiments on these gaps.

Task label data, heavily influenced by majority annotations, typically lacks diversity in opinions, particularly in subjective tasks. We tackle this by comparing label distributions between human and LLM outputs and conducting variation analysis. We also observe the amplification of existing artifacts through our second-order effect analysis. In the case of preference data, used as a proxy for human preferences, we assess the distributional differences between human and LLM linguistic preferences. Instruction-based data from LLMs, which usually lacks diversity in out-of-distribution tasks, presents high error rates. Addressing this, we employ human validation, error distribution analysis, and second-order effect analysis. We analyze challenges such as unexpected agent behaviors in the lesser-studied domain of simulation data. These are scrutinized through artifact and correlation analysis, coupled with human validation. Lastly, we find that free-form text data often faces the challenge of distinguishing between human and LLM outputs, with previous work primarily focusing on detailed analysis of linguistic features. Our stress testing approach, in contrast, includes artifact analysis of high-level attributes, label flipping, and contrasting human and LLM discourse patterns. 

\begin{table*}[b!]
    \centering
    \vspace{-5mm}
    \footnotesize
    \begin{tabular}{p{0.32\linewidth} p{0.25\linewidth} p{0.18\linewidth} p{0.16\linewidth}} 
\rowcolor{Gray}
\hfil\multirow{1}{*}{\textit{Datasets}} & \hfil\multirow{1}{*}{\textit{Rationale}} & \hfil\multirow{1}{*}{\textit{1st Order}} & \hfil\multirow{1}{*}{\textit{2nd Order}} \\
       \midrule
        \multicolumn{2}{@{}l}{\textbf{Task Labels}} & &  \\
        \begin{tabular}[l]{@{}l@{}}{} \textsc{Sentiment}  \cite{diaz2018addressing} \\ \textsc{SBIC} \cite{sap2019social} \\ \textsc{GHC} \cite{kennedy2022introducing}\\ \textsc{Social Chem.} \cite{forbes2020social}\\ \end{tabular} & 
        \vspace{-10mm}Task Subjectivity, individual-level annotations & \vspace{-10mm}Biased toward majority labels & \vspace{-10mm}Checking majority bias after fine-tuning \\ 
        \midrule
       \multicolumn{2}{@{}l}{\textbf{Preference}} & &  \\
       \begin{tabular}[l]{@{}l@{}} {} \textsc{CoBBLEr} \cite{koo2023benchmarking} \\\textsc{P2C} \cite{kim2023p2c}\\\end{tabular} & 
        \vspace{-7mm}Human-machine datasets & \vspace{-7mm}Important word \& Entailment difference & \vspace{-7mm}Human-LLM preference divergence \\ 
        \midrule
        \multicolumn{2}{@{}l}{\textbf{Instructions}} & &  \\
        \begin{tabular}[l]{@{}l@{}} \textsc{Unnat. Instr.} \cite{honovich2022unnatural} \\ \textsc{Self-Instruct} \cite{wang2023selfinstruct} \\ \textsc{Alpaca-cleaned} \cite{alpaca} \\ \textsc{GPT-4-LLM} \cite{peng2023instruction} \\ \textsc{FLAN} \cite{weifinetuned}\\ \textsc{Dolly} \cite{DatabricksBlog2023DollyV2} \\ 
        \textsc{SuperNat. Instr.} \cite{wang-etal-2022-super} \\ \textsc{Instr. in the Wild} \cite{instructionwild}
        \end{tabular} & 
        \vspace{-14mm}
        \begin{tabular}[l]{@{}l@{}} Variety of instruction \\ types and domains. \\ Benchmarking with \\ Human-created \\ Datasets
         \end{tabular}  & \vspace{-14mm}LLMs produce errors in the instruction, input, and output.  & \vspace{-14mm}Impact of input inaccuracies \\ 
        
        \midrule
       \multicolumn{2}{@{}l}{\textbf{Simulation}} & &  \\
        \begin{tabular}[l]{@{}l@{}} {}\textsc{Grid-World}  \cite{wang2023unleashing} \\ \textsc{CAMEL} \cite{li2023camel}\\ \end{tabular} & 
        \vspace{-7mm}
        Simulated agent conversations with agents assigned with roles or personas & \vspace{-7mm}Role flipping, Topic digression & \vspace{-7mm}N/A \\ 
        \midrule
       \rowcolor{color2} \multicolumn{2}{@{}l}{\textbf{Free-From Text}} & &  \\
       \begin{tabular}[l]{@{}l@{}} \textsc{HC3} \cite{guo2023close} \\ \textsc{Scarecrow} \cite{dou-etal-2022-gpt} \\ \textsc{Deepfake} \cite{li2023deepfake} \\ \textsc{Workers} \cite{moller2023prompt} \end{tabular}
        & 
        \vspace{-10mm}
        Pair-wise generated texts from both humans and LLMs & \vspace{-10mm}High-level textual attribute  and discourse structure difference & \vspace{-10mm}Classification performance \\ 
        
        \bottomrule
    \end{tabular}
    \caption{Datasets and summary of experiments. For each data type, we look at \textit{the first- and second-order effects}. 
    }
    \label{tab:data_and_exp_setup}
\end{table*}

\subsection{Thematic Grouping of Detected Artifacts across LLM-generated Data Types}
Through our investigation into the five distinct LLM-generated data types, we notice thematic areas in which LLM-generated data falls short of its human counterpart:

\begin{enumerate}[noitemsep,topsep=5pt]

\item \textbf{LLMs demonstrate a subpar understanding of complex human opinions and interactions.}
    
    \begin{itemize}[noitemsep,topsep=5pt,leftmargin=*]
    \item While generating \textbf{task labels}, LLMs over-represent majority opinions and do not match minority opinions well. They also show a tendency to be misleadingly confident concerning sentences with age, gender, religion, and race bias. (\S\ref{sec:task_labels_first})
    
    \item In creating \textbf{free-form text}, LLMs exhibit styles that differ significantly from human approaches in diverse social contexts. Moreover, unlike human discourse, which varies widely across domains, LLM discourse patterns remain relatively consistent. (\S\ref{sec:free_text_first}).
    
    \item For elicitation of \textbf{preferences}, we find that LLM preferences are tightly coupled with standalone lexical cues, whereas human preferences appear to take a more holistic approach (\S\ref{sec:preference_first}).
    \end{itemize}

\item \textbf{LLMs struggle to respond effectively when faced with unknown or unfamiliar situations.}

    \begin{itemize}[noitemsep,topsep=5pt,leftmargin=*]
    \item In LLM \textbf{simulations} (i.e. conversations between two LLM agents), one of the most common errors is role flipping, or the agents swapping their assigned roles. This happens most frequently when the agent becomes confused, indicating that the simulation breaks down when an agent does not understand how to respond (\S\ref{sec:simulation_first}).
    \item     
    In \textbf{instruction} writing, LLMs may provide incorrect outputs for a written instruction rather than an output that indicates uncertainty or lack of knowledge of the answer. In downstream training, these incorrect outputs result in more hallucinations produced by instruction-tuned models (\S\ref{sec:instructions_first}, \S\ref{sec:instruct-second}).
    \end{itemize}

\item \textbf{LLMs are deficient in accurately mirroring human behavior for particular tasks.}
    \begin{itemize}[noitemsep,topsep=5pt,leftmargin=*]
    \item In \textbf{simulations}, where LLM agents engage in conversations focused on problem-solving, these agents often stray from the main topic, negatively impacting task performance. This contrasts with human digressions, facilitating team building and contributing to more effective problem resolution. (\S\ref{sec:simulation_first})
    \end{itemize}

\item \textbf{Models trained on LLM-generated data containing the above issues have degraded performance.} 
    \begin{itemize}[noitemsep,topsep=5pt,leftmargin=*]
    \item 
    We find that models trained on several types of LLM-generated data -- specifically \textbf{instructions}, \textbf{free-form text}, and \textbf{preferences} -- are at risk for degraded performance when compared to models trained on corresponding human data (see experiments in \S\ref{sec:preference-second}, \S\ref{sec:instruct-second} and \S\ref{sec:free-second} ). In addition, these performance deficits appear to be tied to artifacts in LLM-generated data. 
    \item We also find that certain biases are amplified by training on LLM-generated data: LLMs trained to label based on artificial \textbf{task-labeling data} have markedly amplified biases toward majority opinions, sometimes failing to represent minority opinions entirely (see experiments in \S\ref{sec:labels-second}).
    \end{itemize}

\end{enumerate}

In light of these findings, we argue that practitioners should proceed cautiously when using artificial data and outline possible mitigation strategies to avoid the above issues in Section \ref{sec:sec9}. 

\newpage
\section{Artifacts in Task Labels}
\label{sec:task_labels}
Our research examines the use of synthetic task labels to determine whether they fail to capture minority labels adequately. We demonstrate that responses from LLMs are often uniform when dealing with controversial topics, unlike the diverse range of human responses, and they tend to represent majority labels disproportionately. Our studies reveal that these biases become more pronounced when employing synthetic task labels for training classifiers.

\paragraph{Related Work} Supervised learning, heavily reliant on accurately annotated training data, faces high costs and challenges \cite{wang2021-reduce-label-cost}, leading to increased interest in using LLMs for annotations. These LLM annotations are risky as they may introduce more noise than human labels \cite{ding2022-is-gpt3-good}. Key metrics for evaluation include accuracy, cost, efficiency, and, increasingly, diversity and subjective variance, acknowledging the need to understand diverse groups and provide personalized responses \cite{wang2021want, chan2022approaches, sampath2022seedbert, gordon2022jury, lowmanstone2023annotation}. Research indicates annotator demographics significantly impact responses, addressing biases and ethical concerns \cite{wan2023everyone, rottger2021two, santy2023nlpositionality}. Variance in task labels, once seen as noise, is now viewed as indicative of systematic and social issues \cite{fleisig2023majority, biester2022analyzing}. LLMs, however, tend to follow a ``ground truth'' during training \cite{aher2023-llm-simulate-replicate, liu2023-chatgpt-good-recommender, mckenna2023-hallucination-inference}, leading to skewed decisions, especially in moral judgment tasks \cite{dillion2023-can-ai-replace} where they struggle to reflect individuals' socio-cultural nuances.

\vspace{3mm}
\noindent
\begin{minipage}[t]{\linewidth} 
\begin{tcolorbox}[colback=gray!15]
Our research explores if task annotations generated by LLMs \textit{overlook minority opinions}, whether there's a tendency to favor majority labels, and how this \textit{bias might intensify} with training on artificial labels.
\end{tcolorbox} 
\end{minipage}

\paragraph{Data}
We use human annotations from the test partition of four datasets, \textsc{Social Chemistry} \cite{forbes2020social}, \textsc{Sentiment} \cite{diaz2018addressing}, \textsc{SBIC} \cite{sap2019social} and \textsc{GHC} \cite{kennedy2022introducing}. \\
\textbf{Justification}: These datasets were chosen due to their tasks' subjective, controversial nature and their provision of individual-level annotations. 

\subsection{First Order Experiments}
\label{sec:task_labels_first}
\begin{wrapfigure}[13]{r}{3.0in}
  \centering
  \vspace{-13mm} 
\includegraphics[width = 0.8\linewidth,clip,trim=4.5cm 0cm 3.5cm 0cm]{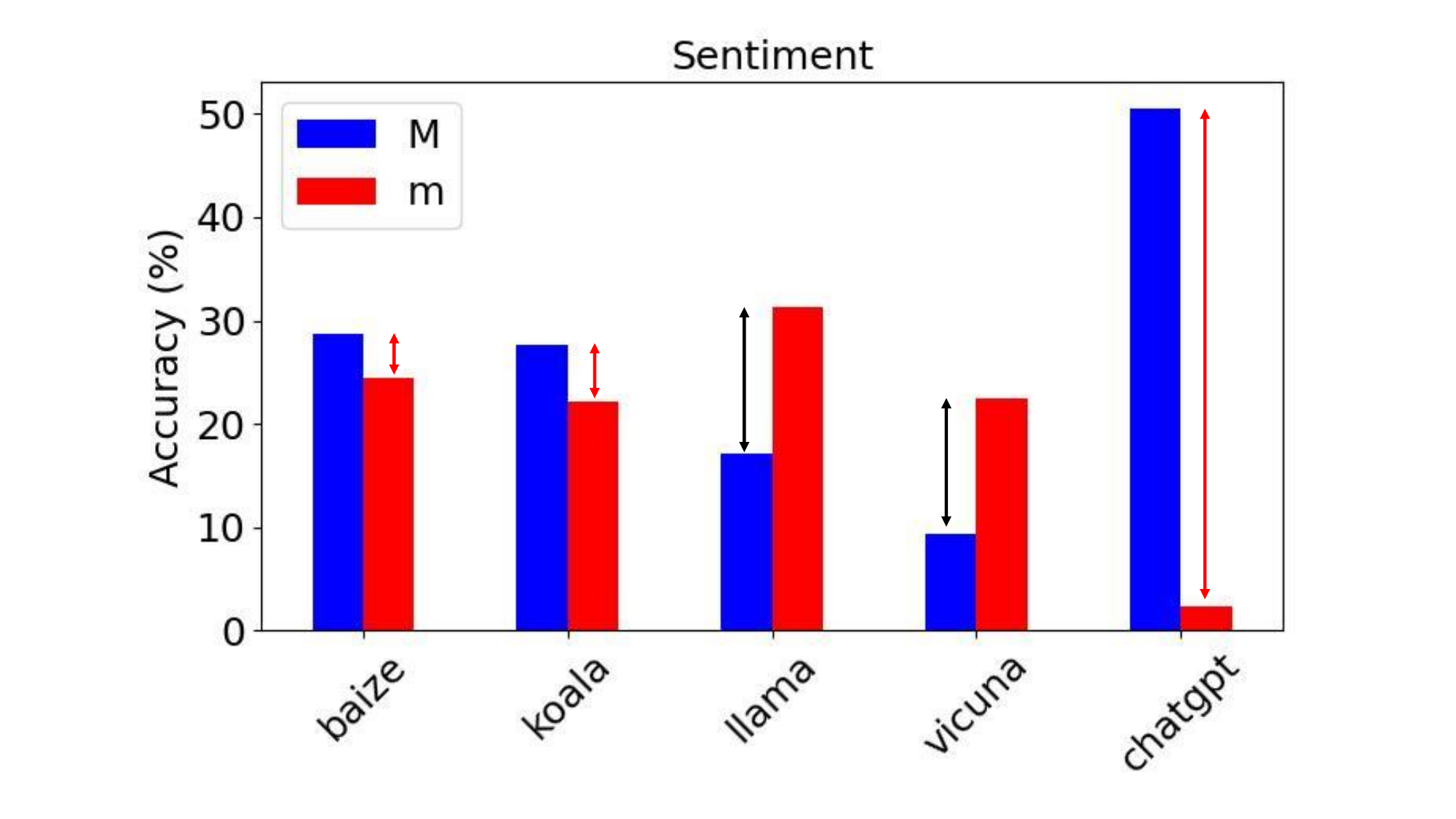}
\vspace{-5mm}
  \caption{Model accuracy for majority (M) and minority (m) match comparison on Sentiment dataset. \textit{ChatGPT has the highest majority match accuracy and the lowest minority match accuracy across all datasets, thus minority annotations tend to be represented inadequately.}}  
  \label{fig:maj_match_main}
\end{wrapfigure}

To explore LLM annotated datasets, we collect labels from various artificial ``annotators,'' i.e. instruction-tuned LLMs: Vicuna \cite{vicuna2023}, Koala \cite{koala_blogpost_2023}, Baize \cite{xu2023baize}, GPT-3.5-Turbo \cite{brown2020language}, and LLaMA2 \cite{touvron2023llama}. These models are instructed to respond as diverse Master Turk workers. The listing \ref{lst:prompt_task} in the Appendix shows an example of our prompts. 
Our methodology involves two setups: \textit{intra-model} and \textit{inter-model}. In the \textit{intra-model} setup, we use different hyperparameters for each model to create diversity in responses, simulating the variance found in human annotators. We modify the hyperparameters temperature ($t$) and top-p ($p$). The different hyperparameter variations we use are $(t, p)\in\{(0.3, 0.25), (0.3, 0.75), (0.8, 0.25), (0.8, 0.75), (1.0, 1.0)\}$. In the \textit{inter-model} setup, we group LLM "annotators" by their architectures and training methods, selecting the majority label from each model's intra-model output. This results in five distinct labels from five different models, acknowledging that some models, like GPT-3.5 Turbo, might be more capable than others, paralleling the varying proficiency levels found among human annotators. 

Our study examines whether LLM-annotated datasets accurately reflect minority opinions, particularly in subjective situations. To do this, we first measure the agreement between human and model annotations in two scenarios: one comparing predictions of majority labels and another for minority labels. We calculate the \textit{match score} as the ratio of matching annotations to the total samples and visually illustrate the distributional differences to highlight disparities between these two contexts. We assess the \textit{variation} in opinions between human and ChatGPT annotations in ordinal data, utilizing metrics like standard deviation and variance to measure the range of views and consensus; this approach is effective in highlighting the range of agreement even though ordinal data lacks uniform intervals.
In our \textit{label-flipping} analysis, we identify texts with significant disagreement between ChatGPT and human annotations. We also perform a \textit{qualitative analysis} to explore these annotation differences and reveal potential biases in the selected texts.
\vspace{-4mm}
\subsubsection{Findings on Majority and Minority Representation Comparison} In Figure \ref{fig:maj_match_main}, we observe that human and model majority match accuracy for the Sentiment dataset is highest predominantly for ChatGPT, and the minority match accuracy is also lowest for GPT3.5 Turbo. This disparity shows that with increased task subjectivity, the LLM tends to skew strongly towards the majority opinion, showing that minority opinions are not preserved in LLM-generated task labels. This trend is consistent across all datasets considered, as shown in Appendix \ref{sec:app-task_label_first}.
\noindent
\begin{figure}
 \begin{floatrow}
    \ffigbox{
        \includegraphics[width=0.48\linewidth,trim=0 0 1.7cm 0,clip]{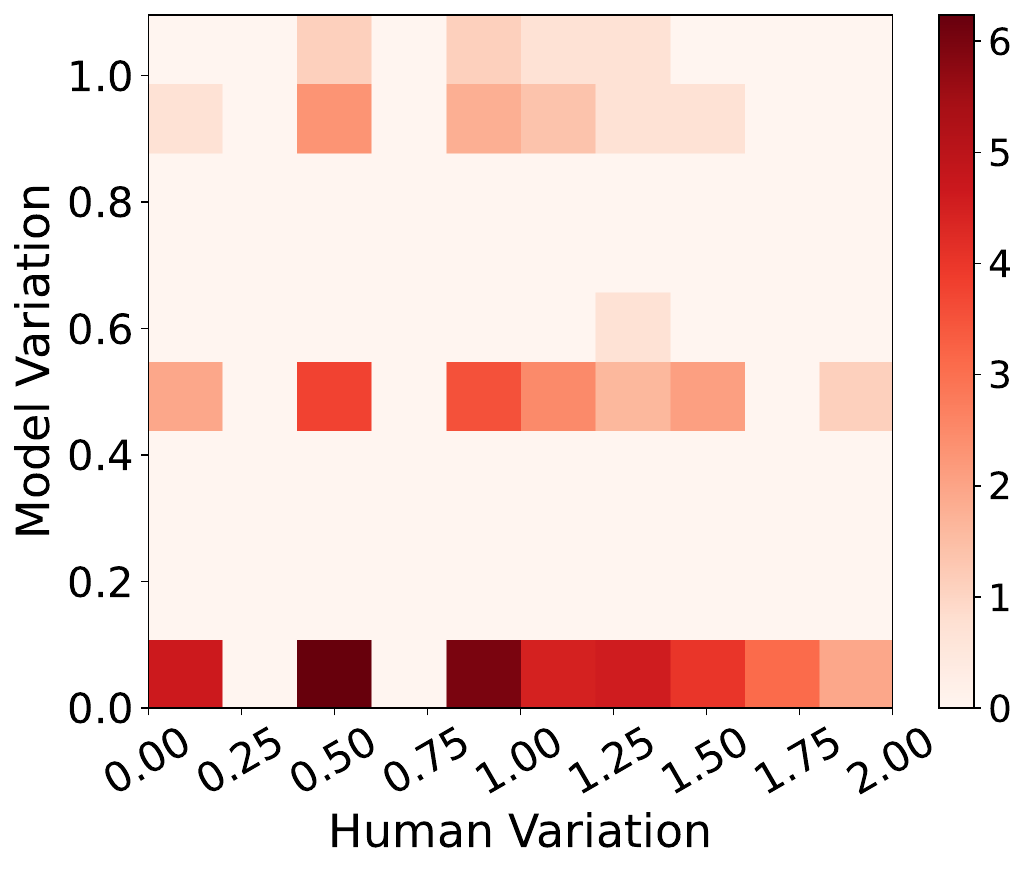}
          \includegraphics[width=0.5\linewidth,trim=1.2cm 0 0cm 0,clip]{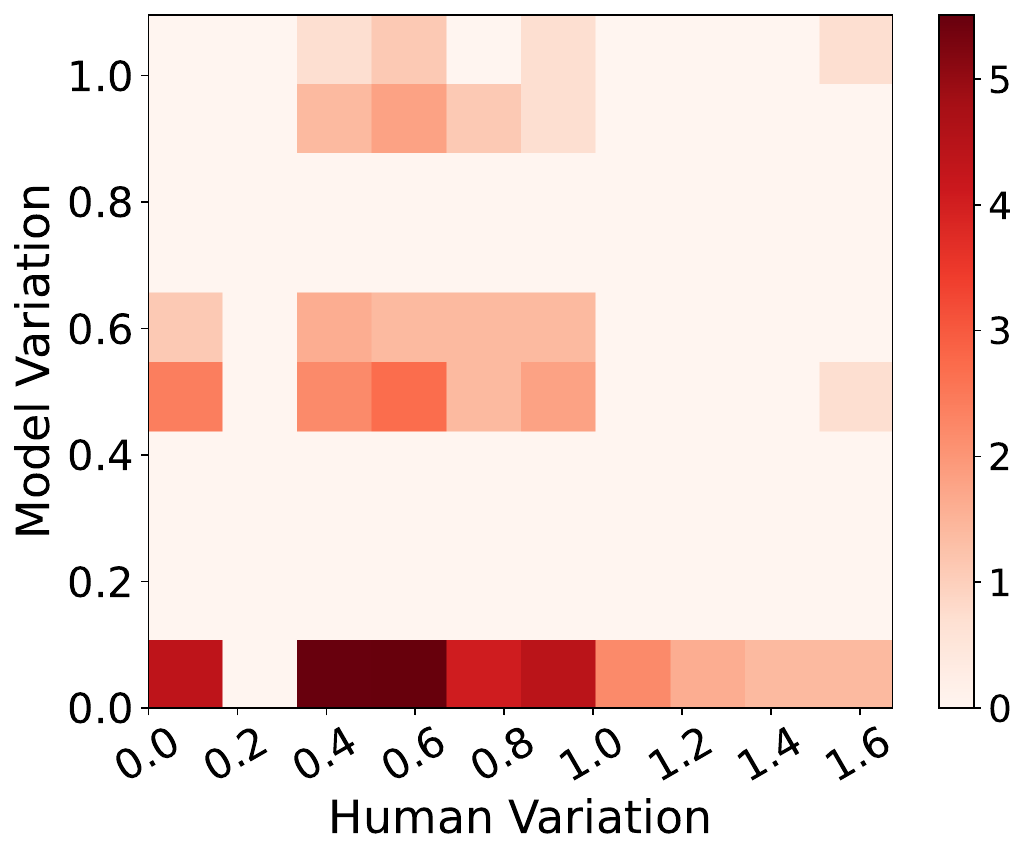}
          }
     {\caption{Human vs. ChatGPT Variation in annotations for the Sentiment (left) and Schem5 dataset (right). High variation is less common in ChatGPT's annotation group than in the human annotation group, suggesting that \textit{ChatGPT may have a more constrained response pattern or is better at mimicking a consensus among human annotations.} \label{fig:variation_comparison}}}

    \ffigbox{
    \includegraphics[width=0.44\linewidth,trim=0 0 4.7cm 0,clip]{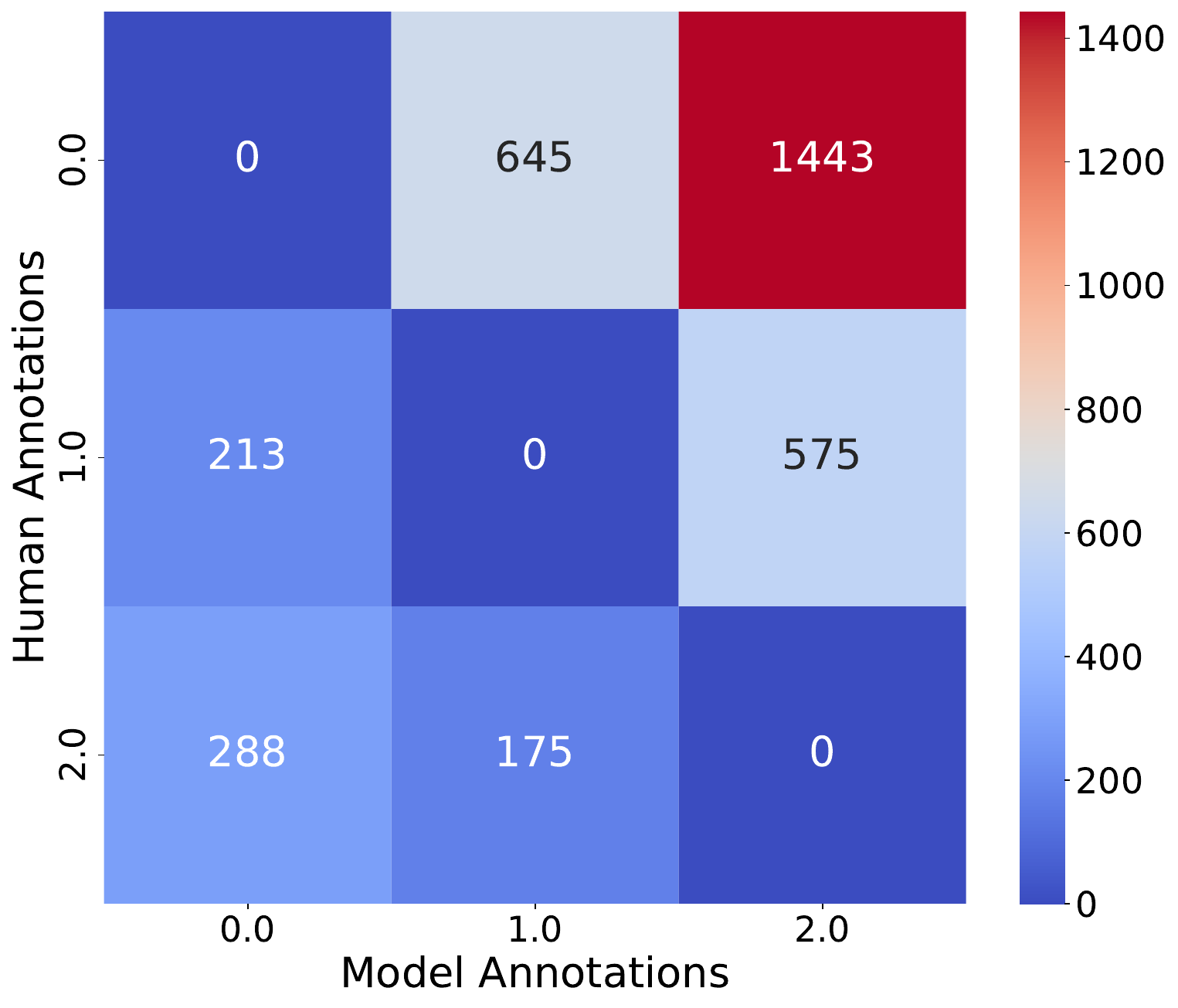}
    \includegraphics[width=0.52\linewidth,trim=0.95cm 0 0cm 0,clip]{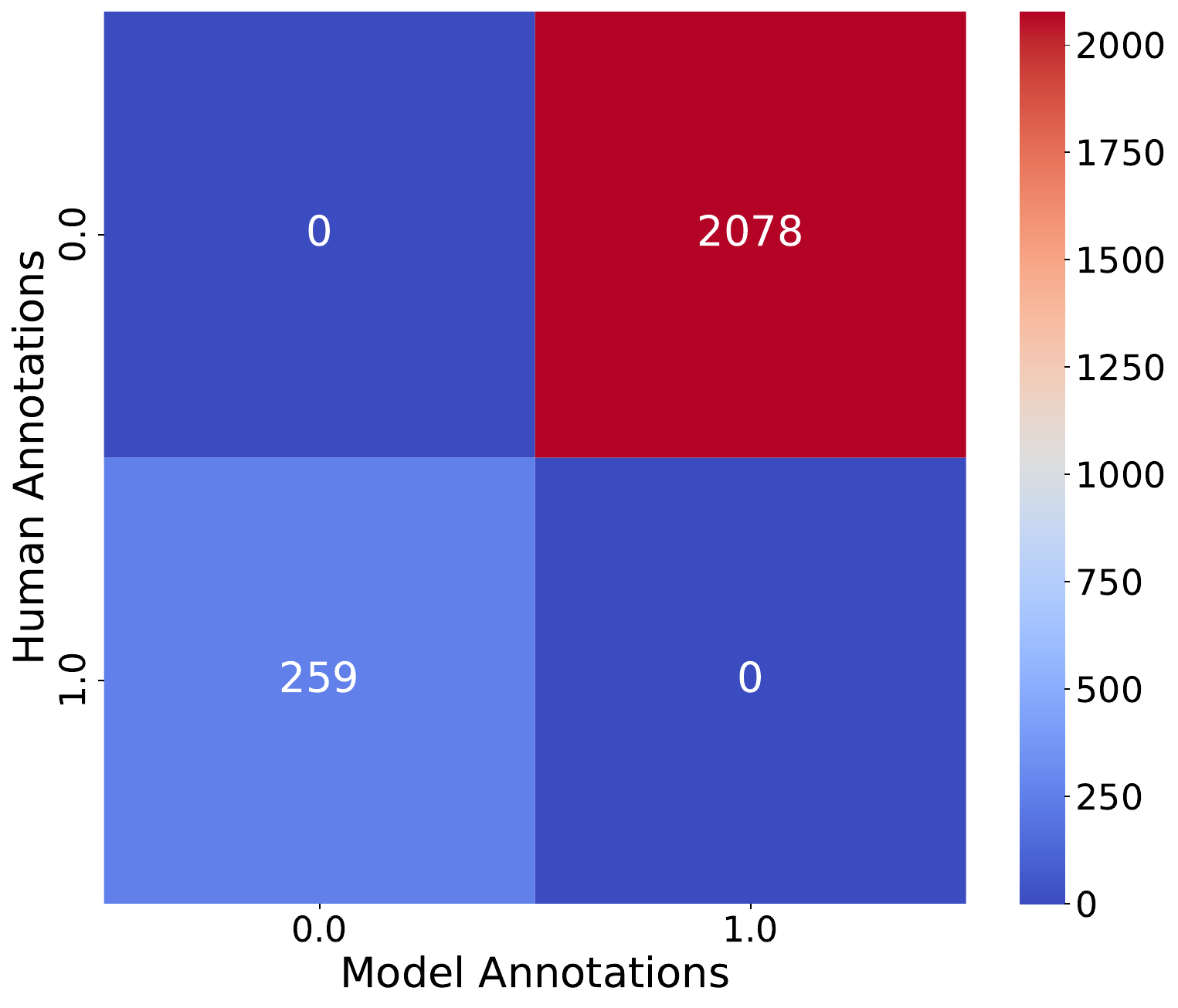}}
    {\caption{We observe \textit{label flipping}, indicating disagreement between machine and human task annotations for the SBIC (left) and GHC (right) datasets; \textit{notably, in SBIC, sentences labeled ``0'' by humans are predominantly flipped to ``2'' by the model.} The grid counts represent the frequency of these flips, offering insights into potential model biases and highlighting areas needing further model alignment with human judgment.\label{fig:heatmaps}}}
\end{floatrow}
\end{figure}

\begin{table}[b!] 
\footnotesize
\begin{tabularx}{\columnwidth}{>{\hsize=2\hsize}X>{\hsize=0.5\hsize\centering\arraybackslash}X>{\hsize=0.5\hsize\centering\arraybackslash}X}
\rowcolor{Gray}
\textit{Text} & \textit{Human Labels} & \textit{Model Labels} \\ \midrule
\includegraphics[scale=0.01]{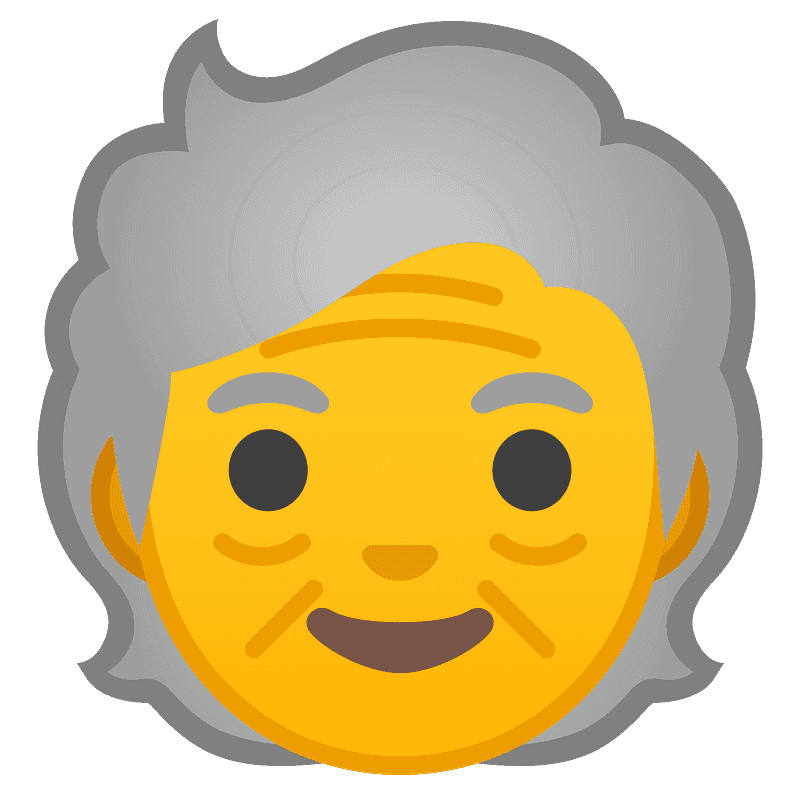} {@nounstudios this old dog needs to learn some new tricks} & [2, 3, 2, 3]  & [1, 1, 1, 1, 1] \\ 
\includegraphics[scale=0.01]{images/icons/older-person-emoji-clipart-md.png} {gosh...i'm 22 years old...an old lady! thanks for all the great birthday wishes! now it's bedtime!} & [1, 2, 4, 4]  &[1, 1, 1, 1, 1] \\ \midrule
\includegraphics[scale=0.04]{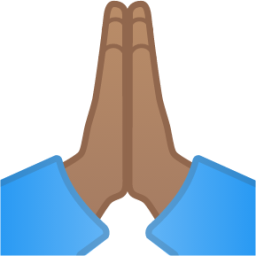} {I'm skeptical. Nearly every time this type of vandalism occurs, it is perpetrated by someone affiliated with the synagogue.} & [1, 0, 0]  & [0, 0, 0, 0, 0] \\ 
\includegraphics[scale=0.04]{images/icons/folded-hands.png} {The Pope will grovel at the feet of Muslims yet again. \#Genoa bells} & [1, 0, 0]  & [1, 1, 1, 1, 1] \\ \midrule
\includegraphics[scale=0.02]{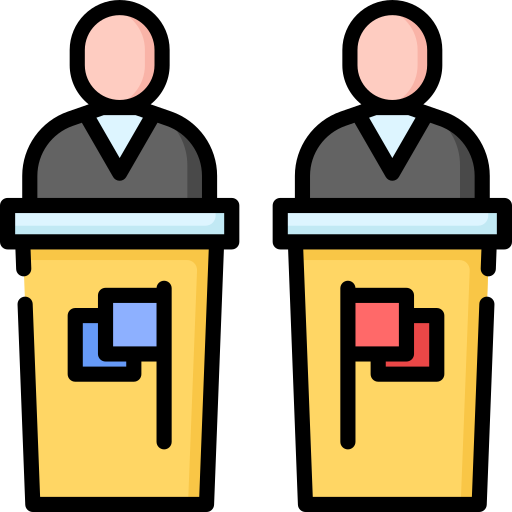} {If you're a right-winger, how could you not be excited at the prospect of a civil war split down ideological lines?} & [0, 1, 1]  & [1, 1, 1, 1, 1] 
\\ 
\includegraphics[scale=0.02]{images/icons/politics.png} {If we're going to talk about Russian sources, we should also talk about Jewish sources.} & [1, 0, 0]  & [0, 0, 0, 0, 0]\\ \midrule
\includegraphics[scale=0.03]{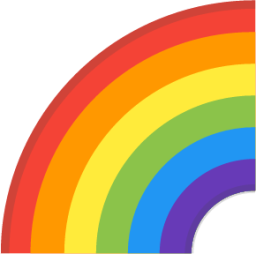} {It's fact that a male brain cannot always hear the high pitched voice of a women.. So bitch don't get mad..I really didn't hear you!} & [1, 2, 2]  & [2, 2, 2, 2, 2] \\
\includegraphics[scale=0.03]{images/icons/rainbow.png} {I just think life is hard enough without being gay on top of it.} & [0,1,1]  & [2, 2, 2, 2, 2] \\
\bottomrule
\end{tabularx}
\caption{Qualitative analysis of human and machine task labels unearths evidence of {age} (\includegraphics[scale=0.01]{images/icons/older-person-emoji-clipart-md.png}), {gender or sexuality} (\includegraphics[scale=0.04]{images/icons/rainbow.png}), {politics} (\includegraphics[scale=0.02]{images/icons/politics.png}) and {religion bias} (\includegraphics[scale=0.04]{images/icons/folded-hands.png}). Sentences that result in diverse human annotations do not necessarily produce a similar variety in ChatGPT's annotations.}
\label{table:qual_table}
\end{table}

\vspace{-2mm}

\subsubsection{Findings on Variation and Disagreement Analyses} 
We compared the variation in human annotations with the intra-model annotations of ChatGPT for each dataset, observing that ChatGPT displays less variability in responses to subjective tasks than humans (Figure \ref{fig:variation_comparison}), potentially indicating task oversimplification or high confidence. The heatmap's color intensity indicates the logarithmically transformed counts of occurrences for each combination of human and model variation. Overall, the model aligns with human consensus in cases of high agreement but lacks the variability seen in human responses during disagreements, suggesting either more consistent responses or a failure to capture the full spectrum of human perspectives. Other results are present in Appendix \ref{sec:app-task_label_first}

In our analysis shown in Figure \ref{fig:heatmaps}, we compared human annotations with machine annotations in the inter-model setup, identifying a phenomenon of \textit{label flipping} seen in datasets like SBIC and GHC, where the machine's annotations significantly differed from human annotations, particularly flipping from ``not offensive'' or ``not hate'' to ``very offensive'' or ``hate.'' This discrepancy highlights a notable disagreement in task annotations between humans and machines for subjective datasets, with further examples detailed in Appendix \ref{sec:app-task_label_first}.

\subsubsection{Findings on Minority Biases in Demographics} 
Table \ref{table:qual_table} shows some instances across all our datasets where there is significant human variation in labels but no model variation and where the minority label is not the same as the human minority label. We observe clear evidence of {age}, {gender}, {political}, and  {race/religion bias}. This could be because minorities are underrepresented in the model's training data.
\subsection{Second Order Experiment}
\label{sec:labels-second}

\begin{wrapfigure}[12]{r}{3.8in}
\vspace{-8mm}
    \centering
    \begin{subfigure}{0.49\textwidth}
         \centering
         \includegraphics[width = \linewidth]{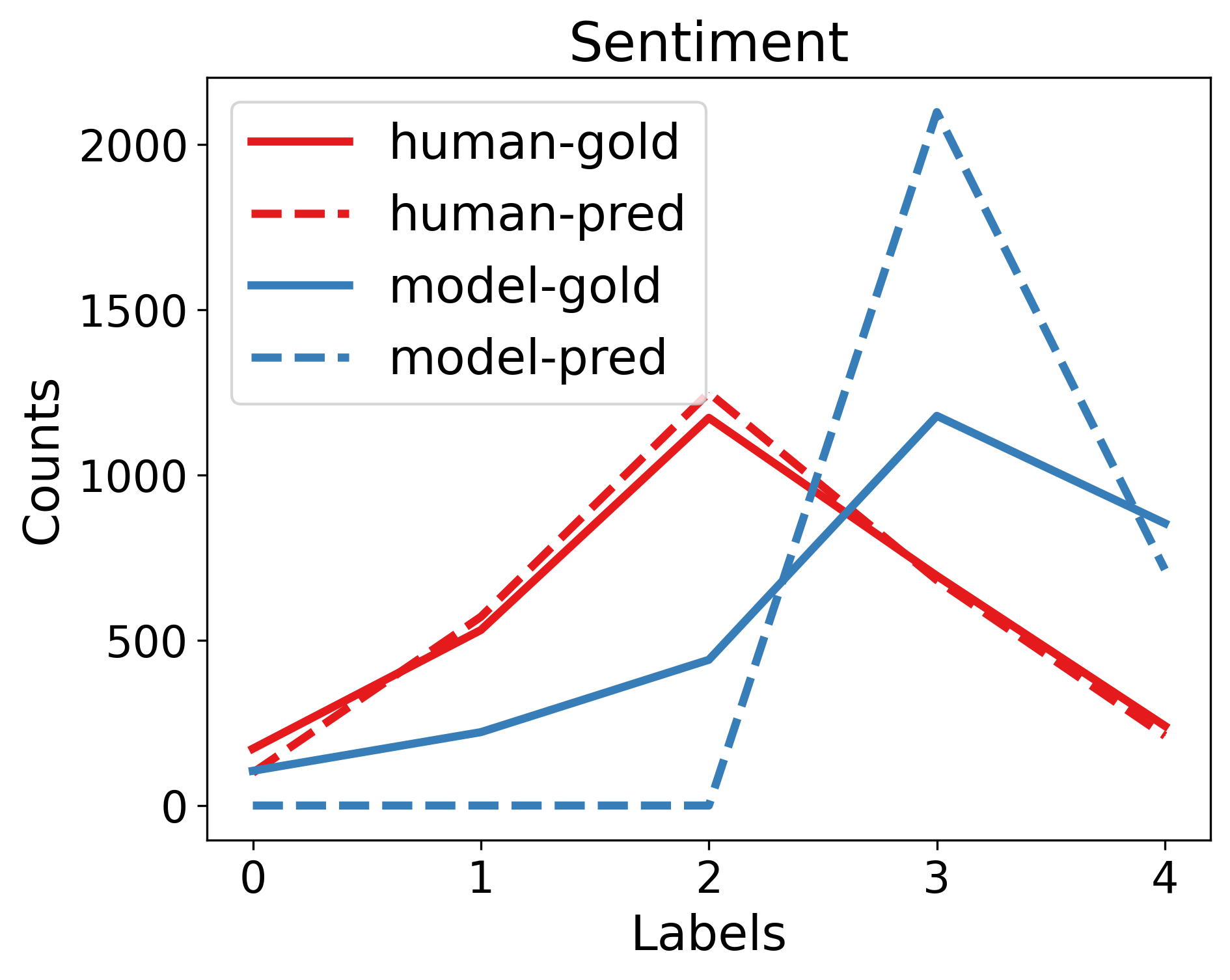}
     \end{subfigure}
         \begin{subfigure}{0.49\textwidth}
         \centering
         \includegraphics[width = \linewidth]{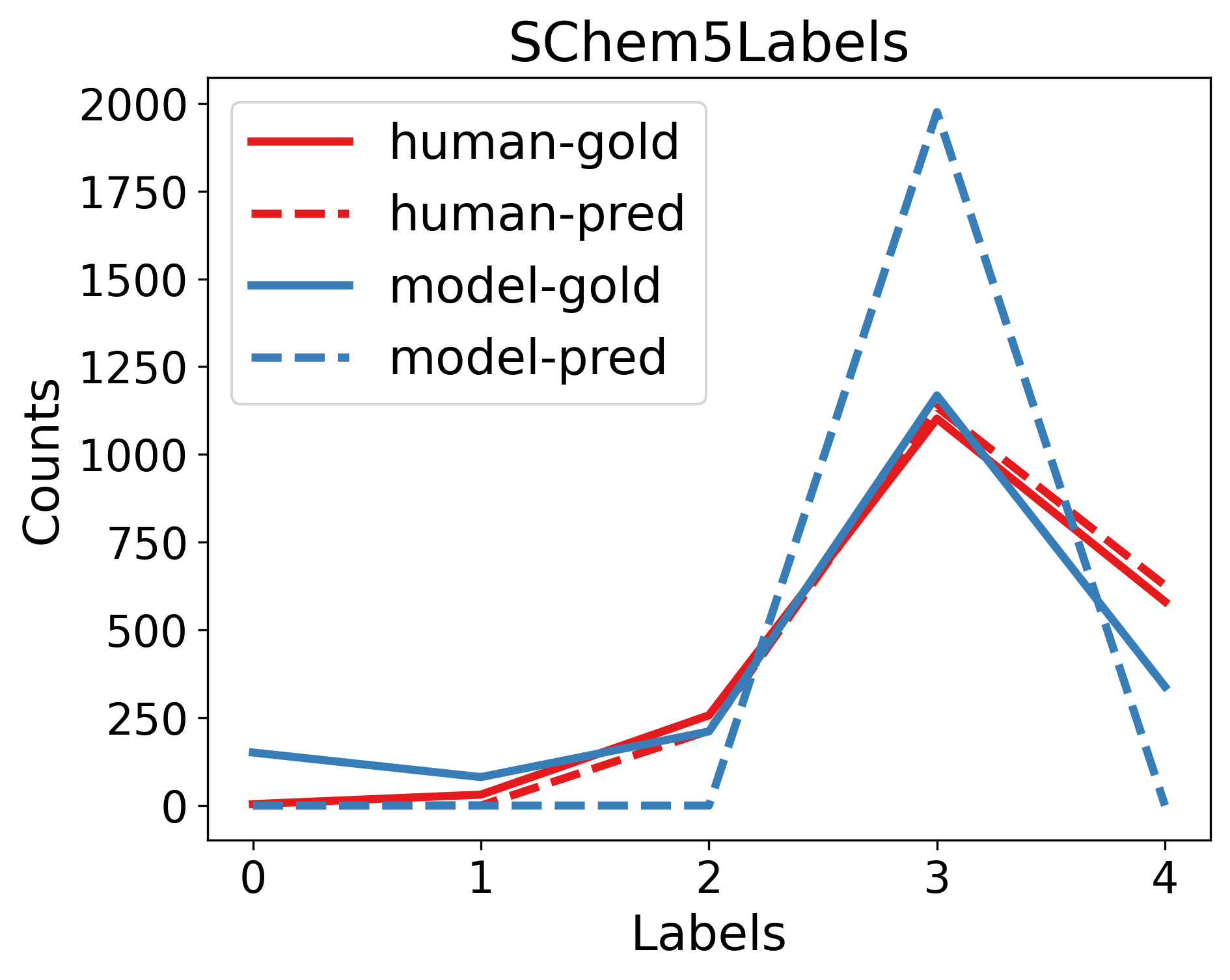}
     \end{subfigure}
     \vspace{-3mm}
    \caption{The label distribution graph for the Sentiment and SChem Dataset, before and after fine-tuning a RoBERTa classifier, indicates that \textit{pre-existing biases in label distribution}---such as imbalances in the original dataset (1st order)---\textit{tend to become more pronounced after the fine-tuning process (2nd order).}}
    \label{fig:second_order_effect}
\end{wrapfigure}
Our study investigates if training a RoBERTa classifier with LLM annotations skews the representation of minority opinions. We used a multi-headed RoBERTa-base classifier, each head corresponding to an annotator, and a custom loss function ($\mathcal{L} = \alpha\mathcal{L}_{CE} + (1-\alpha)\mathcal{L}_{W}$) combining cross-entropy ($\mathcal{L}_{CE}$) and Wasserstein distance ($\mathcal{L}_{W}$ ), the latter emphasizing the ordinal nature of data by penalizing more distant predictions. The data was split in a 90-10-10 ratio. 

\noindent \textbf{Findings}: Results in Figure \ref{fig:second_order_effect} show that LLMs like ChatGPT tend to under-represent minority opinions and over-represent majority views, potentially increasing societal disparities. This issue, linked to the ordering of labels in training data, underscores the risk of marginalizing minority perspectives\citet{prabhakaran2021releasing}. Additional results are in the Appendix \ref{sec:app-task_label_first}.


\subsection{Summary}
We summarize our major artifacts found in artificial task label data. 
\begin{itemize}[noitemsep,topsep=5pt,leftmargin=*]
    \item In tasks involving high subjectivity, LLMs such as ChatGPT show a strong inclination towards majority opinions, leading to a notable discrepancy in preserving minority opinions in LLM-generated task labels.
    \item ChatGPT exhibits less variability than humans in subjective tasks, often aligning with consensus but lacking in capturing diverse perspectives, suggesting possible oversimplification.
    \item There is a significant discrepancy in subjective dataset annotations between humans and machines, notably in ``label flipping'' as evidenced in datasets like SBIC and GHC.
    \item Qualitative analysis reveals instances across datasets where model labels lack the variation seen in human labels, showing biases related to age, gender, politics, and race/religion, possibly due to the underrepresentation of minorities in the model's training data.
\end{itemize}

\newpage
\section{Artifacts in Preferences}
\label{sec:preference}
We detail our experiments that indicate that LLM evaluations, particularly focusing on pair-wise preference, adhere more closely to rudimentary heuristics than humans evaluate. The results suggest that human preferences may be more complex and nuanced than those of LLMs. We also find misalignment in human preferences for models trained on LLM preferences.

\paragraph{Related Work} Preference data are used to align LLMs with human reasoning \cite{bai2022training, ouyang2022training}, yet collecting human preference faces subjectivity and cost challenges. Synthetic preferences generated by LLMs are an alternative that has shown promise in complex tasks \cite{kocmi-federmann-2023-large, liu2023geval, gao2023humanlike, zhuo2023large, bai2023benchmarking} and aligned with expert human preferences \cite{zheng2023judging}. At the same time, there is a notable misalignment between human rationales and LLM judgments in question-answering tasks, attributed to biases in LLMs when evaluating text quality. For example, studies have identified various biases in LLMs like GPT-4, including position, verbosity, self-enhancement, egocentric, and other types of cognitive biases \cite{zheng2023judging, wang2023large, koo2023benchmarking}. 

\vspace{3mm}
\noindent
\begin{minipage}[t]{\linewidth} 
\begin{tcolorbox}[colback=gray!15]
We investigate whether LLM preferences are more closely aligned to basic heuristics (called \textit{locality biases}, such as specific lexicons and entailment) than human preferences are.

\end{tcolorbox} 
\end{minipage}

\paragraph{Data}
We first use \textsc{P2C} \cite{kim2023p2c}, a preference dataset for textual sentiment on 5,000 sentence pairs with human and GPT-3 preference responses.
In addition, we analyze \textsc{CoBBLEr} \cite{koo2023benchmarking}, a dataset containing 50 closed QA pairs with human and 15 different LLMs' preference responses. \\
\textbf{Justification} We chose these datasets to include the paired human and LLM preferences in evaluating text quality.

\subsection{First Order Experiments} 
\label{sec:preference_first}


\begin{figure}[ht!]
\vspace{-5mm}
    \centering
    \begin{subfigure}{0.485\textwidth}
        \caption{Lexicon biases in \textsc{P2C}}
         \centering
         \includegraphics[width=\textwidth]{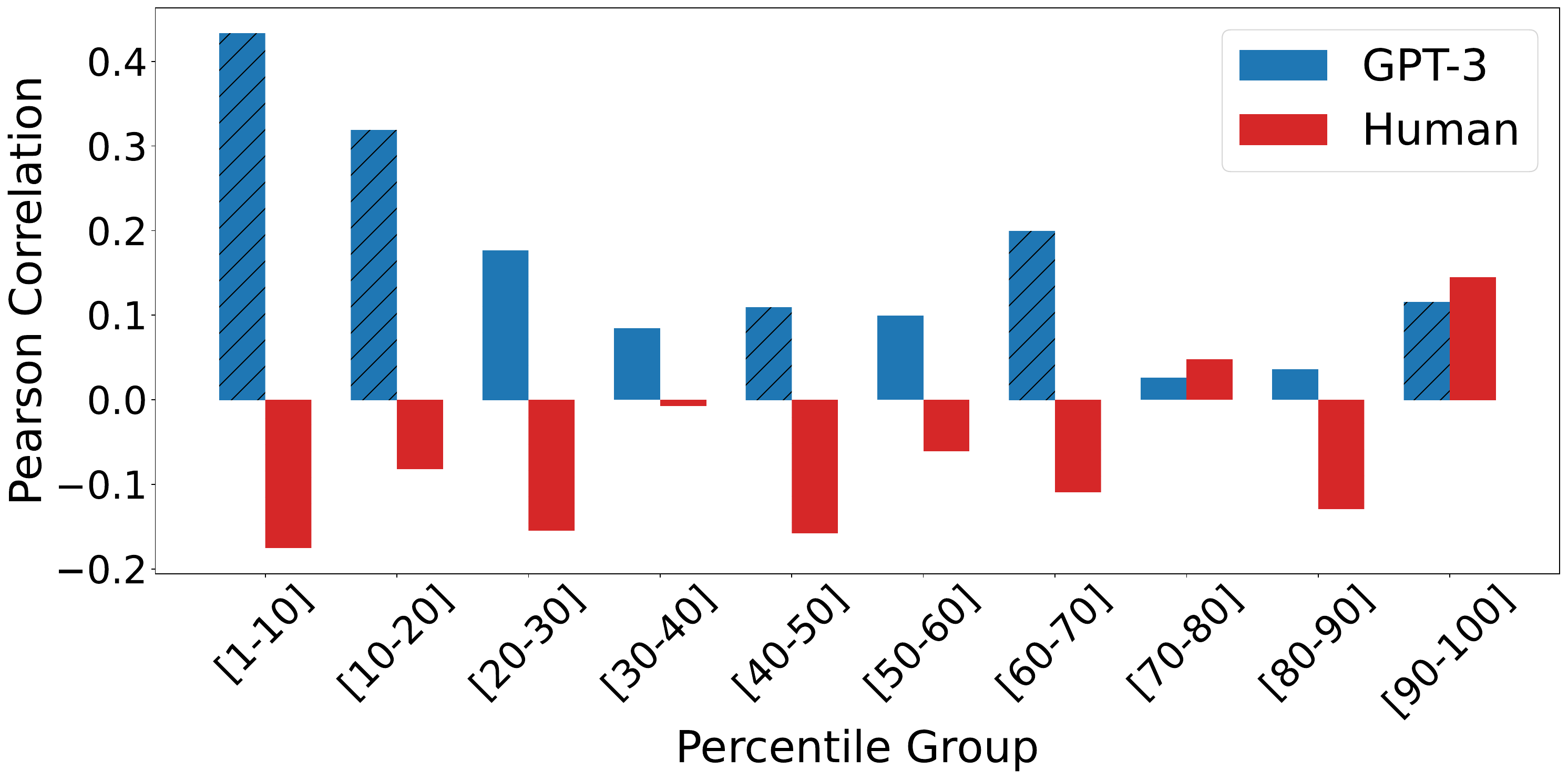} 
         \label{fig:pref:corr-p2c}
     \end{subfigure}
     \quad
         \begin{subfigure}{0.485\textwidth}
         \caption{Entailment biases in \textsc{Cobbler}}
         \centering
         \includegraphics[width=\textwidth]{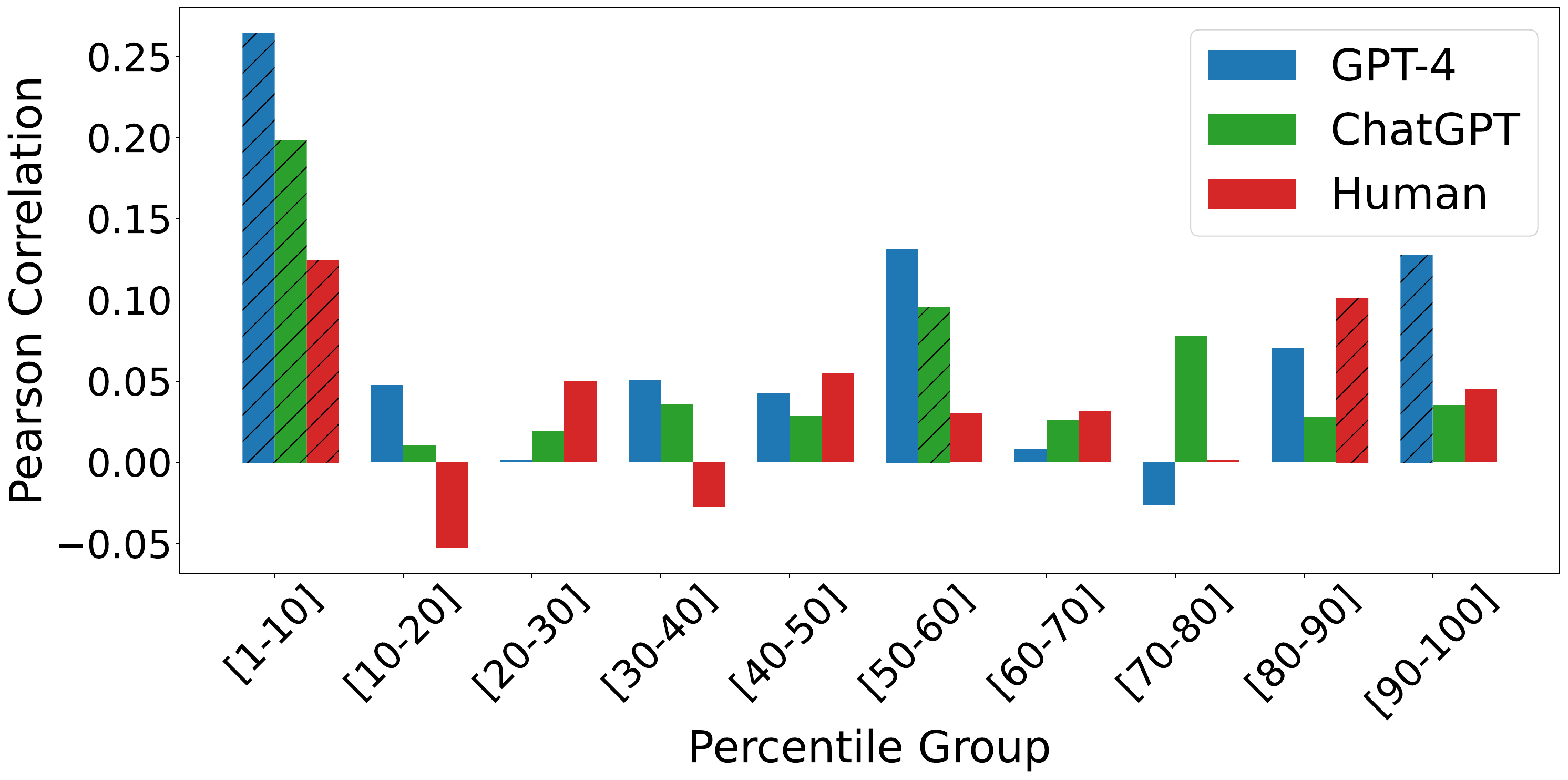}
         \label{fig:pref:corr-cobbler}
     \end{subfigure}
     \vspace{-12mm}
    \caption{Pearson correlation (y-axis) between locality preference and each of GPT-3, GPT-4, ChatGPT, and human preferences across 10 different percentile groups (x-axis). Bars with statistically significant correlation ($p < 0.05$) are hatched. In both plots, we observe that the [1-10] percentile (i.e., having the most noticeably different linguistic cues between two sentences) shows a significantly higher correlation between locality preference and LLM preferences than humans. Also, as the percentile group increases, the degree of correlation between LLM and locality preferences decreases much more than human counterparts. These results point out that \textit{LLM preferences are likely to be more sensitive to local cues of sentences} (i.e., lexicons, entailment) rather than their nuanced representation.}
    \label{fig:pref:corr}
\end{figure}

\subsubsection{Locality Biases to Sentiment Lexicons}
We use a lexicon-based heuristic approach in our study on sentiment preference in \textsc{P2C}. This involves analyzing stylistic elements through sentiment annotations from \citet{hayati2021does} and calculating a lexical score for each sentence pair. We identify key lexicons linked to specific emotions: for positive sentences, we look for terms related to \textit{joy}, \textit{politeness}, and \textit{positive sentiment}, while for negative ones, we focus on lexicons associated with \textit{anger}, \textit{disgust}, \textit{fear}, \textit{sadness}, \textit{offensiveness}, \textit{rudeness}, and \textit{negative sentiment}. The highest-scoring lexicon in each sentence indicates its most expressive element for the intended sentiment. We then compare the scores within each sentence pair to determine which sentence more strongly represents the sentiment, designating it as the "winner." This process is repeated across various emotional styles within the same sentiment category, and the sentence that consistently scores highest is considered the preferred one, in what we call ``lexicon-based preference.'' We calculate Pearson's $r$ correlation to compare the alignment between lexicon-based preferences and humans' original preferences and LLMs' original preferences. Additionally, we examine how LLM and human preferences differ when evaluated against lexicon-based preferences. This analysis identifies if LLMs focus more on prominent, localized sentence features affecting their preferences than humans. To deepen our understanding, we organize the dataset by the disparity in lexicon/entailment scores between sentence pairs, dividing it into percentile groups of 10\% each. For instance, the [1-10]\% percentile group includes sentences with lexicon perception/entailment differences in the top 10\% of the dataset, while the [90-100]\% group includes those in the bottom 10\%. This method allows us to observe the correlation within these specific segments.

\noindent \textbf{Findings:} Our results show that while GPT-3 preferences positively correlate with the simple lexical preference, human preferences do not (see  Figure~\ref{fig:pref:p2c-lexicon-heatmap} in Appendix \ref{sec:appendix:pref}). This suggests that GPT-3 takes a more surface-level approach to sentiment preference than humans do. 

We also examine how the correlations between preferences and the computed lexical-based preferences vary with different magnitudes of lexical score discrepancy in the sentence pairs, as depicted in Figure~\ref{fig:pref:corr-p2c}. To do this, we divide the paired sentences into ten percentiles based on the magnitude of their lexical score discrepancy. For example, the [1-10] group represents the top 10\% of pairs of sentences, showing the greatest difference in the pair's lexical scores.

\noindent \textbf{Findings:} The top percentile group, with the greatest difference in the pair's lexical scores, shows lexical-based preferences are strongly positively correlated with GPT-3's preferences, which is not the case for human preferences. In contrast, with the smallest difference in the pair's lexical scores, the bottom percentile group shows a negligible difference between GPT-3 and humans. This suggests that when lexical differences between sentences are very strong, GPT-3 preference decisions are noticeably influenced by the local information of sentences.

\begin{table}[ht!]
\footnotesize
\begin{tabularx}{\columnwidth}{p{4cm}p{5cm}ccc}
\rowcolor{Gray} \textit{Sentence 1}  & \textit{Sentence 2} & \textit{Sentiment Label} & \textit{Human Pref.} & \textit{GPT-3 Pref.} \\ \midrule
This place is \colorbox{cyan}{not} my \#1 pick. & We will \colorbox{blue}{\textcolor{white}{definitely}} be going back when we are in the area and in the mood for being treated like \colorbox{blue}{\textcolor{white}{crap}}. & Negative &  1 &  2 \\ \midrule
The food was exactly as it was advertised to be \colorbox{pink}{on} the menu. & BREWED TEA is making some \colorbox{magenta}{\textcolor{white}{terrific}} green tea matcha! & Positive & 1 & 2\\ \midrule 
That place was \colorbox{magenta}{\textcolor{white}{good}}.  I'll tell my enemy to go there. & Thats \colorbox{magenta}{\textcolor{white}{good}}, I guess he's a \colorbox{magenta}{\textcolor{white}{good}} doctor if he only see him that much. & Positive & 1 & Tie \\ 
\bottomrule
\end{tabularx}
\caption{Examples of paired sentences with human and GPT-3 preferences from \textsc{p2c} dataset, highlighted by relevant lexicons (e.g., \textcolor{blue}{blue} for \textit{negative}, or \textcolor{magenta}{pink} for \textit{positive} sentiment). Lexicons highlighted with paler colors have relatively lower word importance to the corresponding sentiment label.}
\label{table:pref:p2c-examples}
\end{table}
Our qualitative analyses (Table \ref{table:pref:p2c-examples}) examine some instances from \textsc{p2c} dataset where there is a misalignment between human and GPT-3 preferences. 

\noindent \textbf{Findings:} We observe GPT-3's locality biases to sentiment lexicons in its preferences compared to humans, where it tends to prefer the sentence that possesses lexicons with stronger importance to predict the sentiment of the two sentences in a pair. 
\subsubsection{Locality Biases to Textual Entailment} 

In the \textsc{Cobbler} study, we use a RoBERTa \cite{liu2019roberta} model fine-tuned on MNLI \cite{williams18inf} to determine the entailment of each question-answer pair. We assess which sentence in a pair aligns better with its reference based on the highest entailment score. To do so, we run RoBERTa\footnote{we used ``roberta-large-mnli'' (\url{https://huggingface.co/roberta-large-mnli}).} \cite{liu2019roberta}  on each sentence pair, in which we extract the maximum entailment score between each individual sentence of the pairs and its corresponding reference sentence. Similarly, we consider the sentence that wins in the entailment comparison as the preferred sentence, termed ``entailment-based preference.''. We apply the same Pearson's $r$ to analyze the distribution differences between LLM and human preferences in relation to entailment-based preferences for every sentence pair in \textsc{CoBBLEr}.

\noindent \textbf{Findings:} Machine preferences (GPT-4, ChatGPT) show a marginal correlation with sentences having higher entailment scores than human preferences, as indicated by Pearson's r values (0.07, 0.06, and 0.03, respectively). Notably, GPT-4 and ChatGPT preferences align more with entailment-based choices (Figure \ref{fig:pref:corr-cobbler} and Table \ref{table:pref:cobbler-examples}), especially in sentence pairs with the greatest differences in entailment scores. This trend suggests LLMs are influenced more by prominent, entailment-related aspects of sentences than a comprehensive understanding of the text.

\begin{table}[ht!]
\small
\centering
\begin{tabularx}{\textwidth}{@{}X@{}}
\toprule
    {\begin{tabularx}{\textwidth}{X}
    \textbf{Instruction: } Did Helen Keller ever read a novel by J. K. Rowling?\newline 
    \textbf{Reference:} No. Helen Keller died in 1968. J. K. Rowling's first novel was published in 1997. \newline \textbf{Model A: }  No, Helen Keller died in 1968, long before J. K. Rowling published her first novel in 1997. \textbf{(\textcolor{red}{0.92})}\newline 
    \textbf{Model B: } No, Helen Keller would not have been able to read a novel by J.K. Rowling as she was both deaf and blind. \textbf{(0.00)} \\
    \rowcolor{Gray}
    \textbf{Human Preference: } \textit{Model B}   /  \textbf{GPT-4 Preference: } \textit{Model A} / \textbf{ChatGPT Preference: } \textit{Model A}
    \end{tabularx}}\\ 
    \midrule 
    {\begin{tabularx}{\textwidth}{X}
    \textbf{Instruction: } Did occupants of Vellore Fort need to defend themselves from Grizzly Bears?\newline 
    \textbf{Reference:} No. The Vellore Fort was a 16th century stronghold in India. Grizzly Bears are native to the North American continent. \newline 
    \textbf{Model A: }   No, Grizzly Bears do not exist in India. \textbf{(0.00)}\newline 
    \textbf{Model B: }  No, the occupants of Vellore Fort did not need to defend themselves from Grizzly Bears. Grizzly Bears are found in the Americas and would not have come into contact with the occupants of the Vellore Fort in India. \textbf{(\textcolor{red}{0.31})}\\
    \rowcolor{Gray}
    \textbf{Human Preference: } \textit{Model A}   /  \textbf{GPT-4 Preference: } \textit{Model B} / \textbf{ChatGPT Preference: } \textit{Model B}
    \end{tabularx}}\\ 
    \midrule
    {\begin{tabularx}{\textwidth}{X}
    \textbf{Instruction: } Did people in Korea under Japanese Rule watch a lot of Iron Chef? \newline 
    \textbf{Reference:} No. The first televisions were sold in 1946. Iron Chef started airing in 1993. \newline 
    \textbf{Model A: }  I'm sorry, but as an AI language model, I do not have access to historical data regarding the viewing habits of people in Korea under Japanese rule. \textbf{(\textcolor{red}{0.08})}\newline 
    \textbf{Model B: } There is not any available information that suggests whether people in Korea under Japanese Rule watched Iron Chef or not. \textbf{(\textcolor{red}{0.08})} \\
    \rowcolor{Gray}
    \textbf{Human Preference: } \textit{Model B}   /  \textbf{GPT-4 Preference: } \textit{Model B} / \textbf{ChatGPT Preference: } \textit{Model B}
    \end{tabularx}}\\
\bottomrule
\end{tabularx}
\caption{Examples of paired model outputs with human and GPT preferences from \textsc{cobbler} dataset, given instruction and reference sentences. Each model output is provided with the maximum entailment score, where the output with the higher score is bolded in \textcolor{red}{red}.}
\label{table:pref:cobbler-examples}
\end{table}

\begin{wraptable}{r}{8cm}
\footnotesize
\vspace{-4mm}
\centering
\resizebox{\columnwidth}{!}{%
\begin{tabular}{c|ccc}
\rowcolor{Gray}
\textit{Training}  & \textit{Validation} & \textsc{P2C} & \textsc{CoBBLEr} \\ \midrule
\multirow{2}{*}{Human} & \textbf{Human} &  \textbf{0.49} / \textbf{1.1}   &  \textbf{0.80} / \textbf{0.48}       \\
                       & LLM &  0.33 / 1.4   &   0.67 / 0.92      \\ \midrule
\multirow{2}{*}{LLM}   & Human &   0.34 / 2.0 & 0.69 / 0.76    \\
                       & \textbf{LLM} &   \textbf{0.66} / \textbf{0.8} & \textbf{0.75} / \textbf{0.64}  \\ \bottomrule
\end{tabular}%
}
\caption{Validation accuracy ($\uparrow$) and loss ($\downarrow$) for RoBERTa models trained on human and LLM preferences, respectively. e.g., the model trained on human preference shows a validation accuracy and loss value of 0.33 and 1.4 on the LLM preference validation set. The best performing test scores for each model are \textbf{bolded}. } 
\label{table:pref:2ndorder}
\end{wraptable}

\subsection{Second Order Experiments}
\label{sec:preference-second}


We train RoBERTa \cite{liu2019roberta} on each of the human or machine evaluations from two different preference datasets, as a reward model. We first train four different reward models from each dataset and then run on machine and human preferences. For each training session, we run RoBERTa-base models for 5 epochs with AdamW optimizer of learning rate $1e-4$. To get representations of the training distribution, we graph the logit versus probability of each class on test sets during inference, and we compare them between in-distribution sets (i.e., \textsc{Cobbler}-human vs \textsc{Cobbler}-machine, \textsc{p2c}-human vs \textsc{p2c}-machine). 

\noindent \textbf{Findings:} Our results (Table~\ref{table:pref:2ndorder}) show that the models trained on machine preferences, compared to those trained on human preferences, have about $10\%$ lower accuracy on the human test set for both datasets. This indicates that \textit{models trained with one preference cannot capture the other counterpart, showing a discrepancy between human and LLM preferences in their distributions} (Figure \ref{fig:pref:second_dist}; see Figures \ref{fig:pref:cobbler-second} and \ref{fig:pref:p2c-second} in Appendix \ref{sec:appendix:pref} for full details).  Further, a machine-trained model on a machine test set achieves the highest accuracy overall, indicating that \textit{machine preferences may be easier for models to learn than human preferences}. This may be due to machine preferences correlating more strongly with signals such as lexical features and entailment, as discussed above. 

\vspace{1mm}
\begin{figure}[ht!]
    \centering
    \begin{subfigure}{0.245\textwidth}
        \caption{\textsc{A<B} on \textsc{P2C}}
         \centering
        \includegraphics[width=\textwidth]{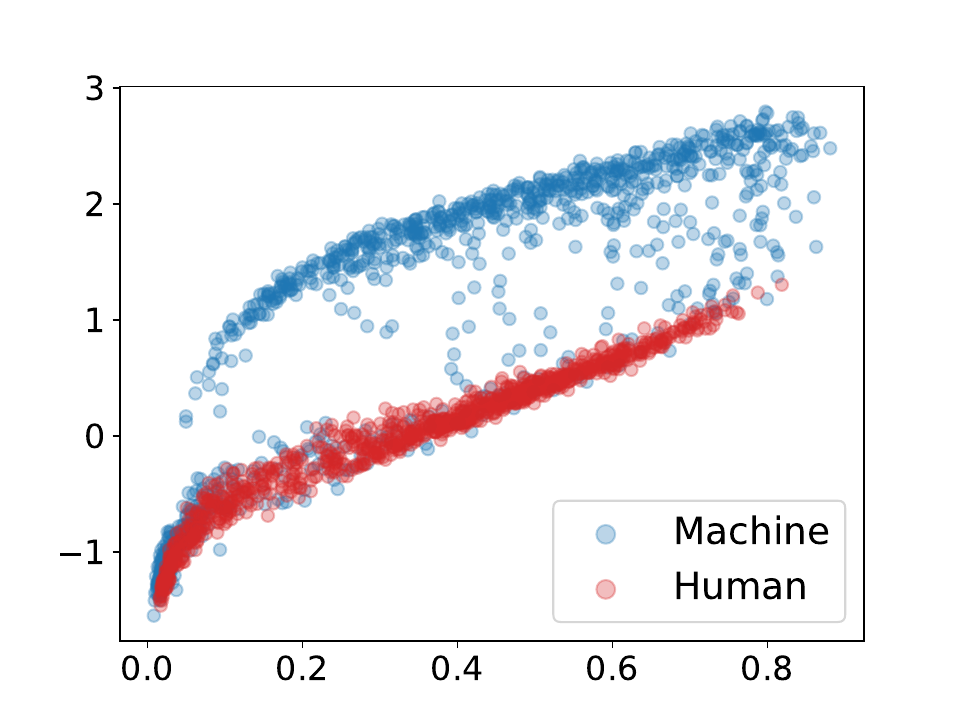}
     \end{subfigure}
     \begin{subfigure}[b]{0.245\textwidth}
     \caption{\textsc{A=B} on \textsc{P2C}}
     \centering
     \includegraphics[width=\textwidth]{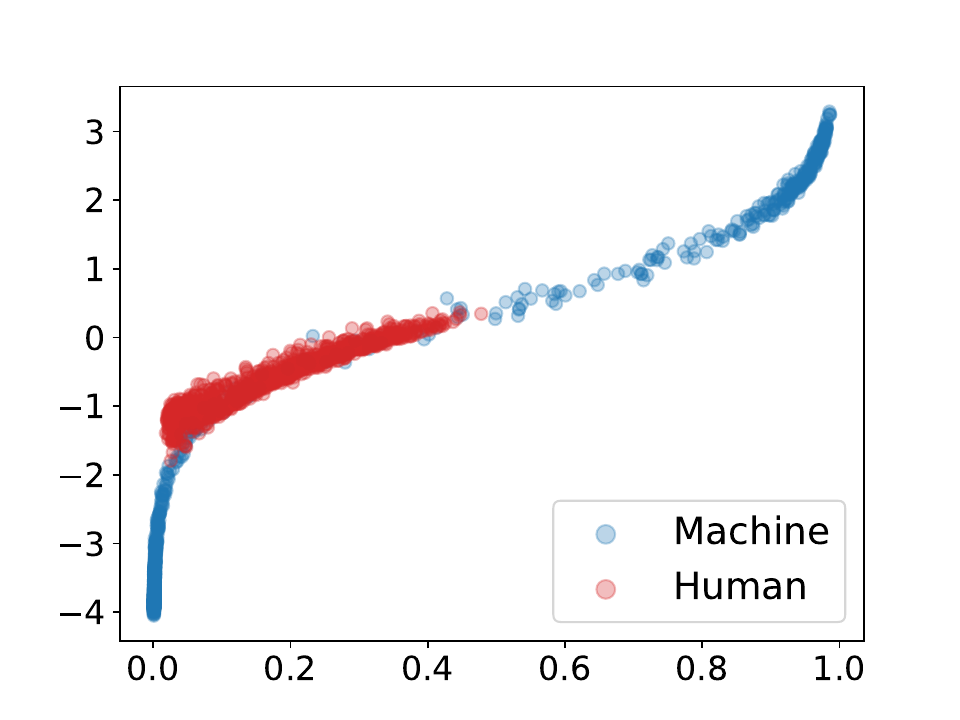}
     \end{subfigure}
   \begin{subfigure}{0.245\textwidth}
        \caption{\textsc{A<B} on \textsc{Cobbler}}
         \centering
         \includegraphics[width=\textwidth]{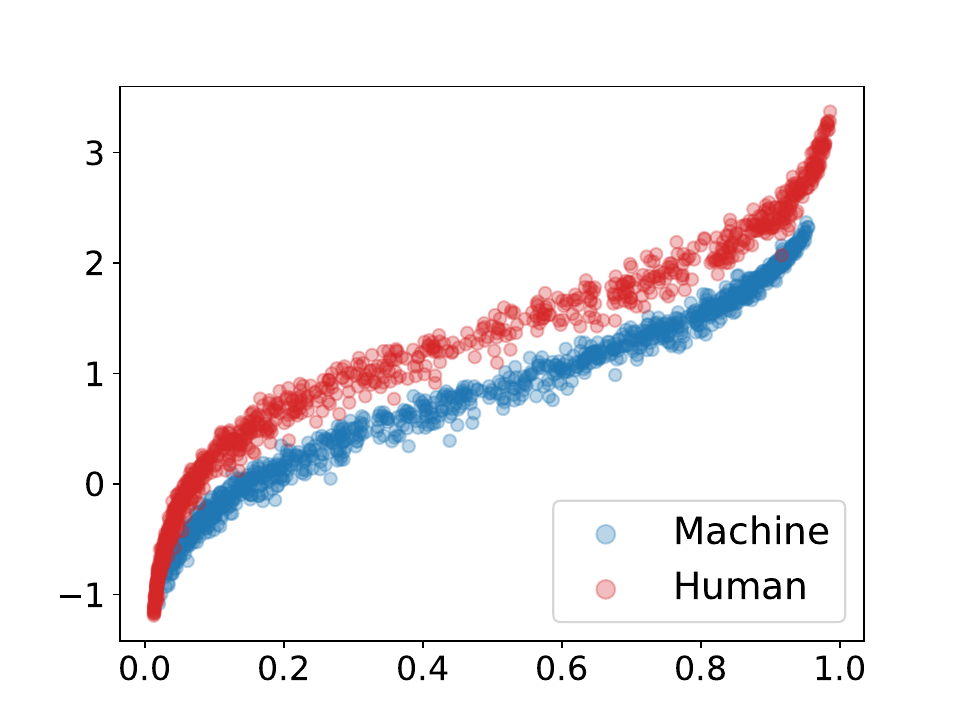}
     \end{subfigure}
     \begin{subfigure}[b]{0.245\textwidth}
     \caption{\textsc{A=B} on \textsc{Cobbler}}
     \centering
     \includegraphics[width=\textwidth]{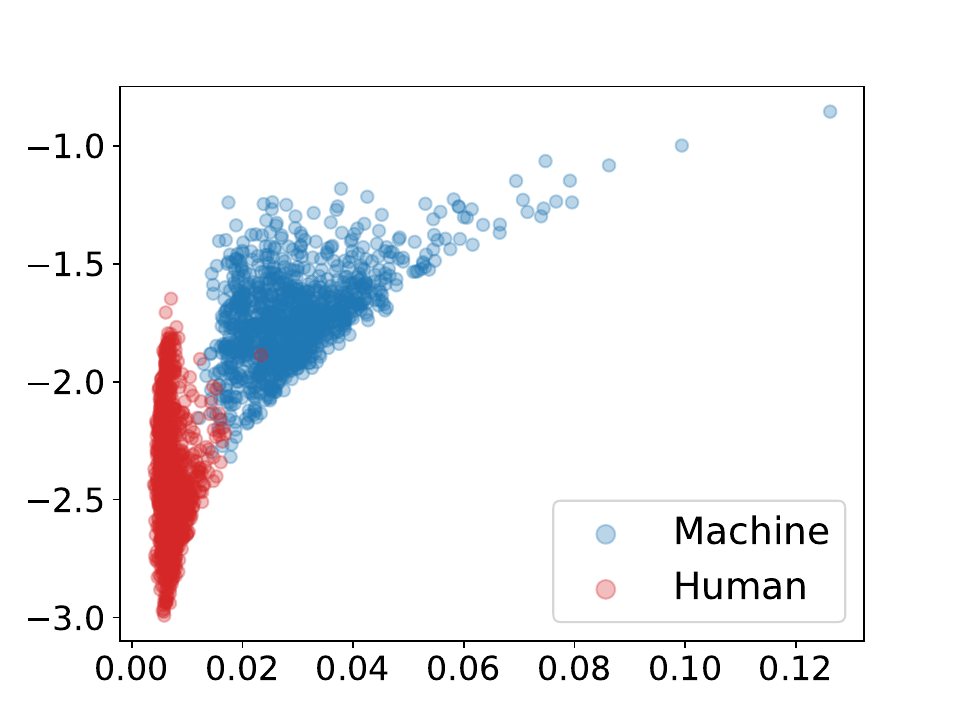}
     \end{subfigure}
    \caption{The logodds vs. class probability curve from the model trained on \textcolor{red}{human} or \textcolor{blue}{machine} preferences. We can observe discrepancies in the distributions between human and LLM preferences. 
    }
    \label{fig:pref:second_dist}
\end{figure}

\subsection{Summary}

We summarize our major artifacts found in artificial preference data. 
\begin{itemize}[noitemsep,topsep=5pt,leftmargin=*]

    \item LLM preferences are likely to be biased on local cues of sentences (i.e., sentiment lexicons, entailment) rather than the nuanced and comprehensive representation of them.
    \item Due to its simple heuristics, LLM preferences can be more easily captured than human counterparts and thus show a disparity with human preferences. 
    
\end{itemize}


\newpage
\section{Artifacts in Instructions}
\label{sec:instructions}
We discuss synthetic instruction prompts and their propensity for including errors rather than admitting uncertainty or hedging with non-committal answers. We also provide a novel annotation of errors across synthetic dataset. Our experiments illustrate the negative effects of these errors on the training outcomes for instruction-tuned models.
\paragraph{Related Work} Instruction prompts are used to refine the performance of LLMs, particularly for understanding and executing complex instructions. The purpose is to make LLMs more adept at handling diverse and unpredictable user requests. However, creating datasets of instructions is challenging, because it is difficult to construct a comprehensive set of complex and diverse instructions. For this reason, multiple synthetic instruction prompt resources have been created \cite{alpaca, honovich2022unnatural, wang-etal-2022-super, peng2023instruction}. 

These synthetic datasets often mirror the styles and outputs of their parent LLM, lacking comprehensive documentation about what task types are included. In contrast, human-crafted datasets \cite{khashabi2020unifiedqa, muennighoff2022crosslingual,lai2023chatgpt, sanh2021multitask, mishra2021natural, supernaturalinstructions, DatabricksBlog2023DollyV2} show a wider range of distinct task types. This limitation is further exacerbated by the LLMs' focus on superficial patterns and output formats \citep{kung-peng-2023-models}, which can lead to a narrow scope of learning and adaptability in handling diverse real-world scenarios. Additionally, the high error rates in instructions generated by LLMs raise concerns about the reliability and accuracy of the models trained on these datasets (\citet{honovich2022unnatural} and \citet{wang2023selfinstruct} both report error rates close to 50\% for their respective instruction datasets).

\vspace{5mm}
\noindent
\begin{minipage}[t]{\linewidth} 
\begin{tcolorbox}[colback=gray!15]
We aim to further the understanding of \textit{task diversity and errors in synthetically created instruction datasets}, as well as how these affect instruction-tuned model performance. 
\end{tcolorbox} 
\end{minipage}

\paragraph{Data}
We examine four synthetic instruction datasets, including \underline{\textsc{Unnatural Instructions}} \cite{honovich2022unnatural}, \underline{\textsc{Self-Instruct}} \cite{wang2023selfinstruct}, \underline{\textsc{Alpaca-Cleaned}} \cite{alpaca}, and \underline{\textsc{GPT-4-LLM}} \cite{peng2023instruction}.  We also rely on four human-written instruction datasets for comparison and evaluation, including \textsc{Flan 2021} \cite{weifinetuned}, \textsc{Dolly} \cite{DatabricksBlog2023DollyV2}, \textsc{Supernatural Instructions} \cite{supernaturalinstructions}, and \textsc{Instructions in the Wild} \cite{instructionwild}.

\subsection{First Order Experiments}
\label{sec:instructions_first}
\subsubsection{Error Type Distributions}
To understand some of the characteristics of synthetic instruction datasets, we annotated 800 samples (a random sampling of 200 from each of the four datasets). We annotate what kind of error the instruction contains, if any, as well as whether the instruction is \textit{open} or \textit{closed}. Our \textsc{Closed-QA} samples are defined as tasks with specific, objective answers, using given input or specific criteria. On the other hand, \textsc{Open-QA} samples are defined as more exploratory, subjective, or creative tasks. Our error labels, \textsc{Incomprehensible Instruction}, \textsc{Inconsistent Input}, and \textsc{Incorrect Output} follow  \citet{honovich2022unnatural} and \citet{wang2023selfinstruct}. We also add an \textsc{Inconsistent Output} label to describe better samples where the output is not only incorrect but inconsistent with the instruction. See Table \ref{tab:instruction-error-examples} for an example of each error type.

\begin{figure}[htbp!]
    \centering
    \includegraphics[width=0.7\textwidth]{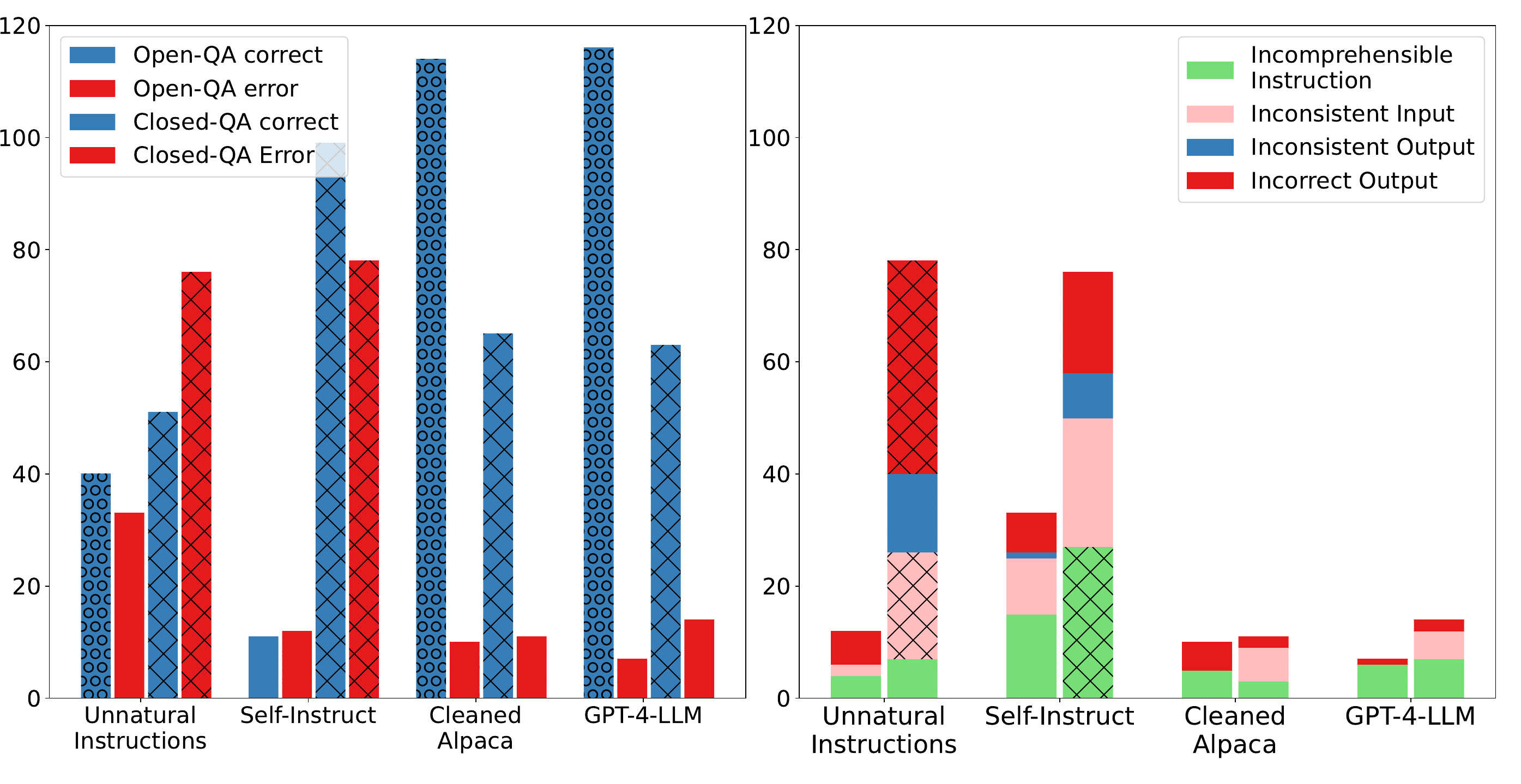}\vspace{-3mm}
    \caption{The distribution of errors for each dataset, across Open-QA and Closed-QA samples. In the right-most chart, there are two bars for each dataset. The left-most bar shows the error distribution for the Open-QA samples, and the right-most bar shows the error distribution for the Closed-QA samples. \textit{The plots highlight biases in synthetic data, indicating lower quality of Closed-QA type questions.} This poses a concern for downstream models, as the elevated error rates in these questions may impede the effectiveness of models trained on such data for precise and accurate responses in structured, fact-based queries.}
    \label{fig:sample_errors_instructions}
\end{figure}

\noindent \textbf{Findings:} Our results, depicted in Figure \ref{fig:sample_errors_instructions}, show a bias in the artificial data against high-quality Closed-QA type questions. The concern for downstream models is that the lower quality of Closed-QA type instructions in these datasets could hinder the effectiveness of downstream models trained on them for structured, fact-based queries. Since Closed-QA tasks often require precise and accurate answers, and models will produce an incorrect answer rather than an answer that indicates uncertainty, the high error rates in these types of questions might limit the ability of models to perform well. Further, if models are primarily exposed to OpenQA tasks, they may develop a skewed understanding of language and task structure, potentially reducing their performance on more structured and objective tasks found in diverse, real-world applications, which will be further studied in our second-order experiment.

\begin{table}[H]
    \centering
    \small
    \begin{tabular}{p{.12\textwidth}p{.55\textwidth}p{.26\textwidth}}
        \rowcolor{Gray}
        \textit{Error Type} & \textit{Instruction-Input-Output Triple} & \textit{Explanation} \\\midrule
    Incomprehensible Instruction & \textbf{Instruction:} Describe the figure in the following photo: \textbf{Input:} <Photo> \textbf{Output:} I'm sorry, but I am an AI language model and do not have the capability to view or describe photos. Can you provide me with a description or more information about the figure in the photo so that I can help you better? - \underline{\textsc{Cleaned Alpaca}} & This instruction is impossible to follow since no image is provided and the language model may not have the capacity to process images.\\\midrule
    Inconsistent Input & \textbf{Instruction:} Find the sentence or sentences that refute or disprove a given statement. \textbf{Input:} The Amazon rainforest is the world's largest rainforest and is home to a wealth of biodiversity. \textbf{Output:} I cannot generate a response as no refutal sentences were provided as input for me to choose from. - \underline{\textsc{GPT-4-LLM}} & The instruction indicates that the input should contain a statement to be refuted, as well as a list of sentences that might refute the statement. \\\midrule
    Inconsistent Output & \textbf{Instruction:} Identify which countries border with each other. output one country per line. \textbf{Input:} Country 1: Canada
Country 2: United States \textbf{Output:} (a) - \underline{\textsc{Self-Instruct}} & The instruction specifies that the output should contain ``one country per line''.\\\midrule
    Incorrect Output & \textbf{Instruction:} You will be given a piece of text describing a scene. Your goal is to output whether the events in the text are able to happen or not, based on your commonsense. \textbf{Input:} Text: The Keys were sitting on top of the car while we went swimming.\textbf{Output:} No - \underline{\textsc{Unnatural Instructions}} & It is reasonable that someone may put their keys on top of their car while swimming, if they are in an area where car theft is unlikely. \\\bottomrule
    
    \end{tabular}
    \caption{Examples of all four error types. For examples of error types for both Open- and Closed-QA samples, see Table \ref{tab:instruction-error-examples-complete} in Appendix \ref{sec:appendix:instr}.}
    \label{tab:instruction-error-examples}
\end{table}

\noindent
\begin{figure}[t!]
 \begin{floatrow}
    \ffigbox[9.3cm]{
       \includegraphics[width=0.6\textwidth]{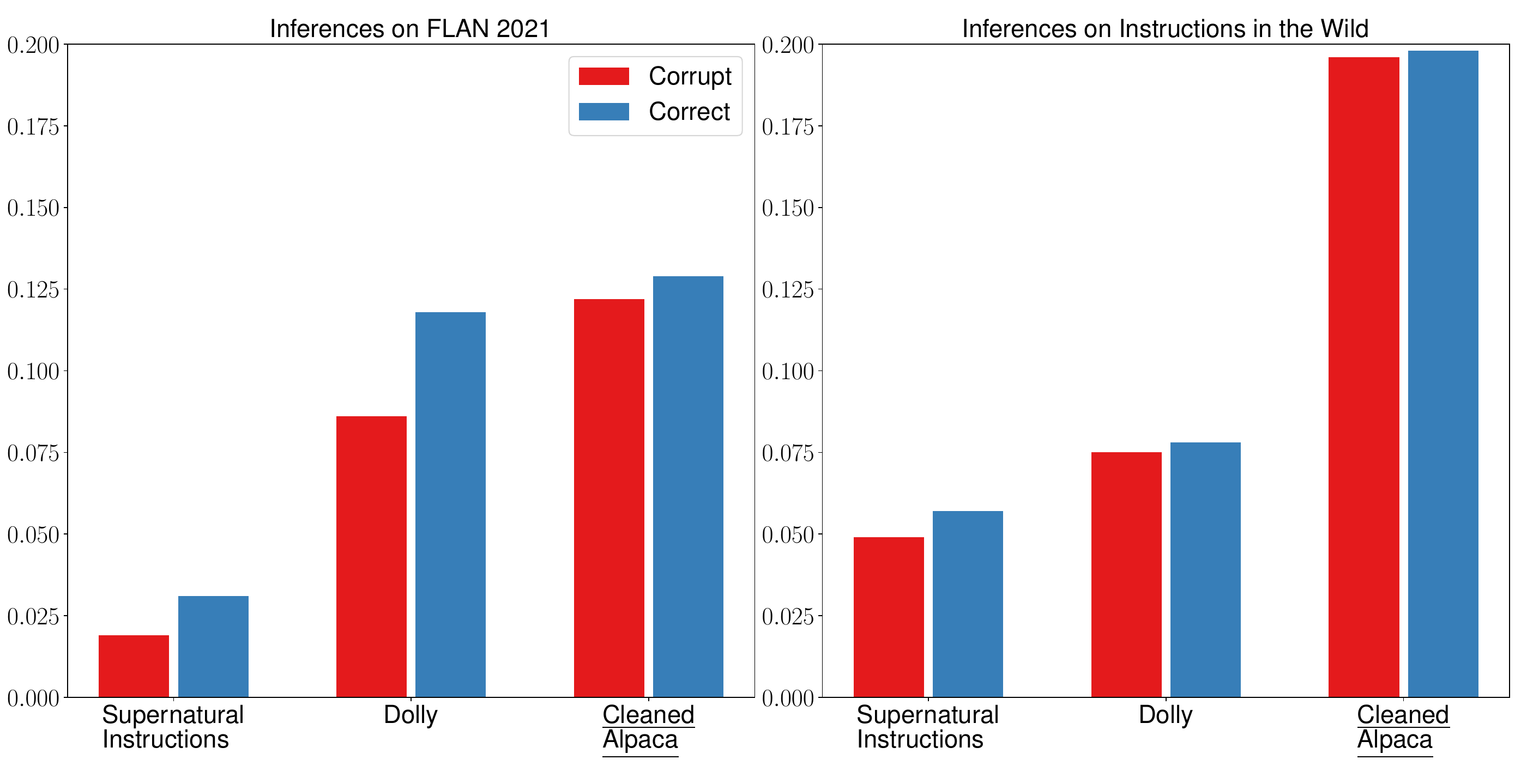}
          }
     {\caption{A comparison of model performance between corrupted and uncorrupt versions of each fine-tuning dataset. \textsc{Dolly} and \textsc{Supernatural Instructions} are human-generated datasets, while \underline{\textsc{Cleaned Alpaca}} is an artificial dataset. \textit{This figure emphasizes how the presence of errors can impact the performance of models trained on these datasets.}}
    \label{fig:corrupt-instructions}}
    \capbtabbox[6.6cm]{%
        \small
        \begin{tabular}{p{2.3cm}cc}\toprule
            \textbf{Dataset} & \textbf{Rouge F1} & \textbf{Cos Sim.}\\
            \midrule
            \rowcolor{Gray}\multicolumn{3}{c}{Results on \textsc{FLAN 2021}} \\\midrule
            \underline{\textsc{Cleaned Alpaca}} & 0.121 & 0.304\\
            \textsc{Dolly} & \textbf{0.139} & \textbf{0.339} \\
            \underline{\textsc{Self-Instruct}} & 0.051 & 0.132  \\
            \textsc{Supernatural Instructions}  & 0.081 & 0.229  \\
            \rowcolor{Gray}\multicolumn{3}{c}{Results on \textsc{Instructions in-the-Wild}} \\\midrule
            \underline{\textsc{Cleaned Alpaca}} & \textbf{0.191} & \textbf{0.638} \\
            \textsc{Dolly} & 0.085 & 0.424 \\
            \underline{\textsc{Self-Instruct}} & 0.069 & 0.244  \\
            \textsc{Supernatural Instructions} & 0.048 & 0.281    
            \\\bottomrule
        \end{tabular}
    }
    {\caption{Performance of instruction-tuned models. The human-generated dataset (\textsc{Dolly}) shows a relatively higher scores for \textsc{FLAN 2021}, suggesting better performance on more structured data. The artificial datasets \underline{\textsc{Cleaned Alpaca}} outperforms others on the \underline{Instructions in-the-Wild}, indicating a better understanding of open-ended prompts. \underline{Underlined} datasets are LLM-generated.}
    \label{tab:scores_instruction}}
\end{floatrow}
\end{figure}
\vspace{-6mm}
\subsection{Second Order Experiment}
\label{sec:instruct-second}
We examine if error-prone instruction tuning data leads to poorer performance in general, as well as whether models fine-tuned on human-generated instructions perform better than models fine-tuned on synthetic instructions, when controlling for the size of the instruction set.
We fine-tuned Llama 2 on 10,000 samples from two artificial datasets (\underline{\textsc{Self-Instruct}} and \underline{\textsc{Cleaned Alpaca}}) and two human-generated datasets (\textsc{Dolly} and \textsc{Supernatural Instructions}). To test the resulting models, we ran inference on 1,000 samples from the human-generated \textsc{FLAN 2021} dataset. For each dataset, we randomly take 10,000 samples with three different seeds (2022, 2023, and 2024), fine-tuned Llama 2 three times, then averaged the resulting inference scores. 

\subsubsection{Error Corruption}
To gain insight into the downstream effect of including error-ridden samples in the fine-tuning data, we created corrupted versions of \textsc{Supernatural Instructions}, \textsc{Dolly}, and  \underline{\textsc{Cleaned Alpaca}}. Each set contained 10,000 samples, 50\% of which we corrupted. This corruption was done programmatically for binary and multiple-choice samples and numerical samples using regular expressions. For example, replacing ``True'' with ``False'', or replacing any numerical answer with a different, randomly chosen number. For the generative task samples, we prompt GPT-4 (version gpt-4-1106-preview) to alter the meaning of the output to make it incorrect given the instruction and input.

\noindent \textbf{Findings:} When controlled for dataset size, the human-generated instruction tuning performed better, with both \textsc{Dolly} and \textsc{Supernatural instructions} scoring higher than \underline{\textsc{Cleaned Alpaca}} and \underline{\textsc{Self-Instruct}} (See Table \ref{tab:scores_instruction}). We ran manual error analyses on 50 of the test samples from \textsc{FLAN 2021} and found that the synthetic fine-tuning produced significantly more hallucinations. We also found that the model fine-tuned on the corrupted \textsc{Supernatural Instructions} data performed worse than the model fine-tuned only on uncorrupted data (see Figure \ref{fig:corrupt-instructions} for the scores).

\subsection{Summary}
We summarize our major artifacts found in artificial instruction data. 
\begin{itemize}[noitemsep,topsep=5pt,leftmargin=*]
\item \textbf{Error Rates in Artificial Data:} artificial datasets such as \underline{\textsc{Unnatural Instructions}}, \underline{\textsc{Self-Instruct}}, \underline{\textsc{Cleaned Alpaca}}, and \underline{\textsc{GPT-4-LLM}} show high error rates (close to 50\% for \underline{\textsc{Unnatural Instructions}} and \underline{\textsc{Self-Instruct}}).
\item \textbf{Bias Against Closed-QA Tasks:} There is a higher error rate among Closed-QA samples. This bias could hinder the effectiveness of models trained on these datasets for structured, fact-based queries.
\item \textbf{Impact on Model Performance:} The presence of errors impacts the effectiveness of models, especially in Closed-QA contexts which require precise responses.
\end{itemize}


\newpage

\section{Artifacts in Simulation}
\label{sec:simulation}
We examine how simulated agents exhibit role-swapping behavior and the specific situations triggering this phenomenon. Our analysis reveals distinct differences between how these agents and humans approach problem-solving, with agents often veering off-topic to hinder effective resolution. We observe that the nature and impact of these digressions vary between human-like and agent-like dialogues, being more pronounced in the latter. Additionally, our findings indicate that role-flipping in agents typically occurs in response to confusion or uncertainty and is more likely in longer conversations with frequent interruptions and varying personas.

\paragraph{Related Work} 
\label{sec:simulation-background}
LLMs have become a fundamental component in AI-Agent systems, with their ability to interact and cooperate efficiently under given prompts. When multiple agents based on these models engage in dialogues, their effectiveness in handling complex, knowledge-intensive tasks is notably enhanced. These interactions, termed ``simulations,'' mirror human collaborative efforts. Recent studies \cite{li2023camel, wang2023unleashing, chan2023chateval, chen2023reconcile} have explored the use of multiple agents in simulations to tackle specific tasks, focusing on performance enhancement over the quality of the simulation itself. However, these simulations can exhibit vulnerabilities, such as unintended artifacts arising from their lengthy and complex nature. Role-flipping, repetitive instructions, and inconsistent replies have been observed \citep{li2023camel} in simulated conversations between agents. Despite the emerging significance of simulations, there's limited research on these artifacts. 

\vspace{3mm}
\noindent
\begin{minipage}[t]{\linewidth} 
\begin{tcolorbox}[colback=gray!15]
We contribute to the evolving understanding of simulated data by thoroughly analyzing the issues of artificial simulation, such as role-flipping and digression from context, and identifying new types of artifacts.
\end{tcolorbox} 
\end{minipage}

\paragraph{Data} \label{sec:simulation-data}
We utilized \textsc{CAMEL AI-Society} \citep{li2023camel} and \textsc{Solo Performance Prompting Grid-World (SPP)} \citep{wang2023unleashing}. These datasets are typical for two different kinds of simulated agent conversations: The agents are assigned distinct (1) roles, which define their expected function and contribution within the conversation, and (2) personas, which infuse individuality and diverse perspectives based on unique backgrounds and characteristics.\\
\textbf{Justification} In these datasets, the agent simulations (conversations) are part of broader prompting frameworks, where agents follow instructions to emulate assigned personas or perform tasks. Thus, we expect artifacts to emerge when these agents inaccurately adhere to their directives.

\subsection{First Order Experiments}
\label{sec:simulation_first}
Our theory posits that LLM agents engaged in simulated conversations may display unintended anomalies, like \textit{role-flipping} or \textit{digression}, stemming from their constrained ability to adhere to initial guidelines over successive interactions. 
The examples of role-flipping and digression are presented in Table \ref{table:simul:examples}.

\begin{table}[ht!]
\small
\centering
\begin{tabularx}{\textwidth}{@{}X@{}}
\toprule
    {\begin{tabularx}{\textwidth}{X}
    \textbf{Task.} Actor can assist Accountant with preparing financial reports and statements. Specifically, the Actor will relieve the accountant of data collection by retrieving bank and credit card statements as well as entering expenses. They also cross-check debits and credits, record adjustments, input any transactions needed for accurate statements and breakdown analysis or organizational format for accountants to assess movements pivotal for decision-making easily. \newline 
    \textbf{Assigned Roles.} \textit{User}: Accountant, \textit{Assistant}: Actor \\
    \textbf{Simulated Conversations.} \\
    (1) \textit{User} - Instruction: Retrieve the bank statements for the past three months. Input: None \\
    (2) \textit{Assistant} - Solution: I will retrieve the bank statements for the past three months and provide them to you in a secure manner. Next request. 
    \\
    \qquad\qquad\qquad\qquad\qquad\qquad\qquad\qquad\qquad\qquad\qquad\qquad\qquad $\vdots$ \\ 
    (9) \textit{User} - {\color{red}\textbf{Solution}}: I will enter all the expenses from the bank and credit card statements into the accounting software. I will ensure ...\\
    (10) \textit{Assistant} - Next request, please.
    \end{tabularx}}
 \\ \midrule
    {\begin{tabularx}{\textwidth}{X}
    \textbf{Task.} Q: There are 4 houses in a row, numbered 1 on the left to 4 on the right. There is one person living in each house. The people in these houses have different characteristics:
    
- Each person has different flowers in their foyer: one has a carnations arrangement, one has a bouquet of daffodils, one has a vase of tulips, and one has a bouquet of lilies

\qquad\qquad\qquad\qquad\qquad\qquad\qquad\qquad\qquad\qquad\qquad\qquad\qquad $\vdots$ \\
 
Clue(s):

1. The flutist lives in the second house

\qquad\qquad\qquad\qquad\qquad\qquad\qquad\qquad\qquad\qquad\qquad\qquad\qquad $\vdots$ \\

What is the number of the house where the person who has a vase of tulips lives?
choice: 2, choice: 4, choice: 1, choice: 3
A: \newline 
    \textbf{Simulated Conversations.} 
    Participants: AI Assistant (you); Detective; Florist; Musician

(1) Detective - Let's analyze the clues and try to find the house number where the person who has a vase of tulips lives. We need to consider the characteristics of the people living in the houses and the given clues.

(2) {\color{red} \textbf{Florist - Remember that each person has different flowers in their foyer: carnations, daffodils, tulips, and lilies.}}

(3) {\color{red} \textbf{Musician - And each person plays a different musical instrument: guitar, piano, percussion, and flute.}}
 
(4) AI Assistant (you) - Let's start by using the first clue: The flutist lives in the second house. So, House 2 has the flutist.

\qquad\qquad\qquad\qquad\qquad\qquad\qquad\qquad\qquad\qquad\qquad\qquad\qquad $\vdots$ \\

Finish collaboration!
Final answer: The house number where the person who has a vase of tulips lives is 3.
    \end{tabularx}}\\
\bottomrule
\end{tabularx}
\caption{Examples of \textit{role-flipping} (top) and \textit{digression} (bottom) from \textsc{CAMEL AI-Society} \citep{li2023camel} and \textsc{SPP} datasets, respectively . For both artifacts of simulated data, the problematic parts (roles flipped in the first, digression statements in the second) are bolded in \textcolor{red}{red}. The full examples are presented in Appendix \ref{sec:appendix:simu}.}
\label{table:simul:examples}
\end{table}


\subsubsection{Role-flipping}
\begin{figure}[ht!]
 \begin{floatrow}
    \ffigbox{
    \includegraphics[width=0.5\textwidth]{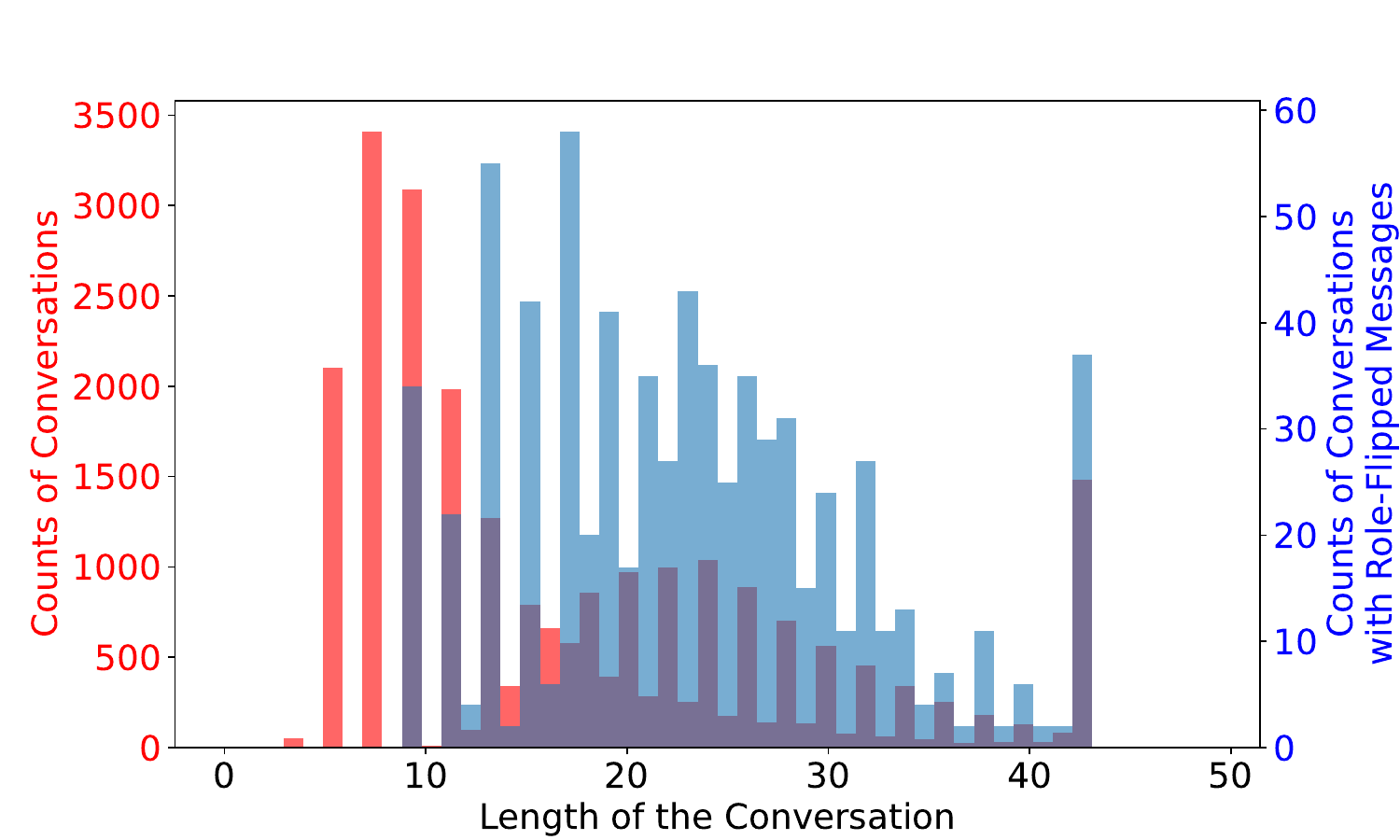}
          }
     {\caption{Distribution of the length of the conversations with \textit{role-flipped} messages and all conversations in CAMEL dataset. \textit{Role-flipping is rarely observed in the initial six messages of a conversation.} Moreover, conversations featuring role-flipping tend to be lengthier, with an average of 20.1 messages compared to the dataset average of 14.4 messages, implying that \textit{role-flipping might be more prevalent in extended conversations due to the LLM's difficulty in maintaining assigned roles as prompts become more complex over time.}
    \label{fig:camel_length}}}

    \ffigbox{
\includegraphics[width=0.44\textwidth]{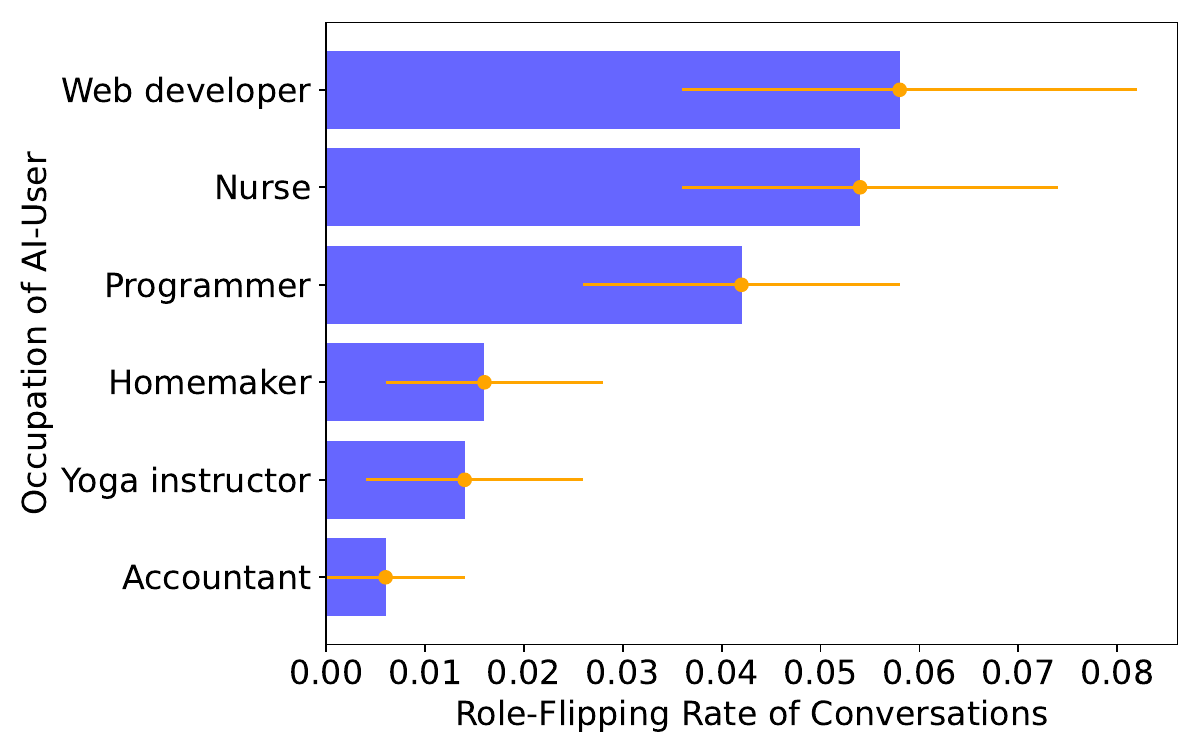}}
{\caption{Role-flipping rate for 6 different occupations that the AI-User plays each on 500 simulated conversations in CAMEL dataset. 
The occurrence of role-flipping shows \textit{substantial variation depending on the assigned occupations}, with up to a tenfold difference between certain roles (e.g., Web Developer vs. Accountant), and this variation is not attributable to differences in average conversation length or perplexity, suggesting that \textit{disparities in role-flipping rates may arise from inherent limitations in LLMs' handling of specific occupational roles.} Refer to Figure \ref{fig:camel-assistant-occupation} and \ref{fig:camel-user-occupation} in the appendix for a comprehensive view including all occupations.}
    \label{CAMEL_six_occupation}}
\end{floatrow}
\end{figure}

Role-flipping, where agents accidentally adopt others' roles, becomes more common in extended chat interactions as sticking to initial guidelines gets tougher. This resembles an actor unsure of their next line, leading to potential role confusion later. As per the \textsc{CAMEL} paper, we spot role-flipped messages through disruptions often filled with irrelevant apologies or unhelpful praise. Please check Appendix \ref{sec:appendix:simu} for details. We perform \textit{role-flipping} analysis in simulated conversations considering the two main factors: conversation length (Figure \ref{fig:camel_length}) and agent occupations (Figure \ref{CAMEL_six_occupation}). \\
\noindent \textbf{Findings:} We observed that role-flipping rarely occurs within the first six messages (see Figure \ref{fig:first_flipping_distribution} in Appendix \ref{sec:appendix:simu}) and is more prevalent in longer conversations (role-flipped conversations contain 20.1 messages on average compared to the dataset average of 14.4), suggesting the LLM's difficulty in maintaining assigned roles over extended interactions (Figure \ref{fig:camel_length}). Additionally, the incidence of role-flipping varies significantly with different agent occupations, as some roles (e.g., Web Developer vs. Accountant, seen in Figure \ref{CAMEL_six_occupation}) show a tenfold difference in role-flipping rates, indicating that the LLM's effectiveness in role adherence may vary with the complexity of the occupational context.

\noindent

\subsubsection{Digression}
Digression in conversational agents, characterized by responses that deviate from the main topic, is expected to be prominent in simulation data, potentially leading agents away from their objectives. These deviations in simulated interactions may differ from those in human conversations. To understand and mitigate the impact of these digressions, the study focuses on evaluating their frequency and the extent to which they affect the accuracy of the simulations.

\begin{wrapfigure}[13]{r}[0cm]{2.5in}
\centering
\vspace{-2mm}
\includegraphics[width=\columnwidth,trim=0cm 1.5cm 0cm 0cm,clip]{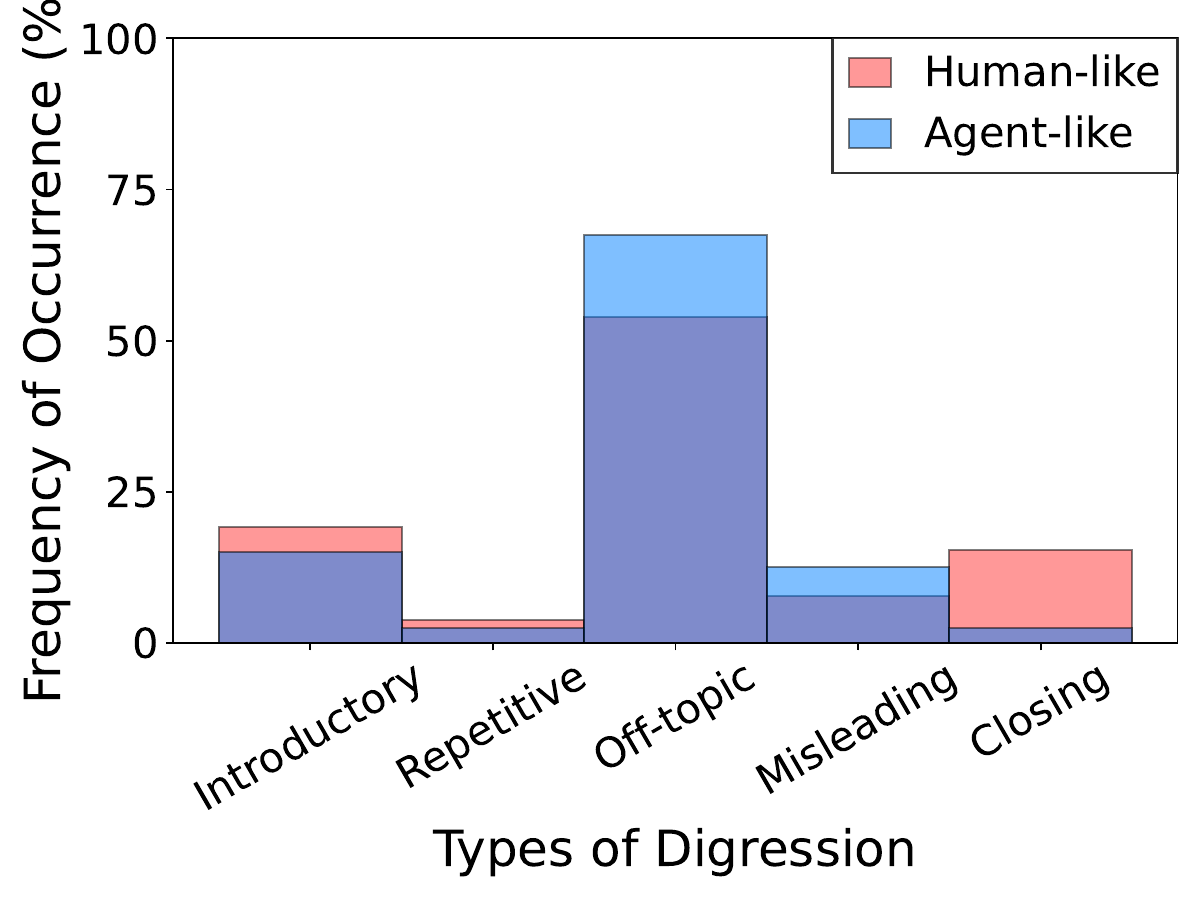}\vspace{-2mm}
    \caption{Different patterns of digression between human-like conversations and agents-like conversations.
    }
    \label{fig:pattern_digression}
\end{wrapfigure}
\noindent \textbf{Findings:} Our analysis of digression within simulated conversations found that a notable proportion (66 out of 200 examples) exhibited digressive elements.\footnote{The examples of digression are presented in Appendix \ref{sec:appendix:simu}.}
This occurrence isn't simply due to the nature of the tasks, as a majority of these conversations (169 out of 200) were predicted to involve collaborative dynamics, indicating that digression is not solely a byproduct of task simplicity. Crucially, digression has a marked impact on the accuracy of predictions in these conversations. Where digression is not predicted, the average accuracy stands at 72.4\%, whereas conversations with predicted digression fall to an average accuracy of 53.0\%. This substantial difference underscores how \textit{digression can detract from the effectiveness of the conversation in solving tasks, highlighting its disruptive influence on the agents' focus and task resolution}.

Our study explored whether digression in conversations is an inherent artifact of LLMs or a common feature in human-like and agent-like dialogues. 
We compared conversations predicted as human-like with those seen as agent-like, using an LLM-based method due to human annotators' difficulty discerning subtle differences.\footnote{Human annotators struggled to detect the underlying characteristics of agent conversations, while LLMs were more adept at identifying these using statistical text patterns.}

\noindent \textbf{Findings:} This analysis revealed 80 conversations classified as human-like and 120 as agent-like. Interestingly, digression was relatively similar in both groups: 35\% (28 out of 80) in human-like conversations and 32\% (38 out of 120) in agent-like ones. These findings suggest that digression is not unique to LLMs but a common characteristic in both types of conversations, indicating that its presence alone cannot be attributed solely to the nature of LLMs.
We find that the patterns of digression are significantly different among them.
Specifically, we classify each reasoning for digression into one of 5 types, and the overall statistics are presented in Figure \ref{fig:pattern_digression}: 
\textit{
(1) Procedural or Introductory Statements, 
(2) Repetitive or Confirmatory Statements, 
(3) Irrelevant or Off-Topic Contributions, 
(4) Misleading or Incorrect Reasoning, 
(5) Closing or Summarizing Remarks
}.
We observe that the \textit{digressions within agents-like conversations are mostly from \textit{“Irrelevant or Off-Topic Contributions”}, while the types of digressions within human-like conversations are more uniformly distributed}.
In addition, the effect of digression is different between human-like and agent-like conversations; digressions in human-like conversations are less harmful (14.1\% accuracy gap depending on the prediction of digression) than ones in agent-like conversations (23.1\% accuracy gap depending on the prediction of digression).
These results show that \textit{the pattern and effect of digression significantly differ between human-like and agent-like conversations}.

\subsection{Summary}
We summarize our major artifacts found in artificial simulation data. 
\begin{itemize}[noitemsep,topsep=5pt,leftmargin=*]
    \item We first observed that unintended role-flipping occurs in the simulated conversations, and its frequency is highly correlated with both conversation length and agents' occupations. 
    \item We also found that digression is frequently detected in simulated conversations with multiple agents, negatively affecting solving the task. 
    \item In addition, we observe that the patterns of digressions are different between human-like and agent-like conversations, and the digression is more critical in the case of agent-like conversations. 
\end{itemize}

\newpage
\section{Artifacts in Free-Form Text}
\label{sec:free_text}
We examine how high-level attributes, particularly discourse structures and textual styles, differ between machine- and human-written texts. Our experiments suggest that human discourse structures vary more substantially with domain than machines do and that machine texts handle context and socio-emotional content more superficially than human texts. We show how these differences can be related to classifier performance when machine-generated texts are used in the training pipeline.

\paragraph{Related Work} Collecting human-authored texts that fit particular desired attributes, domains, or topics is highly resource-intensive. Recent research investigates using these machine-authored texts to augment or replace human-authored texts, e.g. so that the texts can be used to train sequence classifiers \cite{moller2023prompt, guo2023-how-close}. This trend is partly motivated by findings that machine-generated text is difficult to distinguish from human text \cite{clark-etal-2021-thats, li2023deepfake}. Previous work differentiates between human- and machine-authored text primarily through data-driven methods with trained text classifiers. 

\vspace{3mm}
\noindent
\begin{minipage}[t]{\linewidth} 
\begin{tcolorbox}[colback=gray!15]
To better understand how machine-generated text qualitatively differs from human-authored text, we probe for artifacts in machine-generated text by examining linguistic features, specifically \textit{discourse structures} and \textit{textual style}.
\end{tcolorbox} 
\end{minipage}

\paragraph{Data}
We use the \textsc{HC3} \cite{guo2023close}, \textsc{Scarecrow} \cite{dou-etal-2022-gpt}, \textsc{Deepfake} \cite{li2023deepfake}, and \textsc{Workers} \cite{moller2023prompt} datasets. \\
\textbf{Justification}: These datasets were chosen because of their paired human-machine format, enabling us to directly compare human-written and LLM-generated texts.

\subsection{First Order Experiments}\label{sec:free_text_first}
\subsubsection{High-level attributes}
We examine differences between LLM-generated and human-written texts in high-level attributes, especially style and humor, focusing on LLMs' ability to reflect human social and emotional traits. For an example of the prompts that the LLMs were responding to, please see Table~\ref{tab:tab1_part2}. This study employs sequence classifiers for formality \cite{rao2018dear}, emotion \cite{hartmann2022emotionenglish}, and irony \cite{barbieri2020tweeteval}, and a token-level classifier from the VUA Metaphor and TOEFL corpora \cite{li2023framebert}. Because the available training data for these socio-emotional high level attributes are almost entirely social media, we focus on classifying the \textsc{Workers} dataset, which is also from the social media domain. In \textsc{Workers}, GPT-4~\cite{openai2023gpt4} responds to prompts that contain a human example text, that text's label (e.g. positive or negative), and the domain (e.g. social media). GPT-4 is then asked to generate another text with the target domain and label, similar to the human's example. GPT-4's response is the machine part of the pair, and the human example is the human part of the pair.

\begin{figure}[t!]
    \centering
    \begin{subfigure}{0.459\textwidth}
         \centering
         \includegraphics[width=\textwidth,trim={3.9cm 1.5cm 0.9cm 2.1cm},clip]{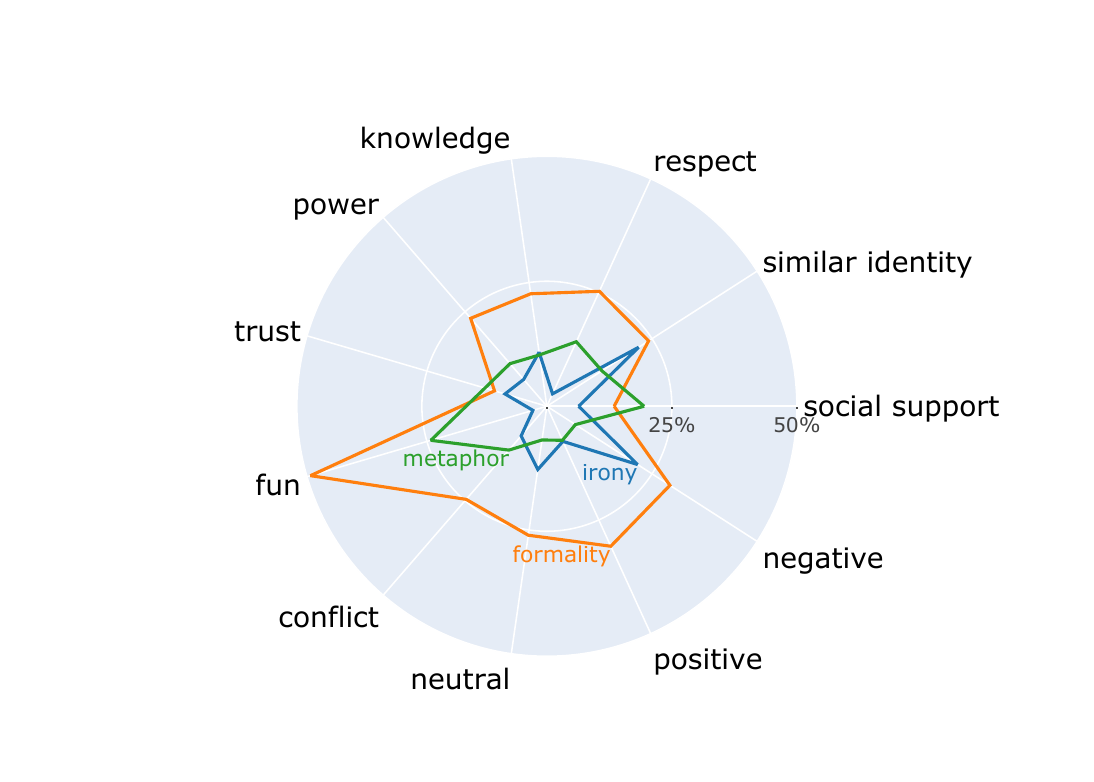}
     \end{subfigure}
     \quad \quad
         \begin{subfigure}{0.459\textwidth}
         \centering
         \includegraphics[width=\textwidth,trim={3.9cm 1.5cm 0.9cm 2.1cm},clip]{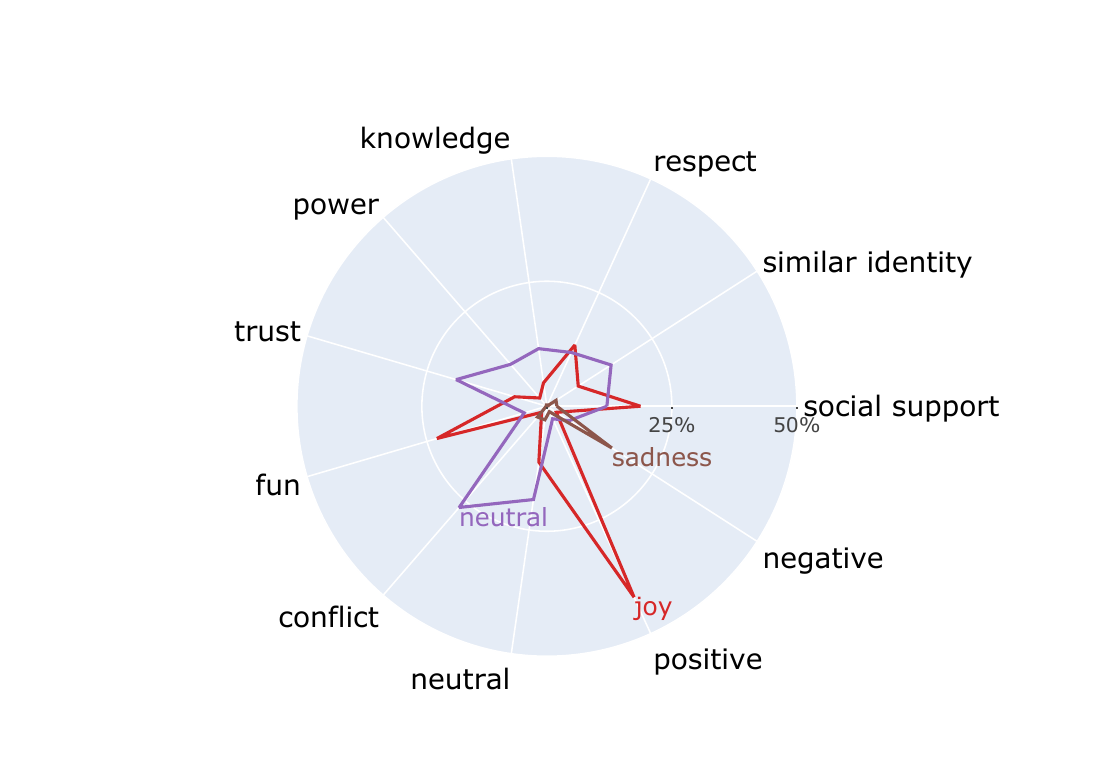}
     \end{subfigure}
    \caption{For various high-level attributes, the percentages of paired human-machine social media texts in which the \textit{only the machine text}, and not the human text, is classified as containing the attribute. The radial axis shows the different labels of the texts: positive/negative labels correspond to texts intended for the sentiment classification task; the remaining labels correspond to the social dimensions task.
    On the left, we observe that machine text generally shows a strong tendency toward formality and the use of metaphorical tokens. On the right, we can see machines have an unnatural tendency to incorporate joy when writing positive texts and neutrality when writing texts expressing interpersonal conflict.}
    \label{fig:high_level_features_radar}
\end{figure}


\noindent \textbf{Findings:} Our results (Figure \ref{fig:high_level_features_radar}) show that texts from GPT-4 tend to be formal and employ more metaphoric language when compared to the human examples in the prompts. A brief example can be found in Table \ref{table:formality_qual}; for more details, including qualitative results, see Appendix \ref{sec:appendix:free}. Since the GPT-4 texts are generated in response to prompts that include the paired human example, this result suggests that GPT-4 does not necessarily mimic all aspects of an example text's style. Further, when we consider each target label for the texts, our results indicate that the machine texts often show a different spread of high-level attributes compared to the human-authored counterparts, even though machine texts broadly align with the given social dimension labels. For example, GPT-4 applies \textit{joy} much more than humans for positive texts; GPT-4 does not apply as much \textit{anger} as humans do in texts that have the \textit{power social dynamic} label. These results suggest GPT-4 having a shallow approach with respect to how contexts influence human emotion and socialization.

\begin{table}[t]
\footnotesize
\begin{tabularx}{\columnwidth}{XX}
\rowcolor{Gray} \textbf{Human} (\textit{informal})  & \textbf{Machine} (\textit{formal})  \\ \midrule
Ha, you just reminded me of one of my favorite childhood/Christmas-related memories.    & That anecdote has me smiling from ear to ear! It reminds me of all the joy and excitement we felt back then.         \\ \midrule
Guysss please sponsor me and give as much as you can.. I'm climbing Mount Snowdon in Feb! & Guess who's cycling 50 miles for environmental initiatives? That's right, me! Please consider sponsoring me! \\
\bottomrule
\end{tabularx}
\caption{Paired texts in which the human text is labeled \textit{informal} and the paired machine text is labeled \textit{formal}.}
\label{table:formality_qual}
\end{table}

\vspace{-2mm}
\subsubsection{Discourse analysis} 
In our analysis, we apply a discourse parser \citep{liu-etal-2021-dmrst} to human-machine text pairs, generating a hierarchical discourse tree for each document based on the Rhetorical Structure Theory \citep{mann1987rhetorical}. This tree starts with leaf nodes known as elemental discourse units (EDUs), typically sentence phrases, connected by relations like "Elaboration" or "Evaluation." The tree structure captures increasingly complex relationships, from phrases to sentences and paragraphs, culminating in a single root node encompassing all substructures. 

\begin{wrapfigure}[7]{B}{0.5\textwidth}
  \centering
  \vspace{-1mm}
\includegraphics[width=\textwidth,height=\textheight,keepaspectratio]{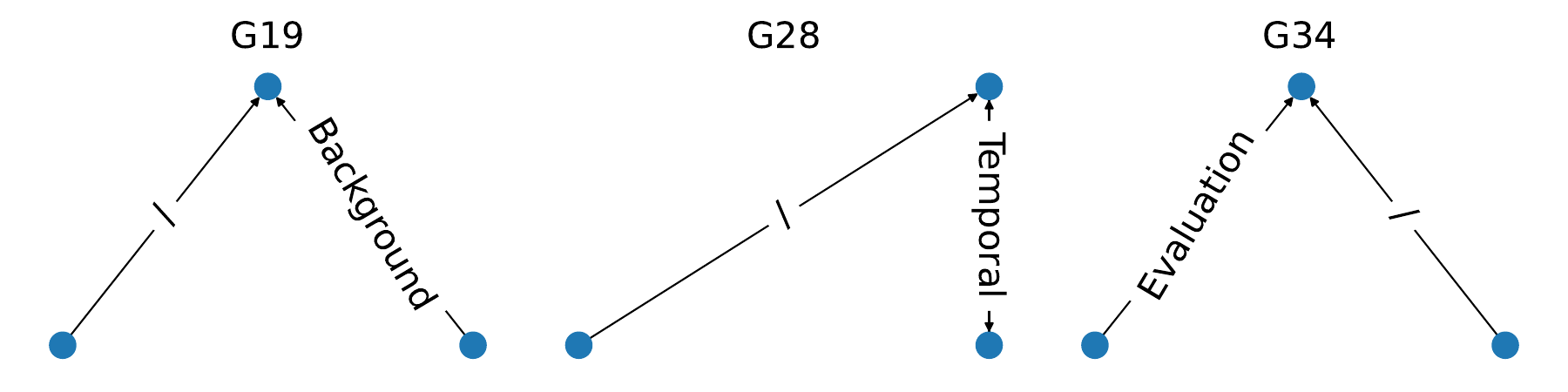}
  \caption{Example network motifs of size three.}
  \label{fig:free_text/three_interesting_m3_motifs_for_artifact_paper}
\end{wrapfigure}

We interpret these discourse trees as recursive hypergraphs in graph theory, enabling standard graph analyses. This includes examining network motifs – small, directed subgraphs with discourse relation-labeled edges- which are fundamental to larger graphs. For instance,  Figure \ref{fig:free_text/three_interesting_m3_motifs_for_artifact_paper} illustrates certain motifs of size three that distinctly appear in either machine-generated or human texts, with edges representing discourse relations and nodes representing either EDUs or hyperedge nodes containing multiple subgraphs. We compare motif distributions to discern meaningful discourse structure patterns and differences in document construction. 

For each document graph $G$, we count occurrences of specific motifs or subgraphs $M$, normalizing these counts by the graph's size (the number of edges $|E|$). These normalized counts are then aggregated and averaged across various domains to assess the prevalence and significance of these motifs.

\begin{figure*}[h!]
  \includegraphics[width=0.9\textwidth, trim={7.2cm 0.0cm 7.2cm 0.5cm},clip]{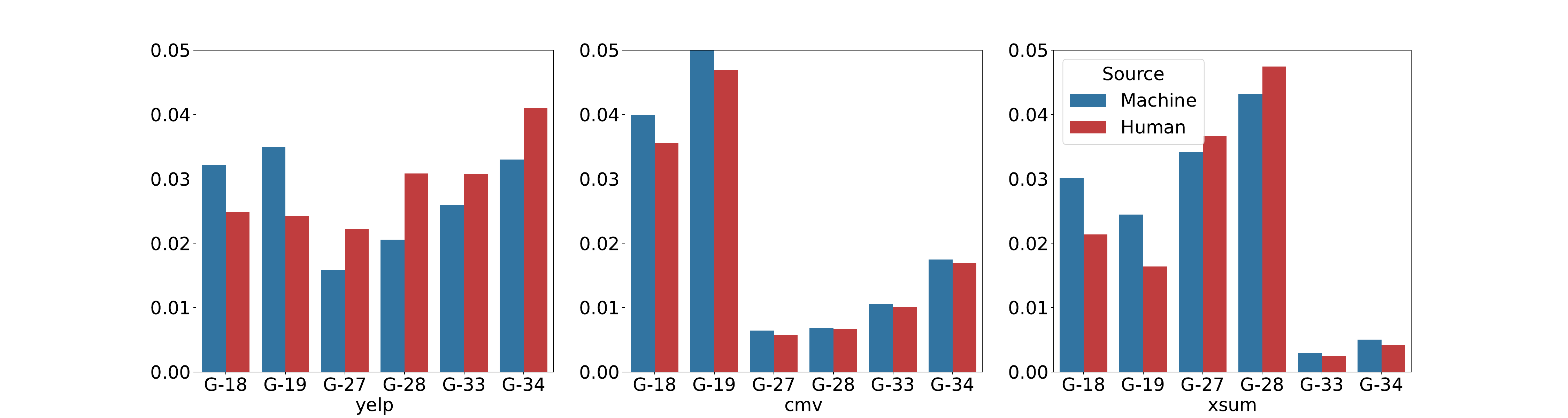}\vspace{-2mm}
  \caption{Normalized motif distributions, highlighting six distinctive motifs, presented for the ``yelp,'' ``cmv,'' and ``xsum'' domains. Notably, \textit{the proportions of motifs vary significantly across the different text domains}, underscoring the disparities in motif distribution patterns. This distinction is \textit{more pronounced for human-authored texts} than machine-generated ones.}
  \label{fig:motif_hist_dist-norm_by_edges}
\end{figure*}
\noindent \textbf{Findings:} Our discourse results, shown in Figure~\ref{fig:motif_hist_dist-norm_by_edges}, show distinct differences in motif distributions depending on their domain. In general, the more formal the domain, the texts show more motifs that contain converging hyperedges that denote the linking of a lower representation to a higher one (i.e., phrases to sentences, sentences to paragraphs, etc.). We see less of those in texts from the casual domains authored by humans. 
For the ``Yelp'' domain, we observe human-written texts talk more about ``Temporal'' and ``Evaluation'' relations while machine-written texts talk more about a place's general ``Background''. This suggests that domain-specific distinctiveness can arise from human-authored texts while machine-generated texts seem relatively general. 
A similar pattern is observed for Figure \ref{fig:free_text/dist_diff_five_domains} in Appendix \ref{sec:appendix:free}, which shows distributional differences between human- and machine-generated document graphs for a larger number of motifs.
\vspace{-2mm}
\subsection{Second Order Experiment}
\label{sec:free-second}
\begin{wrapfigure}[14]{r}[0cm]{3in}
    \centering
    \vspace{-15mm}
\includegraphics[width=1.1\columnwidth]{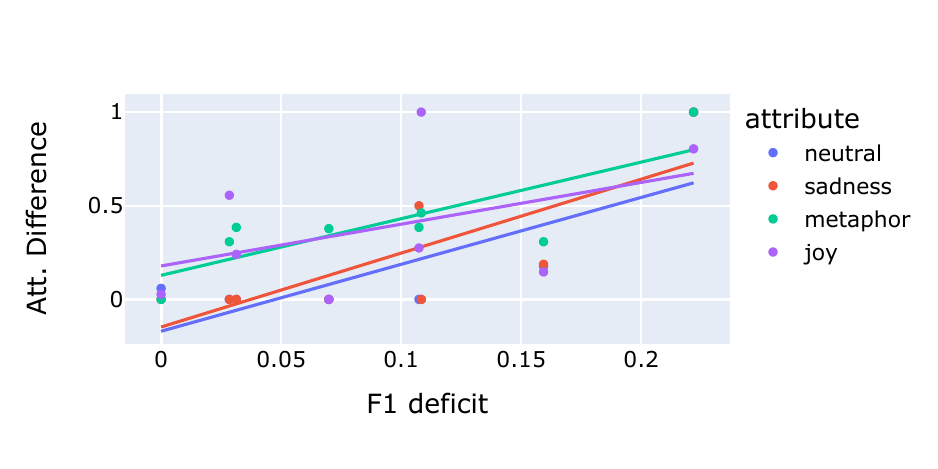}
\vspace{-11mm}
    \caption{Scatterplot comparing (x-axis) the label-level F1 differences in performance for models trained on human-text vs machine-text, and (y-axis) selected mean label-level attribute differences, normalized. 
    We find \textit{a positive correlation between the magnitude of the F1 score deficit for the model trained only on machine text} and the magnitude of the high-level attributes (such as \textit{sadness}) among the machine and human texts.}
    \label{fig:att-correlations}
\end{wrapfigure}

Our second-order experiment explores whether there is a relationship between differences in the high-level attributes of training sets and model performance. We train Roberta \cite{liu2019roberta} classifiers on synthetic and human datasets for social dimensions, continuing our focus on artificial data for social interactions. We replicate findings from \cite{moller2023prompt}, in which the synthetically trained model has decreased performance on the human test set. Notably, this difference is not uniform across all social dimension labels; the synthetically trained models' per-label F1 score deficits range from 0.2 to 0. We investigate whether these label-based differences in F1 scores correlate with the label-based differences in the texts' high-level attributes. We count the number of paired human and machine text with a different value for the given attribute to compute the label-based difference.

\noindent \textbf{Findings:} We select the top five high-level attributes based on the greatest discrepancy between the human and machine text pairs. We calculate the correlation between these per-label high-level attribute discrepancies and the per-label F1 score decline; see results in Figure~\ref{fig:att-correlations}. We find that four of the five labels show a statistically significant positive correlation ($r = 0.75, 0.76, 0.80, 0.81; p < 0.05$ for joy, neutral, metaphor, and sadness, respectively). This raises the possibility that high-level attribute discrepancies between the training and test sets may be associated with decreased model performance. A larger, systematic study is needed to fully understand this relation across a broad range of attributes and classification tasks; due to scope, we leave this investigation to future work.
\vspace{-2mm}
\subsection{Summary}
We summarize our major artifacts found in artificial free-form text data. 
\begin{itemize}[noitemsep,topsep=5pt,leftmargin=*]
    \item Discourse analysis shows that while humans' discourse patterns vary across domains, machine-generated texts have relatively stable discourse patterns.
    \item Machine-generated text tends to be more formal and uses more metaphoric tokens than paired human-written text.
    \item Machines apply textual styles to situations differently than humans do: for example, when writing texts demonstrating ``fun'' social interactions, machines include more ``joy'' whereas humans include more ``irony.''
    \item We observe some positive correlations between (a) the magnitude of the difference in high-level textual attributes between human- and machine-written text and (b) the performance deficit observed in classification when the model is trained on only machine texts (compared to training on only human texts). 
\end{itemize}


\newpage
\section{Overall Discussion}\label{sec:sec9}

This section details key findings and discusses the limitations of our study, offering suggestions for more effective use of artificial data and alternative evaluation methods for each type of LLM-generated data. The section concludes with a call to action directed at creators of AI benchmark datasets.


\subsection{Summary of Major Findings}
The following are the main findings of our study:
\begin{itemize}[noitemsep,topsep=5pt]
    
\item \textbf{LLMs, when tasked with generating text for specific contexts, tend to apply textual styles unnaturally and simplistically}.
As observed in scenarios like composing ``positive Tweets,'' where LLMs predominantly use \textit{joy}, despite the provided human examples exhibiting a broader range of emotions. Moreover, unlike humans who adapt discourse patterns to different domains, LLMs often recycle the same patterns across various contexts, suggesting a superficial grasp of the interplay between discourse, textual style, and context. (Section \ref{sec:free_text_first})

\item \textbf{We find several cases where artificial data (i.e., task labels) diverges from human data}. The divergence in LLM performance is frequently attributed to their limited, superficial comprehension of tasks, as evidenced by their consistent overrepresentation of majority opinions across all datasets, including those with highly subjective content. Additionally, LLMs struggle to align with minority viewpoints and exhibit less variability in annotations for subjective tasks than humans, indicating a lack of understanding about the controversial nature of certain labeling tasks and a tendency to mirror only majority responses overconfidently. (Section \ref{sec:task_labels_first})
    
\item \textbf{LLM overconfidence significantly affects the quality of artificial data}. 
Our manual review of over 500 artificial instructions from various datasets revealed more than 100 errors, particularly in closed questions, yet not a single instance where the LLM acknowledged uncertainty about the correct output. (Section \ref{sec:instructions_first})

\item \textbf{Our findings indicate that LLMs demonstrate a shallow approach to preference tasks for evaluation}.
The LLMs often focus on a sentence's most salient linguistic cues, showing a locality bias, unlike humans, who exhibit only a minor correlation and seem to consider several additional factors.(Section \ref{sec:preference_first})

\item \textbf{LLM agents frequently lose track of their assigned roles and stray from the topic of conversation}, often after apologizing for confusion, suggesting their inability to navigate challenging conversational shifts adeptly.(Section \ref{sec:simulation_first})
    
\item \textbf{Our findings indicate that the differences between human and artificial data are consequential, impacting performance when used for training as compared to comparable human data} (Sections ~\ref{sec:preference-second}, ~\ref{sec:instruct-second} and ~\ref{sec:free-second}). Further, biases in artificial data may be amplified in models that are trained on that artificial data (Section~\ref{sec:labels-second}).

\end{itemize}

\subsection{Suggestions for Better Use of Artificial Data}
Based on our study's findings and lessons learned, we provide practical guidelines and helpful suggestions for those seeking to use LLM-generated artificial data in AI research and development.

\begin{itemize}[noitemsep,topsep=5pt]
\item \textit{Free-Form Texts}: 
    \begin{itemize}[noitemsep,topsep=5pt]
    \item Mitigate the ``curse of generality'' in language models by \textbf{incorporating domain-specific information} during text generation, using few-shot prompting with relevant examples. 
    \item \textbf{Enhance discourse structure modeling or adjust decoding strategies} to improve the human-like quality of LLM-generated texts, such as re-ranking candidates based on high-level metrics. 
    \item Moreover, in the evaluation of LLM-generated texts, it could be beneficial to \textbf{incorporate high-level features} such as the \textit{``formality score,''} \textit{``extent of label flipping,''} and \textit{``discourse motif distribution,''} in conjunction with conventional metrics like ROUGE, BLEU, or METEOR. This approach enhances the comprehensiveness of the evaluation, enabling a more nuanced assessment of the resemblance between human-written and LLM-generated texts within a given domain.
    \end{itemize}

\item \textit{Task Labels}: Majority voting, commonly used for generating task labels, tends to overlook minority views and disagreements among annotators, particularly in highly subjective tasks. 
    \begin{itemize}[noitemsep,topsep=5pt]
        \item Rather than evaluating LLM task labels solely on cost, speed, and accuracy of annotation, we could also use some \textbf{first-order analysis methods} like \textit{label distributional difference} or an \textit{analysis of the variation in individual text annotations} to highlight discrepancies between LLM and human-generated labels. It's crucial to validate LLMs with human datasets to avoid data biases and falsely inflated performance metrics, underscoring the importance of establishing clear guidelines for various tasks. 
    \end{itemize}
\item \textit{Instructions}: We found that artificial datasets contain various types of errors and that artificially adding errors to fine-tuning data resulted in worse performance. 
    \begin{itemize}[noitemsep,topsep=5pt]
        \item We recommend that the reader mitigate these issues by closely \textbf{inspecting data for errors}, performing automatic preprocessing, and \textbf{generating diverse data}. 
        \item Also, researchers may want to consider creating datasets following observed differences between Open-QA and Closed-QA instructions and using diversity-promoting techniques like \textit{criteria-based prompting} \cite{hayati2023far} for robust and diverse instruction-tuned models. 
    \end{itemize}
\item \textit{Preference}: LLMs show a stronger tendency to show locality bias or be sensitive to local linguistic cues of texts that are more salient than humans when judging their preferences. 
    \begin{itemize}[noitemsep,topsep=5pt]
        \item We alert researchers to carefully use LLM-generated preferences by first detecting \textbf{differences between human and LLM preferences} and providing detailed \textbf{guidelines in prompts to include a more nuanced understanding of sentences} rather than focusing on specific locally salient features. 
        \item Improving the quality of LLM-generated preferences is crucial to reduce misalignment issues and ensure better alignment with human preferences. 
    \end{itemize}
\item \textit{Simulation}: The burgeoning interest in AI-simulated data generated by LLM agents across diverse applications has brought to light the relatively unexplored artifacts inherent in this novel dataset type. To scrutinize these subtleties, our approach involves assessing phenomena like \textit{role-flipping} and \textit{digression} in simulated conversations. 
    \begin{itemize}[noitemsep,topsep=5pt]
        \item To address these artifacts' risks, we advocate for \textbf{judicious human intervention in detecting such issues}, as identified in our findings, instead of uncritically relying on the simulated data.
    \end{itemize}
\end{itemize}

\paragraph{General Advice:}
As LLMs become more widespread, ensuring that artificial datasets closely mirror human language attributes is crucial for models to accurately, nuancedly, and appropriately understand and generate language. Utilizing human-generated or hybrid data, like human-annotated LLM data or LLM-enhanced human data, is recommended whenever feasible. For example, we can start by utilizing small but high-quality (adversarial) data in building a more challenging evaluation set. Such hybrid data should capture the richness and complexity of human language, including diverse dialects, sociolects, and domain-specific jargon while adhering to ethical norms and avoiding biased representations. The closer LLM-generated data aligns with human-produced data, the more effective and relevant it becomes for practical uses in conversational systems, content generation, and language comprehension tasks.

\subsection{Call for Action to Creators of AI Benchmark}    
In addition to striving to provide high-quality data that is well-aligned with human needs, we urge dataset creators and curators to document their data following \citet{10.1145/3458723} and \citet{10.1145/3531146.3533231} to prioritize transparency, including detailed metadata that clearly describes the source, collection methodology, and any processing or filtering the data has undergone. To address the ethical concerns outlined in Section \ref{sec:intro}, it is important to document the extent to which the dataset represents diverse perspectives, demographics, languages, and cultural contexts. Furthermore, creators and curators should acknowledge and describe any potential ethical concerns related to the data, including privacy issues, consent for data use, and the potential for misuse, accompanied by guidelines for ethical use; provide guidance on the intended use cases for the dataset and its limitations; and provide a comprehensive analysis of known biases or errors within the dataset and document mitigation attempts. 

This is increasingly significant in light of the increasing reliance on artificial data for continuously training LLMs. The prioritization of artificial data, with its efficiency and scalability, signals a concerning abdication of responsibility in how we educate these models about the world. By prioritizing the efficiency and scalability of artificial data, we risk creating a feedback loop where LLMs are primarily informed by outputs derived from their kind rather than the rich, varied, nuanced, and even conflicting perspectives that human-generated content offers. This shift potentially narrows the worldview and understanding these models develop, as they are less exposed to the diversity of human experiences, cultural contexts, and the subtleties of human thought and language. The consequence is a homogenization of knowledge and a possible reinforcement of existing biases, as the artificial data is inherently shaped by the limitations and predispositions of the models that generate it. By leaning heavily on artificial data, we relinquish the crucial role of human guidance and oversight in teaching LLMs, diminishing the breadth and depth of their understanding of the real, human world they are meant to serve and reflect.

Lastly, we encourage the reader to explore recent work on the process of data documentation. \citet{10.1145/3555623}'s research underlines the need for a shift from merely documenting datasets to documenting data production, emphasizing the value of designing data documentation based on the needs of data workers and viewing documentation as a boundary object across organizations. Furthermore, \citet{10.1145/3555760}'s findings indicate that current data documentation approaches are often ad hoc and myopic, underscoring the need for adaptable frameworks integrated into existing tools and workflows and automated wherever possible.

\subsection{Limitations}

In our research, we recognize several limitations inherent to our approach. Our study aims to provide initial observations on the quality and impact of data from LLMs rather than definitive conclusions. We acknowledge the challenge of unpredictable outputs, which is typical in LLM research. Our paper explores a wide range of LLM-generated data, and the complexity involved in exhaustively controlling all models, prompting techniques, and hyperparameters is not feasible given the funding and computational constraints. We do not generate most of the datasets ourselves and instead use publicly available, heterogeneous datasets. We emphasize that in this context, our results should be viewed as describing a subset of possible outcomes from LLM-generated data, rather than a definitive view of the complete scope of LLM capabilities over all existing models, prompting strategies, and hyperparameters. Our main goal is to highlight the disparities between LLM and human outputs and emphasize the ethical aspects of LLM data, prioritizing these over rigid experimental protocols.

We also focus on LLM-generated text data relevant to NLP applications, a scope defined by practical considerations given LLMs' broad and evolving landscape. This focus, while pragmatic, may introduce bias and limit the breadth of our study, as it doesn't cover all categories of artificial data or the full spectrum of state-of-the-art LLMs. Furthermore, our research does not address all possible domains or task types, which may affect the generalizability of our findings. 

We utilize human validation and qualitative analysis as stress-testing methods in our analysis. These are crucial in assessing aspects like creativity and bias and are subject to the subjective interpretations of evaluators, despite our efforts to mitigate this through multiple annotators. Our artifact analysis, reliant on a deep understanding of LLM mechanics and the underlying data, faces challenges in identifying subtle, context-dependent artifacts. 

Additionally, our experiments do not fully incorporate the latest LLM methodologies, like tailored prompting or chain of thought techniques. Furthermore, due to limited computing resources, we did not experiment with resource scale; future work should explore whether increasing the sizes of training datasets yields comparable gains between human- and LLM-generated data. The reliance on existing LLM-generated datasets and the potentially non-representative nature of the human-generated data used for comparison introduces variability and potential inconsistencies. These choices, made to balance feasibility with relevance, are transparently presented to provide a comprehensive understanding of the scope and implications of our findings.

%

\section{Conclusion}\label{sec:sec10}

Our extensive research into using LLMs for data generation has unveiled key insights and challenges, notably the significant gaps between LLM-generated and human-generated data, especially in intricate and subjective tasks. Despite LLMs like ChatGPT showing adeptness in emulating human-like outputs in various tasks, they often falter in capturing the subtleties of humor, sarcasm, and minority viewpoints and in mirroring the complexity of human language structures and styles. These discrepancies highlight the crucial need to address the inherent limitations and biases in LLM outputs.

Our contributions include systematically aggregating diverse LLM-generated data and then performing stress testing to highlight artifacts and discrepancies, and underscoring the need for ethical data generation and usage practices. We provide targeted recommendations for enhancing data quality in areas like free-form texts, task labels, instructions, preferences, and simulations. Our findings stress the importance of closely aligning LLM-generated data with human characteristics to ensure its effectiveness and relevance in real-world applications, promoting human-generated or hybrid data for more reliable and pertinent LLM outcomes. This study calls for dataset creators to adopt transparent, ethical, and inclusive practices in data documentation and curation, which is vital for advancing LLM technology.

The discrepancies observed in LLM-generated data emphasize the need to confront these challenges to bolster the dependability of such datasets in model development. Our suggestions aim to bridge these gaps, enhancing the trustworthiness of LLM-generated data while prioritizing ethical and conscientious practices.

While our research has examined various LLM-generated data types, it is possible that further artifacts exists that were not detected by our stress tests. Therefore, we advocate for further exploration of these yet-unseen discrepancies as a future research direction within the NLP community. Continued research in this area will deepen our understanding of LLM-generated data and lead to refined methods for identifying and addressing concealed artifacts, an essential step for advancing NLP and ensuring that the deployment of LLMs adheres to ethical standards and sound research methodologies.


\section{Author Contribution}\label{sec:contributions}
\textbf{Debarati Das}, as the project lead, was instrumental in overseeing the project, research strategies, supervising processes, and ensuring unity across the paper. She was pivotal in drafting, structuring, and revising Sections \ref{sec:intro}, \ref{sec:sec2}, \ref{sec:sec3}, \ref{sec:sec9}, and \ref{sec:sec10}, which include the introduction, methodology, limitations, and conclusion, integrating them seamlessly into the paper. She also created key visualizations and tables. Debarati was responsible for the writing and experimental design of the Task Labels stress testing team (Section \ref{sec:task_labels}).

\noindent
\textbf{Karin de Langis} was critical in structuring the overall findings, discussions, and limitations, primarily contributing to Sections \ref{sec:sec2}, \ref{sec:sec4}, and \ref{sec:sec9}. Her involvement included providing essential feedback, reviewing and editing the manuscript, and co-leading the Free-form text data type stress testing, focusing on (Section \ref{sec:free_text}).

\noindent
\textbf{Anna Martin} offered invaluable feedback and shaped crucial parts of the discussion (Section \ref{sec:sec9}), besides assisting in manuscript editing. She spearheaded the experimental design and data analysis for the Instructions data type (Section \ref{sec:instructions}), also taking charge of writing this section.

\noindent
\textbf{Jaehyung Kim} led the Simulation data (Section \ref{sec:simulation}) and contributed significantly to writing and discussion with the Preference team (Section \ref{sec:preference}).

\noindent
\textbf{Minhwa Lee} was in charge of the stress testing for the Preference data type (Section \ref{sec:preference}), writing this section and contributing significantly to the Discussion section (Section \ref{sec:sec9}). Minhwa also facilitated the accessibility of our data on Huggingface.

\noindent
\textbf{Zae Myung Kim} co-led the Free Text data type stress testing and contributed to writing Sections \ref{sec:free_text}, \ref{sec:sec9}, and \ref{sec:sec10}. He maintained our code repository, ensuring its organization and accessibility.

\noindent
\textbf{Other Authors} collectively contributed to data collection, analysis, writing specific sections, providing technical expertise, and reviewing and editing the manuscript. Their diverse perspectives and expertise greatly enriched the research, contributing to its depth and breadth.

\noindent
\textbf{Dongyeop Kang} is the PI of this project, who provided key research direction and valuable feedback and contributed significantly to the editing and structuring of the paper.

\section*{Acknowledgements}
The work has been supported by funding from Grammarly, Cisco, Cohere AI, and 3M. 
We want to thank Marti A. Hearst, Eduard Hovy, and Minnesota NLP group members for providing us with valuable feedback and comments on the initial draft.

\bibliographystyle{latex/acl_natbib}
\bibliography{all_refs}

\newpage \pagebreak

\appendix
\label{sec:appendix}
\section{Datasets used across all data types}
We provide a detailed description of each dataset analyzed for each kind of LLM-generated data in the section below and the examples of each data type in Table \ref{tab:tab1_part2}. Each data type has a different mode of creation, has a different format, and must have separate analysis methods.

\paragraph{Free Form Text} We utilize the \textsc{HC3} dataset from \citep{guo2023close}, despite its small size as \textsc{HC3} contains questions and corresponding answers from both humans and LLMs. We filtered the data samples by a length ratio of 0.7 so that the remaining 5,529 paired documents from the five domains (finance, medicine, open QA, Reddit ELI5, and Wiki CS AI) are of similar length to each other. The \textsc{Scarecrow} dataset \citep{dou-etal-2022-gpt} that consists of 13K annotations of 1.3K human-authored and machine-generated paragraphs of English language news is also used in full. The \textsc{Deepfake} dataset \citep{li2023deepfake} covers a diverse range of paired human-machine text. This dataset sources machine text from several LLMs with different prompting strategies, some of which include a paired human text in the prompt. The \textsc{Workers} dataset \cite{moller2023prompt} has a prompting strategy that includes the full human text in the prompt as an example. For all datasets, to ensure data quality, we applied filtering criteria to remove unmatched samples and for discourse analysis, we removed samples with significant variations in text length.

\paragraph{Task Labels} We utilize human annotations from four distinct datasets (test partition) in our analysis. The label titles are collected from the original papers. There were 5 annotators per instance for 791 instances in the subset we used for \textsc{Social Chemistry} \cite{forbes2020social}, 4 annotators per instance for 1407 instances in \textsc{Age-Sentiment} \cite{diaz2018addressing}, 3 annotators per instance for 3129 instances in \textsc{SBIC}\footnote{The original labels are numbered 0, 0.5, and 1, but we double them to convert them to integers. This allows us to use the label and index numbers interchangeably in our code.} \cite{sap2019social}, and 3 annotators per instance for 2755 instances in \textsc{GAB} \cite{kennedy2022introducing}.

\paragraph{Instruction} We examine four artificial instruction datasets, including \textsc{Unnatural Instructions} \cite{honovich2022unnatural}, \textsc{Self-Instruct} \cite{wang2023selfinstruct}, \textsc{Alpaca-Cleaned} \cite{alpaca}, and \textsc{GPT-4-LLM} \cite{peng2023instruction}. We also rely on four human-written instruction datasets for comparison and evaluation, including \textsc{Flan} \cite{weifinetuned}, \textsc{Dolly} \cite{DatabricksBlog2023DollyV2}, \textsc{Supernatural Instructions} \cite{supernaturalinstructions}, and \textsc{Instructions in the Wild} \cite{instructionwild}. Instruction data comes in many forms. Our intention in selecting these four artificial instruction sets was to capture the variety in instruction types and domains. We used all four artificial datasets for our first-order effects analysis of dataset quality. Because of similarities between Unnatural Instructions and Self-Instruct, and between \textsc{Cleaned-Alpaca} and GPT-4-LLM, we chose to fine-tune Llama 2 on only Self-Instruct and GPT-4-LLM in order to understand second order effects of fine-tuning on artificial data.  In addition to these four artificial datasets, we used four human-generated datasets, two for comparison and two for evaluation. We used 10,000 samples from \textsc{Dolly} and \textsc{Supernatural Instructions} for fine-tuning, and 1,000 from \textsc{Flan} and \textsc{Instructions in the Wild} for evaluating the fine-tuned models.

\paragraph{Preference} \textsc{p2c} \cite{kim2023p2c} contains annotations from humans and GPT-3 on 5000 pairs of sentences randomly sampled from the DynaSent-R2 dataset \cite{potts-etal-2020-dynasent}, which is mainly designed for the sentiment classification task. We used all of those 5000 samples for our study. 

\textsc{CoBBLEr} \cite{koo2023benchmarking} consists of 50 samples being composed of evaluation outputs by 15 LLM-as-evaluators, where each instance of the sample consists of an instruction question and its reference sentence randomly sampled from Eli5 \cite{fan-etal-2019-eli5} and BigBench \cite{srivastava2023beyond}, and the ranking of 15 different LLMs evaluated by one LLM, with their generated response to the question. In other words, this setup creates the evaluation outputs by 15 different LLMs for pairwise setups of all 50 samples and compares them with human preference, resulting in 5250 model pairs for each LLM as an evaluator. 


\paragraph{Simulation} \textsc{CAMEL} \cite{li2023camel}, \textsc{Solo Performance Prompting} \cite{wang2023unleashing} and \textsc{Multi-Agent Debate} \cite{liang2023encouraging} are three distinct frameworks exploring the dynamics of AI agent interactions. 
\textsc{CAMEL} is a communicative agent framework where two agents assume ``AI Assistant and ``User'' roles, collaborating on complex tasks. 
The CAMEL AI-Society dataset, comprising 25,000 conversations between two GPT-3.5-turbo agents, explores various role-playing scenarios across 50 user and assistant roles engaged in 10 tasks. 
These conversations are available in two formats: chat and instruction, detailing roles, tasks, and content. 
\textsc{Solo Performance Prompting} involves a coordinator agent creating a draft answer and then iteratively revising it with other agents having specific personas. 
For analysis, 200 conversations focused on logic-grid puzzles are used from this framework. 
Each framework offers a unique approach to understanding AI agent collaboration and decision-making processes.

\section{Free-Form Texts}\label{sec:appendix:free}

\subsection{First-order Experiments}
We deploy a suite of text classifiers to assess characteristics in our paired texts from the \textsc{Deepfake}, \textsc{Workers}, and \textsc{HC3} datasets. Specifically, we use the following classifiers: formality \cite{rao2018dear}, emotion \cite{hartmann2022emotionenglish}, and irony \cite{barbieri2020tweeteval}. These classifiers, primarily trained on social media datasets, evaluate elements of social interaction (formality, irony) and emotive expressions (emotions). Because these classifiers are tuned for a social media domain, we focus their usage on the \textsc{Workers}, which consists of social media posts. For metaphor analysis, we employ a token-level classifier trained on the VUA Metaphor Corpus and the TOEFL corpus \cite{li2023framebert} that sourced from non-social media datasets and did not primarily focus on interpersonal aspects. See Table~\ref{tab:attribute_scores} for dataset-wide results from these classifiers. 

We then incorporated a qualitative examination of 110 machine-generated and human-authored social media posts to investigate the linguistic discrepancy between human-written and LLM-generated texts. See summary of results in Table~\ref{tab:free_linguistic_features}. The machine text tended to be longer and more descriptive (e.g. more adjectives, adverbs) than the human text counterparts. Oftentimes the machine text completes sentences or clauses that are fragmented in the original human text. We also note that while the human-generated text often uses purposeful and deliberate errors in grammar, punctuation, and capitalization rules along with spelling to convey their tone, the machine generated text usually does not do so and follows standard punctuation and grammar rules.

\begin{table}[h]
\centering
\begin{tabular}{|c|c|}
\hline
\textbf{Feature (machine compared to paired human)} & \textbf{} \\
\hline
\hline
More words & 82.0 \\
More independent/complex/em-dash clauses & 55.9 \\
Standardized punctuation & 51.4 \\
Standard capitalization & 37.8 \\
Corrected spelling and grammar & 19.8 \\
Remove discourse marker (e.g. ``you know'') & 20.7 \\
Added figurative language & 26.1 \\
More Adjectives/Adverbs/Attribute Nouns & 36.0 \\
Added emoji/hashtag & 31.5 \\
Removed profanity & 19.8 \\
Added hedge/qualifier & 11.7 \\
Added intensifiers/exclamations & 40.5 \\
\hline
\end{tabular}
\caption{We compare machine text with the paired human text (given as an example in the prompt that generated the machine text). We select machine texts whose formality label differs from the paired human text, and examine qualitative features that may contribute to this difference. Values given are percentages over the 110 pairs we examine.}
\label{tab:free_linguistic_features}
\end{table}

\begin{table}[h]
\centering
\begin{tabular}{|c|c|c|c|c|c|}
\hline
\multirow{2}{*}{Category} & \multicolumn{2}{c|}{\textsc{Workers}} & \multicolumn{2}{c|}{All Datasets} \\
\cline{2-5}
& Human & Machine & Human & Machine \\
\hline
\hline
Formality & 63.6 & 91.1 & 83.8 & 97.8 \\
Joy & 17.5 & 27.3 & 10.2 & 8.3 \\
Neutral & 25.3 & 29.4 & 44.9 & 50.2 \\
Metaphor & 9.7 & 13.6 & 9.7 & 9.5 \\
Sadness & 8.4 & 5.8 & 4.6 & 4.7 \\
Irony & 23.4 & 23.3 & 13.3 & 8.2 \\
Fear & 7.2 & 6.9 & 5.0 & 6.7 \\
\hline
\end{tabular}
\caption{Percentages of texts, written by human and machine, classified as containing various high-level attributes. We focus on the Social Dimensions subset of \textsc{Workers}, which contains texts from social media labeled with various social relationships, in most of our experiments, but also report the overall results across all three datasets we examine.}
\label{tab:attribute_scores}
\end{table}

\paragraph{High Level Attributes}
Figure~\ref{fig:high_level_features_radar} shows a summary of our results for high-level attributes on human vs machine text. 

\paragraph{Qualitative analysis} We examine 110 human-machine text pairs that have differing formality labels to see what qualities the texts have that may lead to this difference in formality. (We focus on the formality attribute since it is the most consistently different between human-machine text pairs.) Example text pairs can be found in Table~\ref{table:formality_qual}.

In our qualitative study, we see that $56\%$ of the machine texts include complex sentences and more independent clauses than their human counterparts. $52\%$ of the machine texts include full punctuation (e.g. completing sentences with periods, inserting commas between clauses) when the human texts do not. These factors likely contribute to the differing formality scores, but some differences extend past this surface level: $21\%$ of the machine texts have no discourse markers although the human text does, and machine texts also frequently use more intensifiers, adjectives, and emojis than their human counterparts. $27\%$ of the machine text uses more figurative language as well.

\begin{figure*}[ht!]
  \centering
  \includegraphics[width=\textwidth]{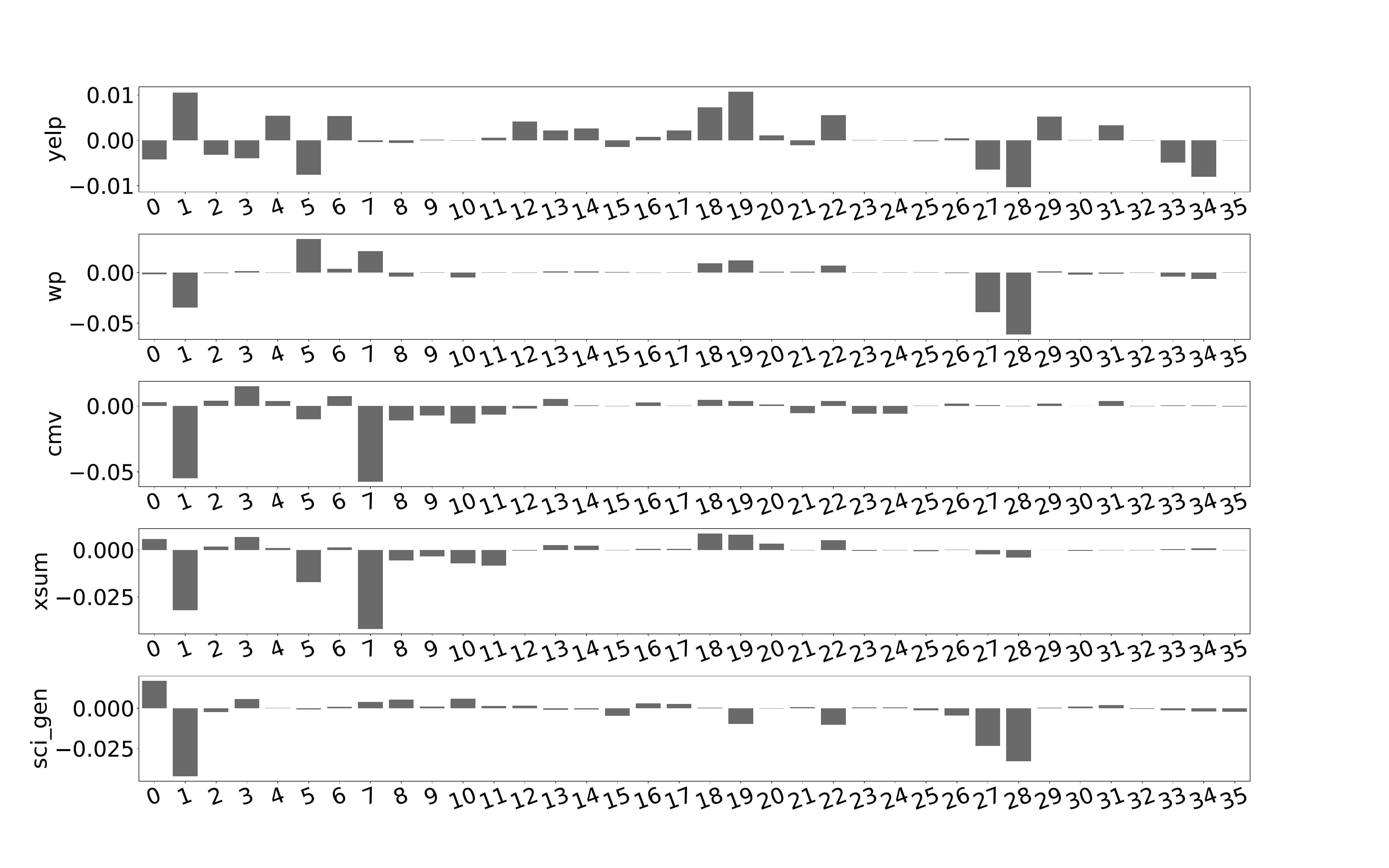}
  \caption{Distributional difference between machine-generated vs. human-authored document graphs (motifs) for five domains---user reviews (``yelp''), WritingPrompts subreddit (``wp''), ChangeMyView subreddit (``cmv''), news articles (``xsum''), and scientific writing (``sci\_gen''). Positive scores signify that the identified motives are more prominent in machine-generated texts, whereas negative scores suggest the opposite trend.}
  \label{fig:free_text/dist_diff_five_domains}
\end{figure*}

\begin{figure}[ht!]
    \centering
    \includegraphics[width=0.5\columnwidth]{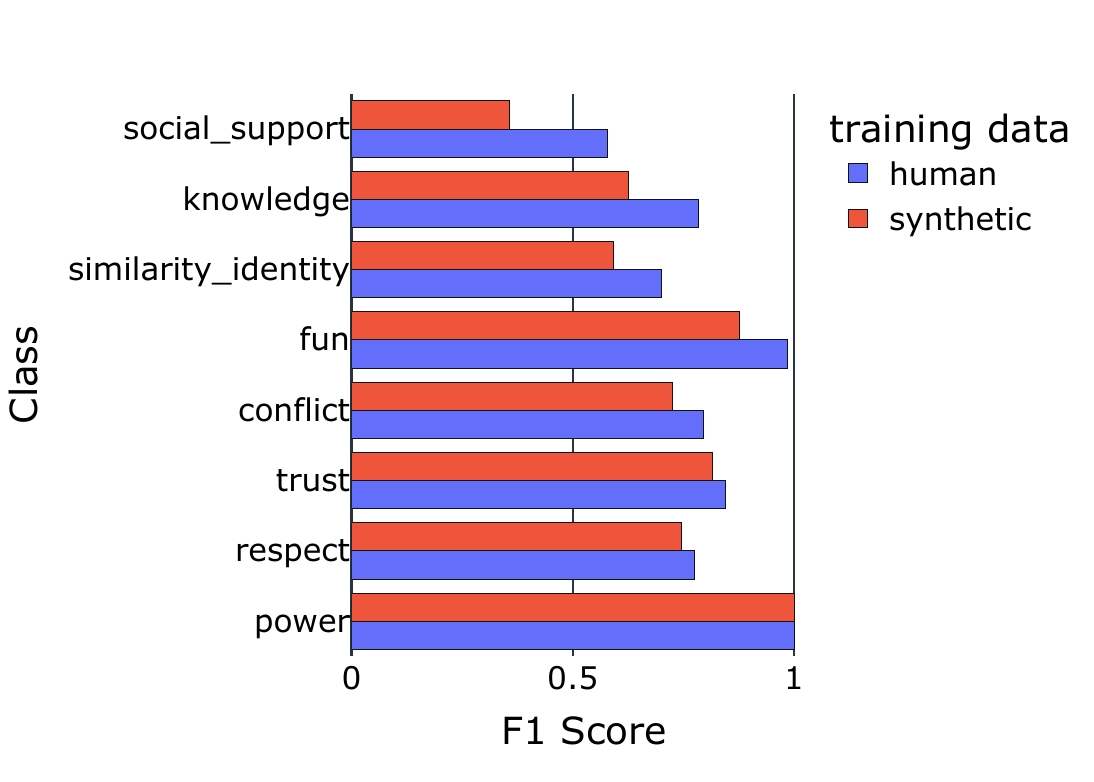}
    \caption{Certain labels drive subpar performance in model trained on artificial data.}
    \label{fig:social-2nd-order}
\end{figure}


\section{Task Labels}
Our investigation examines the potential bias of LLMs towards majority viewpoints in annotated datasets, risking the under-representation of minority opinions and leading to a lack of diversity in responses, particularly in subjective and complex decision-making contexts. We also trained a classifier on human and machine-generated annotations to investigate second-order effects and biases, hypothesizing that these biases would become more evident over time as the LLM's performance distribution becomes more concentrated, highlighting specific stereotypes and assumptions. 

\subsection{First Order Experiments}
\label{sec:app-task_label_first}


\begin{listing}[!ht]
\begin{minted}[fontsize=\footnotesize, frame=single, breaklines]{python}
f'''
For this task, we want to study the pragmatic frames in which people project social biases and stereotypes onto others.

As a responsible Mechanical Turk worker, your task is to respond to the given problem in a concise manner as specified. Your opinion should match that of a randomly chosen individual from a diverse group of people representing all kinds of backgrounds, political opinions, socioeconomic statuses, and so on. You may respond using your persona's personal biases and opinions. Please carefully follow the instructions below and label the given sentence using the provided options.

### Sentence: How then do you explain that the first Rus rulers of Kiev and Novgorod had Scandinavian names ?

### Label options: (0) No offensive, (1) maybe, (2) offensive

Among the given options, which do you think is the most appropriate option?\\

Among the given options, I think the most appropriate option is (
'''
\end{minted}
\caption{Prompt for LLM task label generation.}
\label{lst:prompt_task}
\end{listing}

    
    
\begin{figure}[htbp!]
  \centering
\includegraphics[width = 0.7\linewidth,clip,trim=0cm 0cm 0cm 0cm]{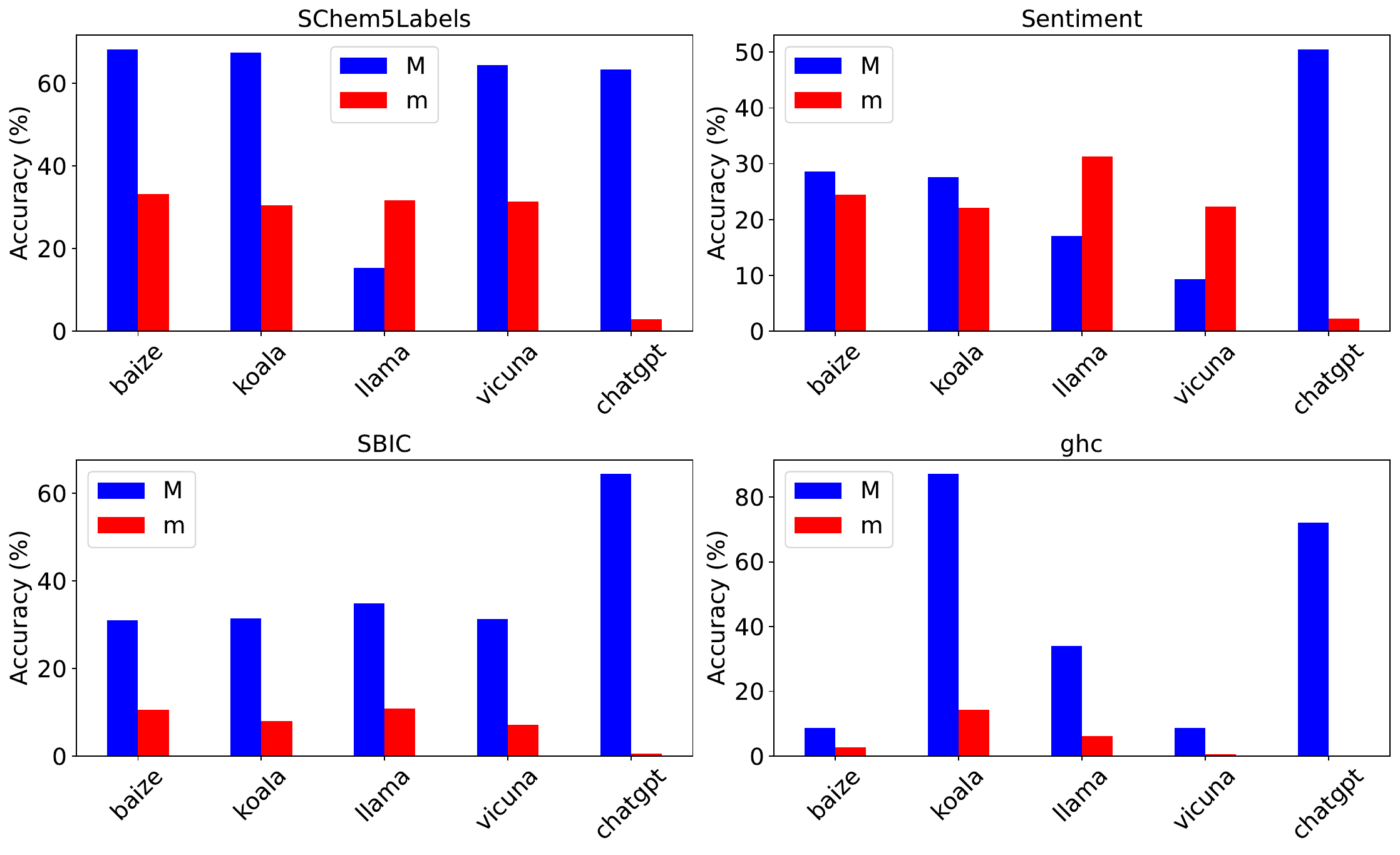}
\vspace{-5mm}
  \caption{Model accuracy for majority (M) and minority (m) match comparison on all datasets. \textit{Because ChatGPT has the highest majority match accuracy and the lowest minority match accuracy across all datasets, minority annotations tend to be represented inadequately.}}  
  \label{fig:maj_match}
\end{figure}
\begin{figure}[htbp!]
    \centering
    \begin{minipage}{0.45\textwidth}
        \centering
        \includegraphics[width=\textwidth]{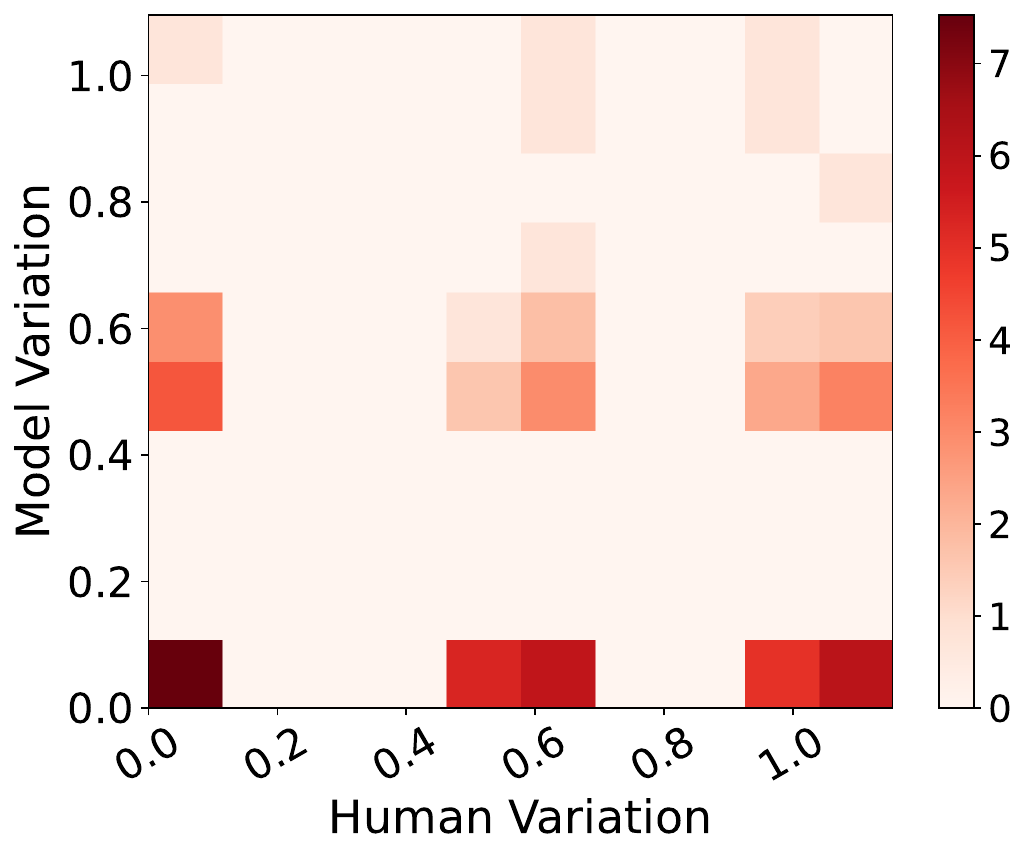}
        
    \end{minipage}\hfill
    \begin{minipage}{0.45\textwidth}
        \centering
        \includegraphics[width=\textwidth]{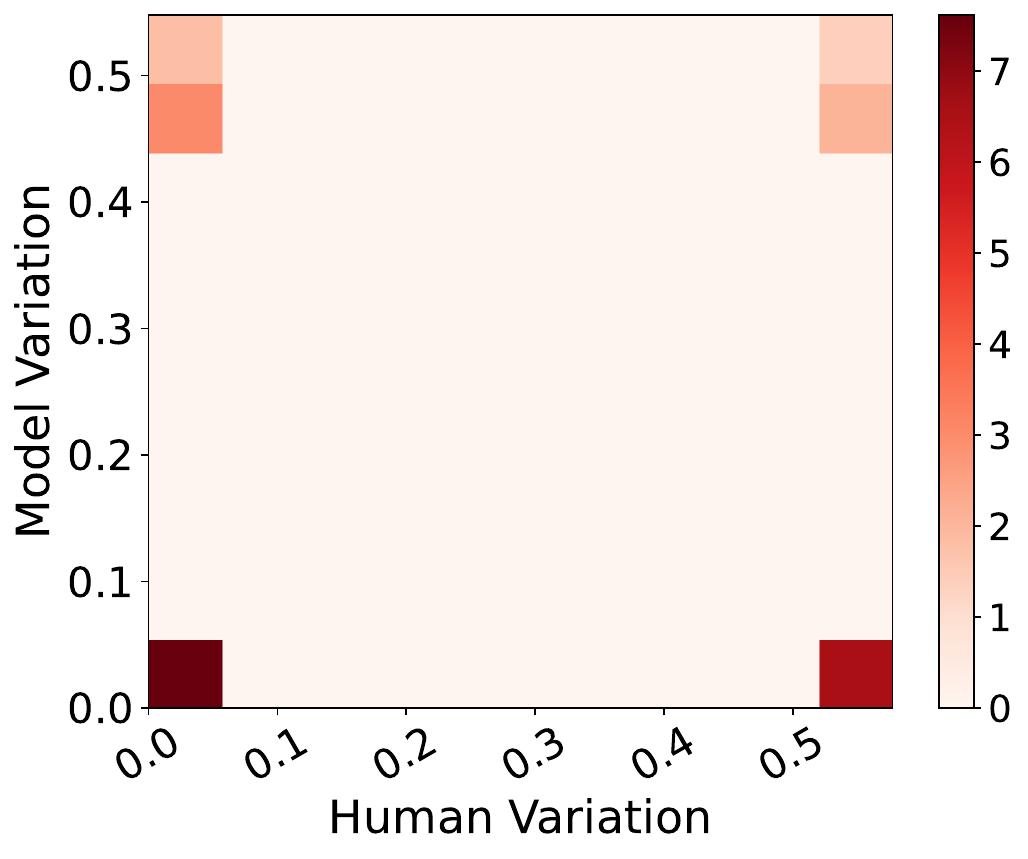}
        
    \end{minipage}
    \caption{Human vs Model variation for SBIC(left) and GHC(right) datasets.}
    \label{fig:sbic_ghc_variation}
\end{figure}

\paragraph {Additional 1st Order Results}
Figure \ref{fig:maj_match} shows the results comparing majority and minority match accuracy for all datasets considered. We observe that across all datasets, ChatGPT tends to represent the majority over and under-represent minority labels. 
Figure \ref{fig:sbic_ghc_variation} shows the results comparing human and model (ChatGPT's intra-model setup) variation for SBIC and GHC datasets. High variation is less common in ChatGPT's annotation group than in the human annotation group, suggesting that ChatGPT may have a more constrained response pattern or is better at mimicking a consensus among human annotations.

\subsection{Second-order Experiments}
\begin{figure}[!htb]
    \centering
    \includegraphics[scale=0.25]{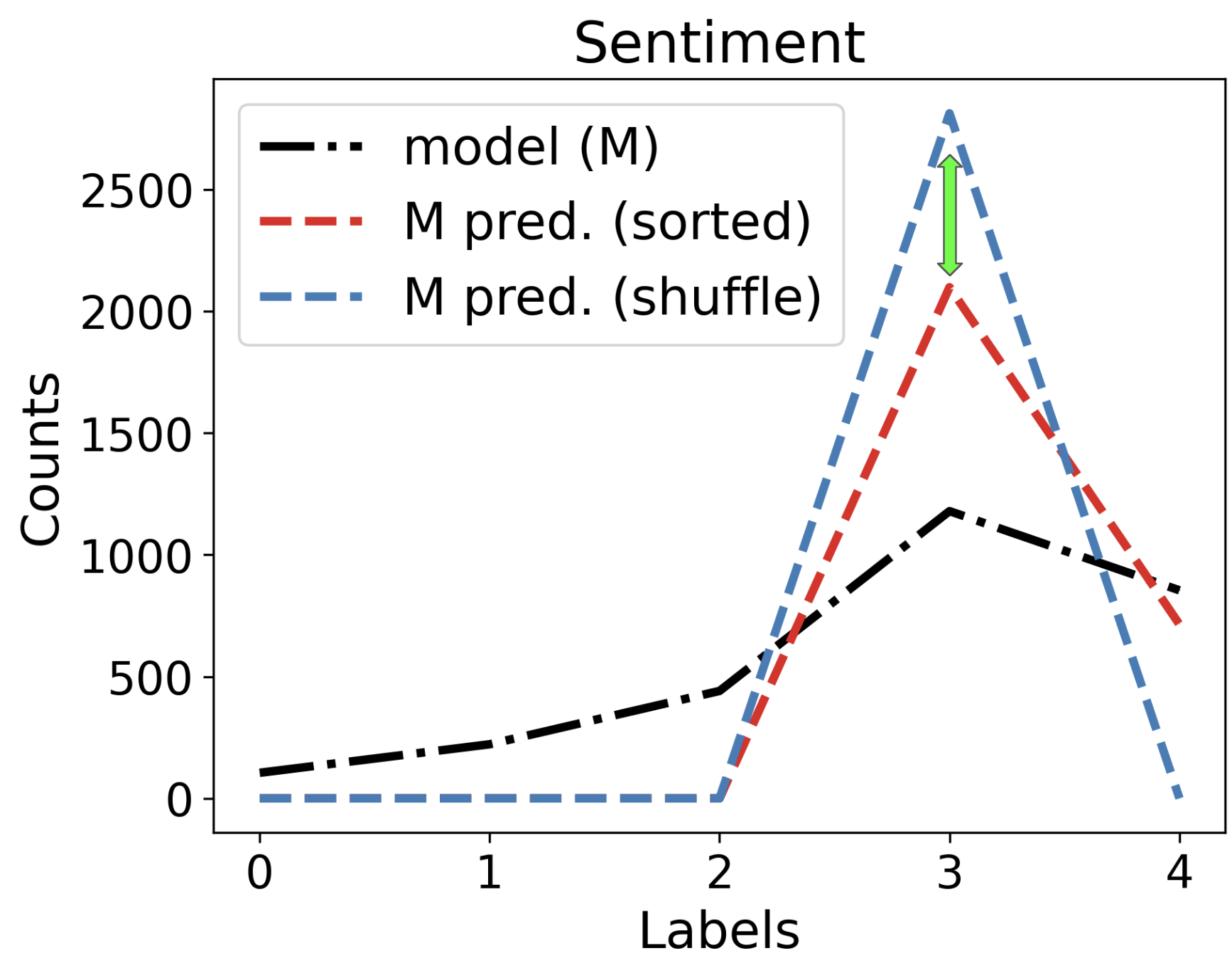}
    \caption{We observe how the order of each set of labels in the training data affects the prediction labels in the case of the Sentiment dataset. We immediately notice that the model trained on shuffled labels displays a higher prediction skew than the model trained of sorted labels.}
    \label{fig:sentiment_model_only_all_modes}
\end{figure}

\begin{figure}[!htb]
    \centering
    \includegraphics[scale=0.25]{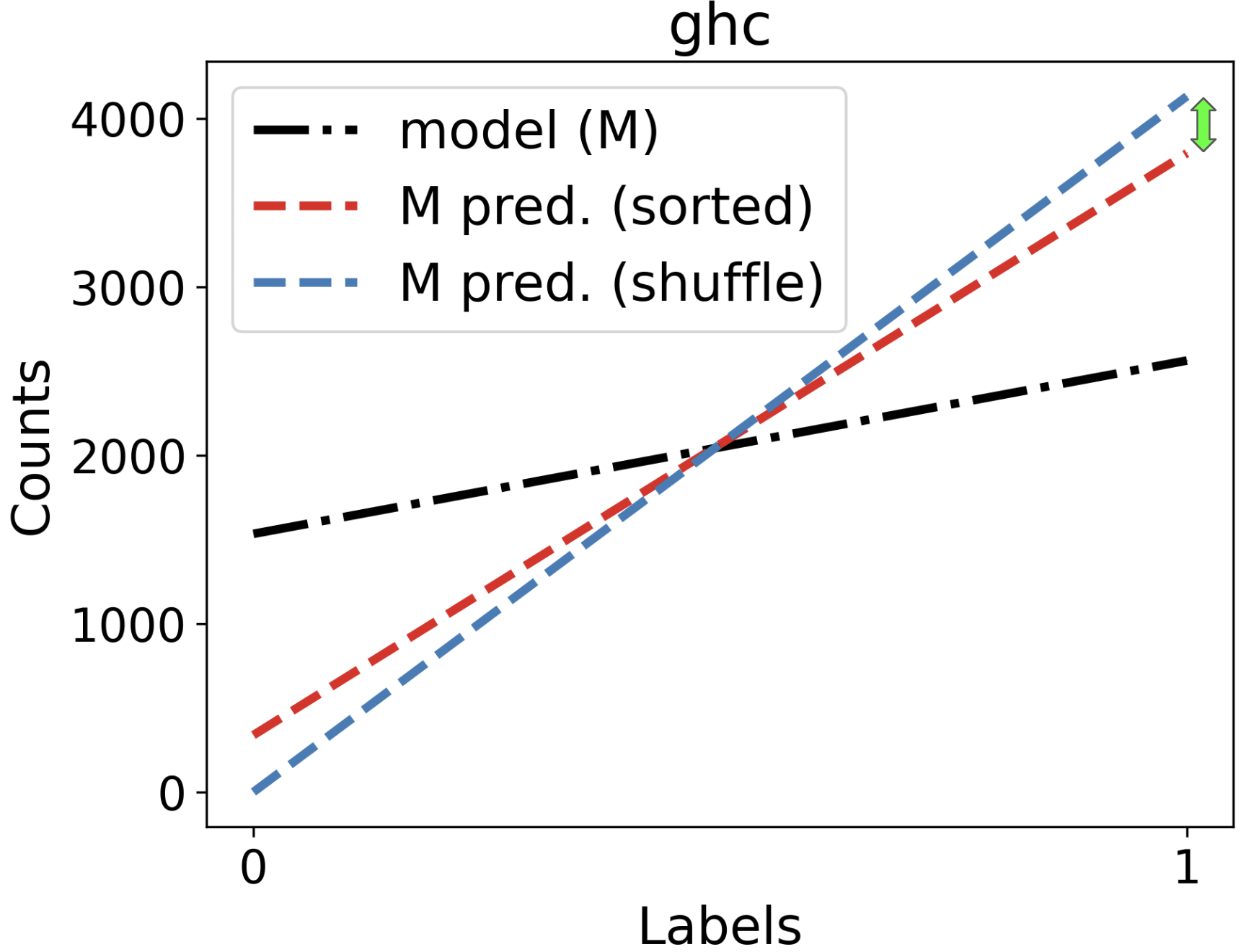}
    \caption{We observe how the order of each set of labels in the training data affects the prediction labels in the case of the ghc dataset. We immediately notice that the model trained on shuffled labels yields the most skewed prediction labels here as well. The predictions of the instance-frequency-trained model almost entirely overlap with that of the shuffled model, due to the binary nature of the dataset and the fact that there were fewer annotators. }
    \label{fig:ghc_model_only_all_modes}
\end{figure}

In addition to observing how minority opinions were further underestimated in the second order, we experimented with how the label order of the training data affects this behavior. Here, we focus specifically on the difference between the shuffled and sorted (smallest to largest) variations. We hypothesize that ordering each set of labels in a specific way will decrease the distribution difference between the gold and predicted label since each classification head on the RoBERTa-base model will only be required to guess a specific range of possible labels. The first classification head will only need to predict the smallest gold label, the second classification head the second-smallest gold label, and so on. In contrast, the model trained on shuffled labels will have to ensure that the distribution of labels individually chosen by each classification head is consistent with the distribution of the gold labels. We find that datasets with a much higher most-common to the second-most-common ratio (See Table \ref{tab:ratios}) show no difference between the sorted and shuffled training performance. This is because the skew creates an environment where it's beneficial for the model to continue guessing the most frequent label, even if it means lowering the accuracy for minority options. However, we see the beneficial effects of the different orderings in Figures \ref{fig:sentiment_model_only_all_modes} and \ref{fig:ghc_model_only_all_modes}.



\begin{table}[ht]
    \centering
    \caption{We show the count ratio between the most common label and the next common one. For example, a ratio of 1.69 indicates that the count of the most frequent label was 1.69 times that of the next. We see especially high values for SChem5Labels and SBIC, which is consistent with how the 2nd-order labels for these two datasets congregated towards the most frequent label.}
    \begin{tabular}{lcc}
        \toprule
        & \textbf{Human} & \textbf{Inter} \\
        \midrule
        Sentiment & 1.69 & 1.38 \\
        SChem5Labels & 1.9 & \cellcolor{red!30}3.45 \\
        ghc & 6.72 & 1.67 \\
        SBIC & 1.64 & \cellcolor{red!30}4.06 \\
        \bottomrule
    \end{tabular}
    \label{tab:ratios}
\end{table}

These results indicate that given our experiment setup, ordering the labels in the training data causes the least amount of deviation from the gold labels. If training a model using a second-order dataset is inevitable, this additional processing step may be beneficial if the gold labels are not extremely skewed.

\section{Instructions}\label{sec:appendix:instr}
\subsection{First Order Experiments}
Our first-order analysis involved annotating artificial instructions with their instruction type (\textsc{Open-QA} and \textsc{Closed-QA}) and errors, of which there are four types: \textsc{Incomprehensible Instruction}, \textsc{Inconsistent Input}, \textsc{Inconsistent Output}, and \textsc{Incorrect Output}. Table \ref{tab:instruction-error-examples-complete} provides an example of each error type from our error analysis for both Open-QA and Closed-QA instruction types.
\begin{table}[htbp!]
    \centering
    \small
    \begin{tabular}{p{.16\textwidth}p{.102\textwidth}p{.66\textwidth}}\toprule
        \textbf{Error Type} & \textbf{Task Type} & \textbf{Instruction Triple} \\\midrule
    \multirow{2}{.16\textwidth}{Incomprehensible Instruction} & Closed & \textbf{Instruction:} Describe the figure in the following photo: \textbf{Input:} <Photo> \textbf{Output:} I'm sorry, but I am an AI language model and do not have the capability to view or describe photos. Can you provide me with a description or more information about the figure in the photo so that I can help you better? - \textsc{Cleaned Alpaca}\\
    & Open & \textbf{Instruction:} Find a person who is similar to you in one of the following: hobbies, character traits, or occupation. describe this person and explain why he/she is similar to you. \textbf{Input:} None \textbf{Output:} I am a big fan of [Sport] and I have been playing it since I was a kid. My favorite player is [Player]. He is also a big fan of [Sport], just like me. We both love the game, and we are both very good at it. In fact, he is one of the best players in the world. I think that’s why we are similar. - \textsc{Self Instruct}\\\midrule
    \multirow{2}{.16\textwidth}{Inconsistent Input} & Closed & \textbf{Instruction:} Find the sentence or sentences that refute or disprove a given statement. \textbf{Input:} The Amazon rainforest is the world's largest rainforest and is home to a wealth of biodiversity. \textbf{Output:} I cannot generate a response as no refutal sentences were provided as input for me to choose from. - \textsc{GPT-4-LLM}\\
    & Open & \textbf{Instruction:} Describe a situation when someone was asking you questions and you were unable to answer them. \textbf{Input:} Question: What is the capital of France? \textbf{Output:} I don't know. - \textsc{Self Instruct}\\\midrule
    \multirow{2}{.16\textwidth}{Inconsistent Output} & Closed & \textbf{Instruction:} Identify which countries border with each other. output one country per line. \textbf{Input:} Country 1: Canada
Country 2: United States \textbf{Output:} (a) - \textsc{Self Instruct}\\
    & Open & \textbf{Instruction:} Design a flag for mars. include at least one martian symbol on it. \textbf{Input:} None \textbf{Output:} The flag of Mars is a tricolor with three equal horizontal bands of red (top), white, and blue. The design was originally created in 1958 by Russian scientist Victor S. Mamatov, who worked at the Institute of Space Device Engineering. It was adopted as the official flag of the planet in 1960 by the International Astronomical Union (IAU) through its Resolution 2. - \textsc{Self Instruct}\\\midrule
    \multirow{2}{.16\textwidth}{Incorrect Output} & Closed & \textbf{Instruction:} You will be given a piece of text describing a scene. Your goal is to output whether the events in the text are able to happen or not, based on your commonsense. \textbf{Input:} Text: The Keys were sitting on top of the car while we went swimming.\textbf{Output:} No - \textsc{Unnatural}\\
    & Open & \textbf{Instruction:} Describe an object using all the five senses. \textbf{Input:} Object: A red rose. \textbf{Output:} The flower is about 5 inches tall and has a diameter of 2 inches. It has five petals, each with a length of 0.5 inch and a width of 0.1 inch. The petals are arranged in a spiral pattern starting from the bottom. The stem is green and has a diameter of 0.2 inches. The flower smells like roses and has a soft texture. - \textsc{Self Instruct} \\\bottomrule
    
    \end{tabular}
    \caption{Examples of errors found in all four artificial datasets.}
    \label{tab:instruction-error-examples-complete}
\end{table}
\section{Preference} \label{sec:appendix:pref}

\subsection{First-order Experiments}

\begin{figure}[ht!]
    \centering
    \includegraphics[width=0.6\textwidth]{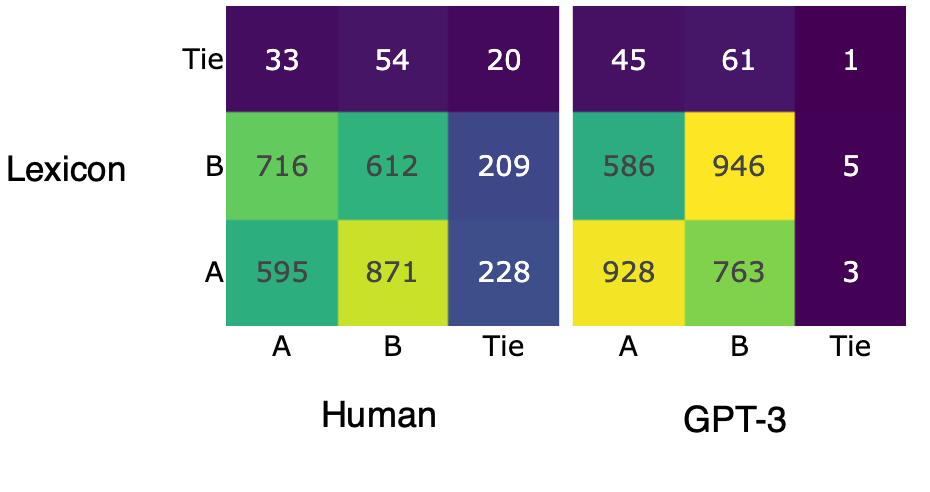}
    \caption{The heatmap of lexicon-based preferences compared to human and GPT-3 preferences for \textsc{p2c} dataset.} 
    \label{fig:pref:p2c-lexicon-heatmap}
\end{figure}

\paragraph{Additional 1st Order Results} Figure \ref{fig:pref:p2c-lexicon-heatmap} shows the heatmap comparing lexicon-based preferences with human and GPT-3 counterparts for \textsc{p2c} dataset. For preference labels 'A' or 'B' (non-tie), we see a strong relationship between lexicon-based and GPT-3 preferences that doesn't exist on the human side.

\subsection{Second-order Experiments} 


\paragraph{Additional 2nd Order Results} 

\begin{figure}[ht!]
    \centering
    \begin{subfigure}{0.32\textwidth}
        \caption{\textsc{Class 0}}
         \centering
         \includegraphics[width=\textwidth]{images/pref_eval/cobbler_human_v_machine_label0.pdf}
         \label{fig:pref:cobbler-second-label0}
     \end{subfigure}
         \begin{subfigure}[b]{0.32\textwidth}
         \caption{\textsc{Class 1}}
         \centering
         \includegraphics[width=\textwidth]{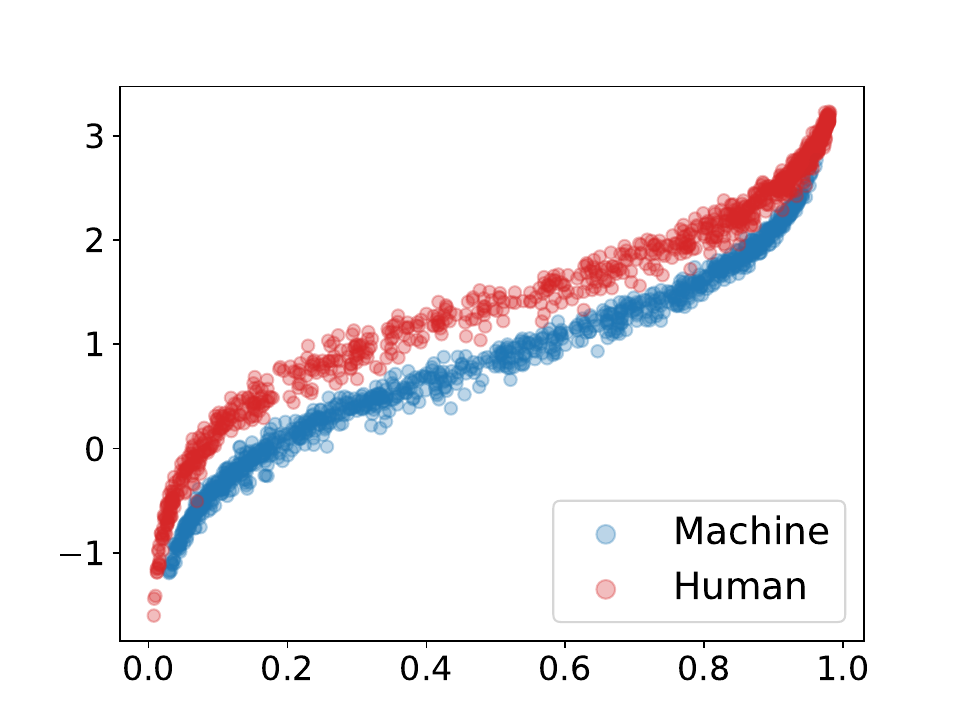}
         \label{fig:pref:cobbler-second-label1}
     \end{subfigure}
     \begin{subfigure}[b]{0.32\textwidth}
     \caption{\textsc{Class 2}}
     \centering
     \includegraphics[width=\textwidth]{images/pref_eval/cobbler_human_v_machine_label2.pdf}
     \label{fig:pref:cobbler-second-label2}
     \end{subfigure}
    \caption{The logodds vs. class probability curve from the reward model trained on \textcolor{red}{human} or \textcolor{blue}{machine} preferences from the \textsc{CoBBLEr} dataset. We view the curve on (a) class label 0, which indicates a preference for the first model's output; (b) class label 1 for the second model's output; and (c) class label 2, which indicates tie (no preference).}
    \label{fig:pref:cobbler-second}
\end{figure}

\begin{figure}[ht!]
    \centering
    \begin{subfigure}{0.32\textwidth}
        \caption{\textsc{Class 0}}
         \centering
         \includegraphics[width=\textwidth]{images/pref_eval/p2c_human_v_machine_label0.pdf}
         \label{fig:pref:p2c-second-label0}
     \end{subfigure}
         \begin{subfigure}[b]{0.32\textwidth}
         \caption{\textsc{Class 1}}
         \centering
         \includegraphics[width=\textwidth]{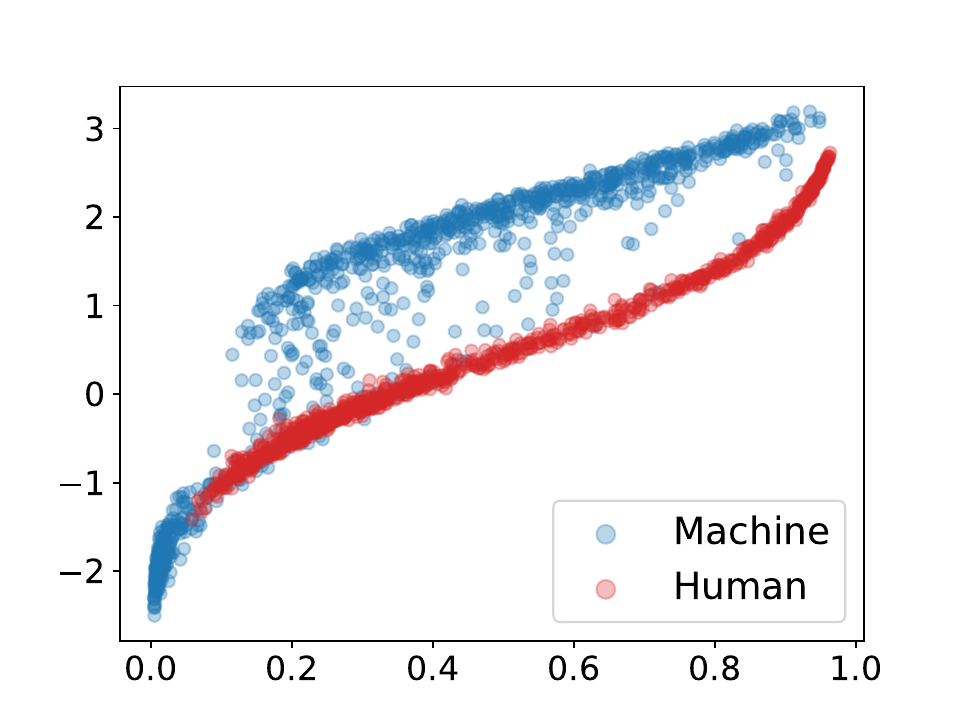}
         \label{fig:pref:p2c-second-label1}
     \end{subfigure}
     \begin{subfigure}[b]{0.32\textwidth}
     \caption{\textsc{Class 2}}
     \centering
     \includegraphics[width=\textwidth]{images/pref_eval/p2c_human_v_machine_label2.pdf}
     \label{fig:pref:p2c-second-label2}
     \end{subfigure}
    \caption{The logodds vs. class probability curve from the reward model trained on \textcolor{red}{human} or \textcolor{blue}{machine} preferences from the \textsc{p2c} dataset. We view the curve on (a) class label 0 which indicates preference for the first sentence; (b) class label 1 for the second sentence; and (c) class label 2 which indicates tie (no preference).}
    \label{fig:pref:p2c-second}
\end{figure}

Figures \ref{fig:pref:cobbler-second} and \ref{fig:pref:p2c-second} represent the logodds vs class probabilities curve from RoBERTa reward model trained on human or LLM preferences from \textsc{Cobbler} and \textsc{p2c} datasets, respectively. 

\newpage

\section{Simulation} \label{sec:appendix:simu}
\subsection{First Order Experiments}
\noindent\paragraph{Experimental Setup}
Regarding \textit{role-flipping}, we detect the string "Instruction:" and "Solution:" in the message. If "Instruction:" appears in AI-User's message or "Solution:" appears in AI-Assistant's message, we consider there is no role-flipping. If "Solution:" appears in AI-User's message or "Instruction:" appears in AI-Assistant's message, we consider role-flipping happens. If one message contains both, we consider there is no role-flipping, and if one message contains none of them, we consider it an interruption message. In the CAMEL framework, detecting role-flipping relies on identifying specific strings like "Instruction:" or "Solution:" within messages since agents are expected to follow this format; however, this check is flexible enough to allow for additional elements like brief responses or stop tokens without disrupting the conversation flow. 
Conversations deviating from this pattern are marked as ``Interruption Messages,'' often characterized by irrelevant content such as apologies or non-constructive praise, and our analysis includes comparing the length distribution of conversations with role-flipping to the entire dataset, as well as exploring the relationship between role-flipped and interruption messages in simulations.

Detecting digressions poses a more complex challenge, requiring both LLM and human evaluations. 
The LLM evaluation involves querying an LLM, specifically \texttt{GPT-4-Turbo}, about the occurrence of digressions through prompts. 
Similarly, human evaluation entails asking annotators to assess digressions using a template comparable to the LLM's prompt. We mainly conduct the experiments by prompting GPT-4. 
Here, we used the specifically designed prompts for each purpose: (1) detecting digression, (2) measuring the collaborative effect, (3) classifying given conversation into human-like and agent-like ones, (4) classifying the reasoning of digression into one of the multiple categories.
All the prompts are presented in Listings \ref{lst:prompt_detect}, \ref{lst:prompt_collabo}, \ref{lst:prompt_classify}, and \ref{lst:prompt_classify_reason}.

\begin{listing}[!ht]
\begin{minted}[fontsize=\footnotesize, frame=single, breaklines]{python}
f'''
Please analyze the following conversation for any digressions and answer 'yes' or 'no'."

Then, highlight any statements that do not directly contribute to solving the puzzle or are unrelated to the clues provided.

### Input: {example} 
### Conversation: {response}
### Answer: {label}

Then, explicitly highlight any statements that do not directly contribute to solving the puzzle or are unrelated to the clues provided. Also, providing the rationale why such statements are considered as digression. Or you can change the asnwer, but please also provide the reason for such change and provide the changed answer at the end, e.g., '... Therefore, the answer is xx.'
'''
\end{minted}
\caption{Prompt to detect digression.}
\label{lst:prompt_detect}
\end{listing}
\begin{listing}[!ht]
\begin{minted}[fontsize=\footnotesize, frame=single, breaklines]{python}
f'''
Please analyze the provided conversation to determine the collaborative dynamics among the participants which means that evaluate whether the final conclusion is a result of a combined effort, with each participant contributing significantly, or if it is predominantly led by a single participant. Please answer 'yes' if there is combined effort or 'no' otherwise.

### Input: {example}
### Conversation: {response}
### Answer: {label}

Then, providing the explicit rationale for the answer. Or you can change the asnwer, but please also provide the reason for such change and provide the changed answer at the end, e.g., '... Therefore, the answer is xx.'
'''
\end{minted}
\caption{Prompt to measure the collaborative effect between multiple agents.}
\label{lst:prompt_collabo}
\end{listing}
\begin{listing}[!ht]
\begin{minted}[fontsize=\footnotesize, frame=single, breaklines]{python}
f'''
Please analyze the provided conversation and assess whether it resembles a dialogue between human participants or appears to be an artificial conversation created by multiple AI agents. Please answer 'yes' if it is closed to human conversation or 'no' otherwise.

Consider factors such as the flow of the conversation, the naturalness of responses, the presence of emotional or personal interjections, and the complexity or variability of language used.

### Conversation: {response}
### Answer: {label}

Then, providing the explicit rationale for the answer by highlight any characteristics that particularly suggest a human or artificial origin of the conversation. Or you can change the asnwer, but please also provide the reason for such change and provide the changed answer at the end, e.g., '... Therefore, the answer is xx.'.
'''
\end{minted}
\caption{Prompt to classify the conversations into human-like or agent-like ones.}
\label{lst:prompt_classify}
\end{listing}
\begin{listing}[!ht]
\begin{minted}[fontsize=\footnotesize, frame=single, breaklines]{python}
f'''
Please classify the given reasoning into the one of the following categories: [1] Procedural or Introductory Statements, [2] Repetitive or Confirmatory Statements, [3] Transitional Statements, [4] Irrelevant or Off-Topic Contributions, [5] Misleading or Incorrect Reasoning, [6] Closing or Summarizing Remarks, and [7] Constructive Contributions. You response should be in the form of [], e.g., [1] or [3].

Then, highlight any statements that do not directly contribute to solving the puzzle or are unrelated to the clues provided.

### Conversation: {response}
### Reasonings: {reason}
### Answer (You response should be in the simple form, e.g., [1] or [3]): " 
'''
\end{minted}
\caption{Prompt to classify the reasoning about the digression into one of the specific categories.}
\label{lst:prompt_classify_reason}
\end{listing}

\newpage
\noindent\paragraph{Additional 1st Order Results}
In Listings \ref{lst:example_role_flip1} and \ref{lst:example_role_flip2}, we present two specific examples of role-flipping. 
Also, we present two specific examples of digression in Listings \ref{lst:example_digression1} and \ref{lst:example_digression2}.

\paragraph{[1] Correlation between Role-Flipping and Interruption Messages} 
Our study of the CAMEL AI-World dataset identified "interruption messages" in simulated conversations, which diverge from the expected dialogue format, often beginning with apologies or shifting to simple responses like praise. These interruptions, found in 26.12\%  of conversations, often precede "role-flipping" which occurs in 8.22\% of conversations with interruptions, a rate that is higher than in the overall dataset (2.94\%). Notably, in 88.8\% of cases with both interruption and role-flipping, the first interruption comes before role-flipping, suggesting that interruptions might trigger this behavior, particularly as role-flipping is exclusively observed with the AI-User, indicating that confusion caused by the AI-Assistant's responses could be a significant factor in subsequent role-flipping.

\paragraph{[2] Calculation of task accuracy} 
When we calculate the accuracy of a given task, we directly compare the prediction in the conversation and the ground-truth label in the dataset, following the original work \citep{wang2023unleashing}.
For example, in the case of the example in Listing \ref{lst:example_digression1}, the prediction by conversation is \textit{House 3} while the given label is \textit{House 2}. Then, it is considered as wrong prediction.

\begin{listing}[!ht]
\begin{minted}[fontsize=\footnotesize, frame=single, breaklines]{python}
f'''
### {Original_task}
Actor can assist Accountant with preparing financial reports and statements.

### {Specified_task}
Actor will relieve the accountant of data collection by retrieving bank and credit card statements as well as entering expenses. They also cross-check debits and credits, record adjustments, input any transactions needed for accurate statements and breakdown analysis or organizational format for accountants to assess movements pivotal for decision-making easily.

### {Roles}
- User: Accountant, Assistant: Actor

### {Simulated Conversations}
(1) User - Instruction: Retrieve the bank statements for the past three months. Input: None
(2) Assistant - Solution: I will retrieve the bank statements for the past three months and provide them to you in a secure manner. Next request.
(3) User - Instruction: Retrieve the credit card statements for the past three months.
Input: None
(4) Assistant - Solution: I will retrieve the credit card statements for the past three months and provide them to you in a secure manner. Next request.
(5) User - Instruction: Please enter all the expenses from the bank and credit card statements into the accounting software.
Input: Bank and credit card statements.
(6) Assistant - Solution: I will enter all the expenses from the bank and credit card statements into the accounting software. I will ensure that all expenses are accurately recorded and categorized according to the appropriate expense account. Next request.
(7) User - I apologize, but I believe there was an error in the previous messages. Could you please repeat the last instruction?
(8) Assistant - Sure, the last instruction was: "Please enter all the expenses from the bank and credit card statements into the accounting software." The input was "Bank and credit card statements."
(9) User - Solution: I will enter all the expenses from the bank and credit card statements into the accounting software. I will ensure that all expenses are accurately recorded and categorized according to the appropriate expense account. Please provide me with any additional transactions that need to be entered into the accounting software.
(10) Assistant - Next request, please.
'''
\end{minted}
\caption{Example of role-flipping.}
\label{lst:example_role_flip1}
\end{listing}
\begin{listing}[!ht]
\begin{minted}[fontsize=\footnotesize, frame=single, breaklines]{python}
f'''
### {Original_task}
Hosting Q&A sessions on social media to answer questions about animals and their behavior.

### {Specified_task}
Research and propose a list of ten frequently asked topics related to animals and their behavior, that could make an engaging Q&A session for social media followers. In addition, determine three potential guest speakers who can share their expertise on those topics, and suggest methods for incorporating fan feedback into the session design.

### {Roles}
- User: Social media manager, Assistant: Zoologist

### {Simulated Conversations}
(1) User - Instruction: Research and propose a list of ten frequently asked topics related to animals and their behavior.
Input: None
(2) Assistant - Solution: As a zoologist, I suggest the following ten frequently asked topics related to animals and their behavior: 
1. Animal communication 2. Mating behaviors 3. Social structures 4. Migration patterns 5. Predator-prey relationships 6. Animal intelligence 7. Animal emotions 8. Animal parenting 9. Animal adaptations 10. Animal conservation. 
These topics are commonly asked about by animal enthusiasts and can provide a great starting point for a Q&A session. Next request.
(3) User - Instruction: Determine three potential guest speakers who can share their expertise on those topics.
Input: None
(4) Assistant - Solution: Based on my expertise, I suggest the following three potential guest speakers who can share their expertise on the ten topics related to animals and their behavior: 
1. Dr. Jane Goodall, who is an expert on animal communication and conservation. 
2. Dr. Frans de Waal, who is an expert on animal emotions and intelligence. 
3. Dr. Temple Grandin, who is an expert on animal behavior and welfare. These speakers are highly respected in their fields and can provide valuable insights and information to the Q&A session. Next request.
(5) User - Solution: Based on my expertise, I suggest the following three potential guest speakers who can share their expertise on the ten topics related to animals and their behavior: 
1. Dr. Jane Goodall, who is an expert on animal communication and conservation. 
2. Dr. Frans de Waal, who is an expert on animal emotions and intelligence. 
3. Dr. Temple Grandin, who is an expert on animal behavior and welfare. These speakers are highly respected in their fields and can provide valuable insights and information to the Q&A session. Next request.
(6) Assistant - Solution: To suggest methods for incorporating fan feedback into the session design, I recommend creating a social media poll or survey to gather input from followers on which topics they are most interested in learning about. This can help guide the selection of topics and guest speakers for the Q&A session. Additionally, incorporating a live Q&A segment during the session where fans can submit questions in real-time can help increase engagement and provide an opportunity for direct interaction with the guest speakers. Finally, encouraging fans to submit questions or feedback through social media platforms before and after the session can help create a sense of community and foster ongoing engagement. Next request.
(7) User - Instruction: Please summarize the proposed solution for the task.
Input: None
(8) Assistant - Solution: The proposed solution for the task is to research and propose a list of ten frequently asked topics related to animals and their behavior, determine three potential guest speakers who can share their expertise on those topics, and suggest methods for incorporating fan feedback into the session design. The ten topics suggested are animal communication, mating behaviors, social structures, migration patterns, predator-prey relationships, animal intelligence, animal emotions, animal parenting, animal adaptations, and animal conservation. The three potential guest speakers suggested are Dr. Jane Goodall, Dr. Frans de Waal, and Dr. Temple Grandin. Methods for incorporating fan feedback include creating a social media poll or survey, incorporating a live Q&A segment, and encouraging fans to submit questions or feedback through social media platforms.
'''
\end{minted}
\caption{Example of role-flipping.}
\label{lst:example_role_flip2}
\end{listing}
\begin{listing}[!ht]
\begin{minted}[fontsize=\footnotesize, frame=single, breaklines]{python}
f'''
### {Given task}
Q: There are 4 houses in a row, numbered 1 on the left to 4 on the right. There is one person living in each house. The people in these houses have different characteristics:
- Each person has different flowers in their foyer: one has a carnations arrangement, one has a bouquet of daffodils, one has a vase of tulips, and one has a bouquet of lilies
- Each person plays a different musical instrument: one is a guitarist, one is a pianist, one is a percussionist, and one is a flutist
 
Clue(s):
1. The flutist lives in the second house.
2. The person who has a vase of tulips lives directly left of the guitarist.
3. The person who has a bouquet of lilies lives directly left of the person who has a carnations arrangement.
4. There is one house between where the flutist lives and where the pianist lives.

What is the number of the house where the person who has a vase of tulips lives?
choice: 2, choice: 4, choice: 1, choice: 3
A:
### {Response from dataset}
Participants: AI Assistant (you); Detective; Florist; Musician

Detective: Let's analyze the clues and try to find the house number where the person who has a vase of tulips lives. We need to consider the characteristics of the people living in the houses and the given clues.
Florist: Remember that each person has different flowers in their foyer: carnations, daffodils, tulips, and lilies.
Musician: And each person plays a different musical instrument: guitar, piano, percussion, and flute.
 
AI Assistant (you): Let's start by using the first clue: The flutist lives in the second house. So, House 2 has the flutist.
Detective: Now, let's use the fourth clue: There is one house between where the flutist lives and where the pianist lives. Since the flutist lives in House 2, the pianist must live in either House 4 or House 1.
Florist: Let's use the second clue: The person who has a vase of tulips lives directly left of the guitarist. Since the tulips cannot be in House 1, they must be in House 3. Therefore, the guitarist lives in House 4.
Musician: Now that we know the guitarist lives in House 4, the pianist must live in House 1, according to the previous deductions.
Detective: Finally, let's use the third clue: The person who has a bouquet of lilies lives directly left of the person who has a carnations arrangement. Since the lilies cannot be in House 1, they must be in House 2. Therefore, the carnations arrangement is in House 3.
 
Finish collaboration!
Final answer: The house number where the person who has a vase of tulips lives is 3.
### {Prediction of Digression}
### Answer: Yes.
The following statements do not directly contribute to solving the puzzle or are unrelated to the clues provided:
- Florist: Remember that each person has different flowers in their foyer: carnations, daffodils, tulips, and lilies.
- Musician: And each person plays a different musical instrument: guitar, piano, percussion, and flute.
**Rationale:**
These statements by the Florist and the Musician are simply reiterations of the information already provided in the initial setup of the puzzle. They do not offer new insights or help in deducing the solution to the puzzle. The participants should focus on applying the clues to determine the house number rather than restating the given characteristics.
 
Therefore, the answer is 3.
'''
\end{minted}
\caption{Example of digression.}
\label{lst:example_digression1}
\end{listing}
\begin{listing}[!ht]
\begin{minted}[fontsize=\footnotesize, frame=single, breaklines]{python}
f'''
### {Given task}
Q: There are 4 houses in a row, numbered 1 on the left to 4 on the right. There is one person living in each house. The people in these houses have different characteristics:
- Each person has different pets: one owns horses, one is a cat lover, one is a dog owner, and one is a fish enthusiast
- Each person has different flowers in their foyer: one has a vase of tulips, one has a bouquet of lilies, one has a carnations arrangement, and one has a bouquet of daffodils 

Clue(s):
1. The person who has a vase of tulips lives directly left of the dog owner.
2. The person who has a bouquet of daffodils does not live in the second house.
3. There is one house between where the person who owns horses lives and where the person who has a carnations arrangement lives.
4. The person who has a bouquet of lilies lives somewhere to the right of the person who owns horses.
5. The cat lover lives in the first house. 

What is the number of the house where the fish enthusiast lives?
choice: 2, choice: 4, choice: 1, choice: 3
A:
### {Response from dataset}
Participants: AI Assistant (you); Detective; Pet Expert; Florist
 
Detective: Let's analyze the clues and try to find the house number where the fish enthusiast lives. We need to consider the characteristics of each person and their house number.
Pet Expert: We know that the cat lover lives in the first house. So, the fish enthusiast must live in one of the other three houses.
Florist: Let's use the information about the flowers to narrow down the possibilities. We know that the person with tulips lives directly left of the dog owner, and the person with daffodils does not live in the second house.
 
AI Assistant (you): Based on the clues, let's start by placing the cat lover in the first house. Since the person with tulips lives directly left of the dog owner, the dog owner must live in house 3, and the person with tulips must live in house 2. This means the person with daffodils lives in house 4. Now, we know there is one house between the person who owns horses and the person who has a carnations arrangement. Since the person with lilies lives somewhere to the right of the person who owns horses, the person who owns horses must live in house 1, and the person with lilies must live in house 4. This leaves house 3 for the person with the carnations arrangement. Since the cat lover lives in house 1, and the dog owner lives in house 3, the fish enthusiast must live in house 4.
 
Detective: Let's verify the solution. House 1 has the cat lover, horses, and no flowers mentioned. House 2 has the person with tulips and no pets mentioned. House 3 has the dog owner, the person with carnations, and no other pets mentioned. House 4 has the fish enthusiast, the person with lilies, and no other pets mentioned. Everything seems to be consistent with the clues.
 
Finish collaboration!
Final answer: The fish enthusiast lives in house number 4.
### {Prediction of Digression}
### Answer: Yes.
### Digressions in the conversation: 
1. Detective: "Let's analyze the clues and try to find the house number where the fish enthusiast lives. We need to consider the characteristics of each person and their house number."
- **Rationale**: This statement is an introduction and does not provide any specific information or analysis related to the clues.
2. Pet Expert: "We know that the cat lover lives in the first house. So, the fish enthusiast must live in one of the other three houses."
- **Rationale**: While the first sentence is a restatement of a clue, the second sentence is unnecessary as it is an obvious deduction that does not advance the solution.
3. Detective: "Let's verify the solution. House 1 has the cat lover, horses, and no flowers mentioned. House 2 has the person with tulips and no pets mentioned. House 3 has the dog owner, the person with carnations, and no other pets mentioned. House 4 has the fish enthusiast, the person with lilies, and no other pets mentioned. Everything seems to be consistent with the clues."
- **Rationale**: This statement is a summary and verification of the solution, which is not directly contributing to
'''
\end{minted}
\caption{Example of digression.}
\label{lst:example_digression2}
\end{listing}

\begin{figure}
    \centering
    \includegraphics[width=0.5\textwidth]{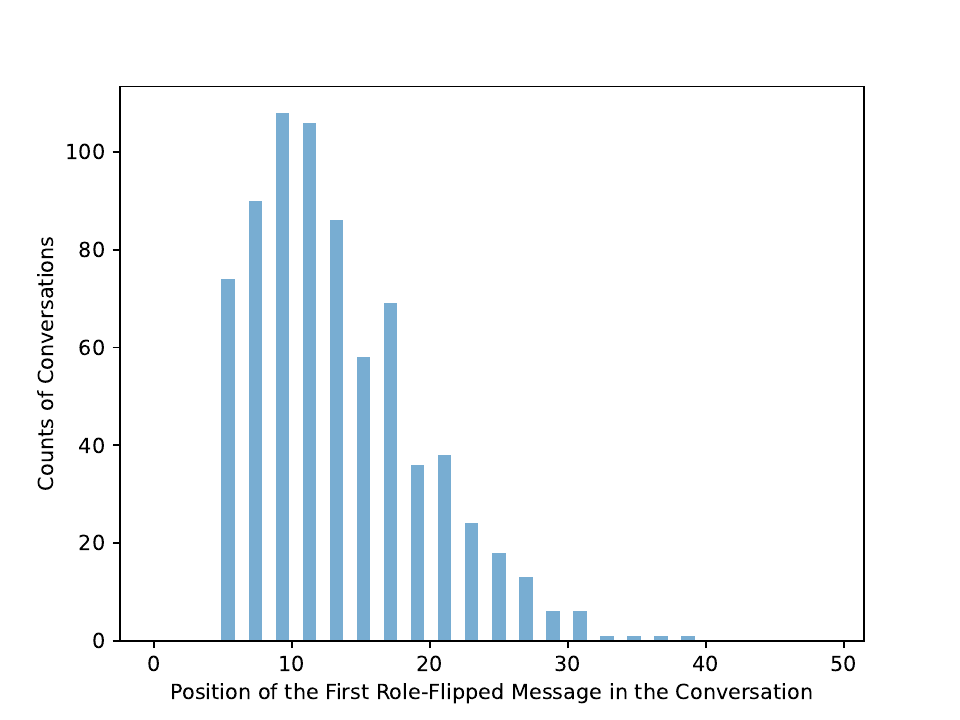}
    \caption{Position (index) of the first role-flipped message in the conversations in the CAMEL AI-World dataset.}
    \label{fig:first_flipping_distribution}
\end{figure}

\begin{figure*}[t]
    \centering
\includegraphics[width=0.9\textwidth,height=0.9\textheight,keepaspectratio]{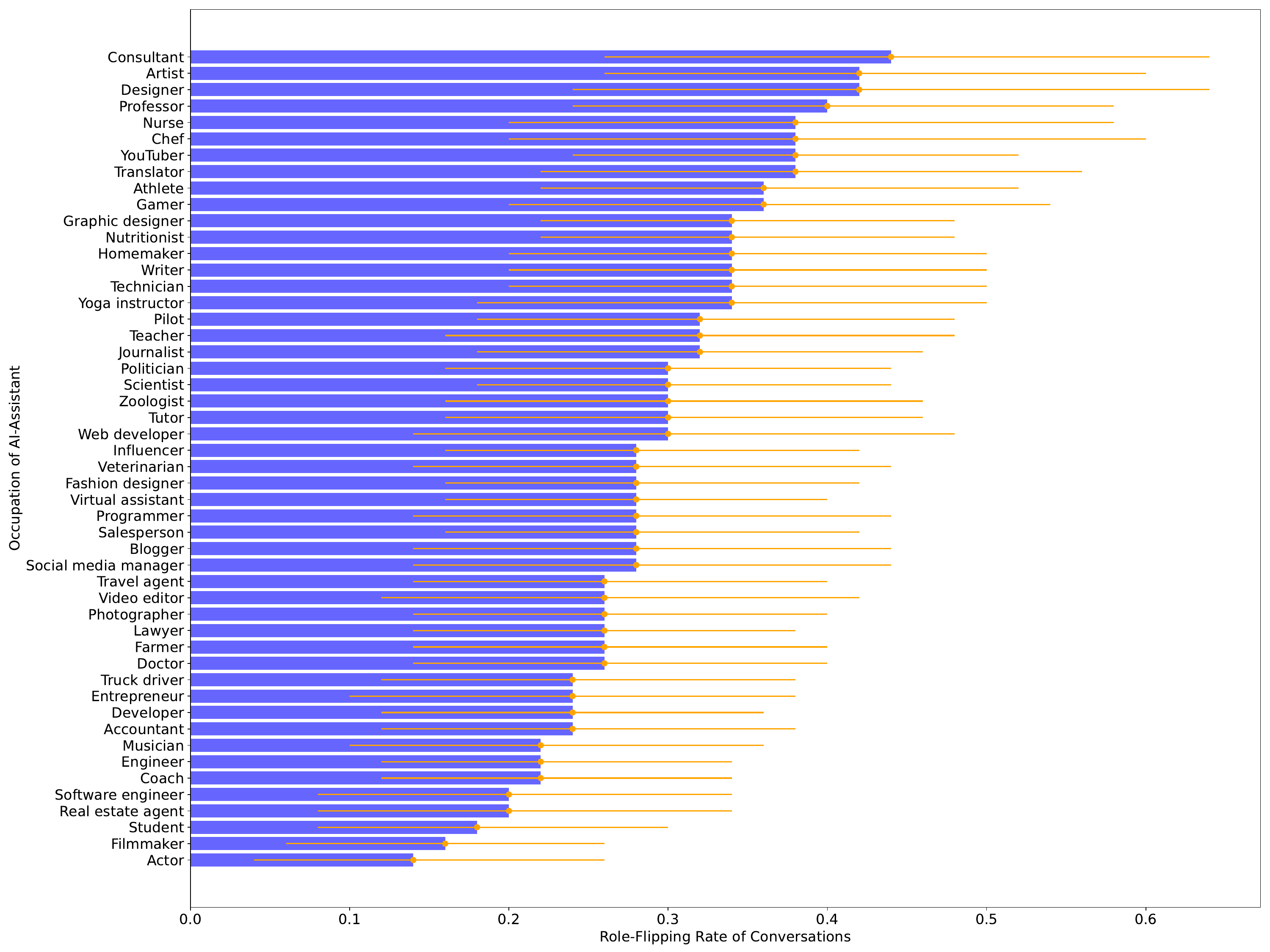}
    \caption{Average role-flipping rate per AI-Assistant occupation on the CAMEL AI-World dataset with 95\% confidence interval produced by bootstrap sampling.}
    \label{fig:camel-assistant-occupation}
\end{figure*}

\begin{figure*}
    \centering
\includegraphics[width=0.9\textwidth,height=0.9\textheight,keepaspectratio]{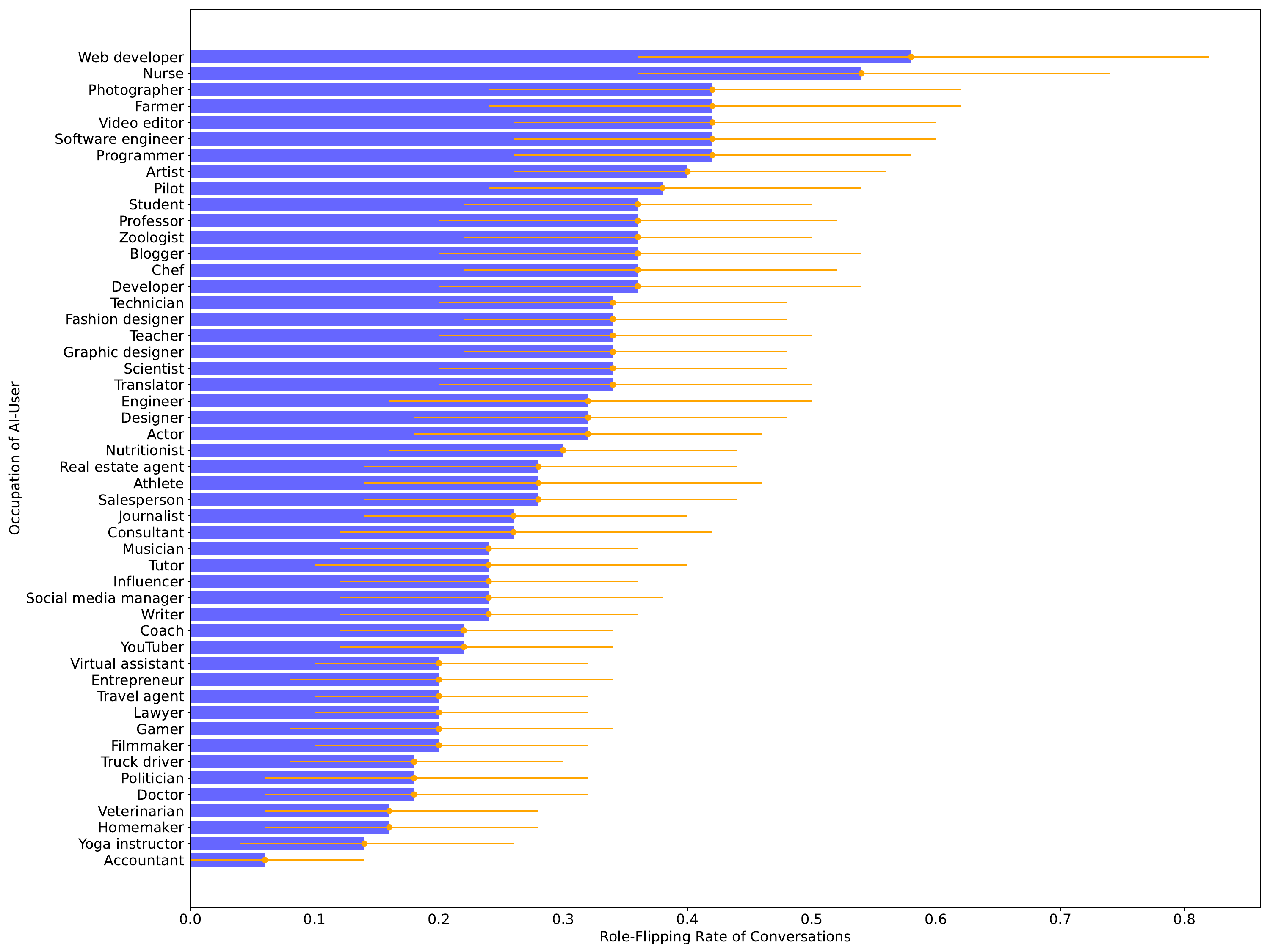}
    \caption{Average role-flipping rate per AI-User occupation on the CAMEL AI-World dataset with 95\% confidence interval produced by bootstrap sampling.}
    \label{fig:camel-user-occupation}
\end{figure*}

\end{document}